\documentclass{article}

\usepackage[numbers, sort&compress]{natbib}

\usepackage[final]{neurips_2023}

\usepackage{adjustbox}

\usepackage[toc,page,header]{appendix}
\usepackage{minitoc}

\usepackage[utf8]{inputenc} 
\usepackage[T1]{fontenc}    
\usepackage{hyperref}       
\usepackage{url}            
\usepackage{booktabs}       
\usepackage{amsfonts}       
\usepackage{nicefrac}       
\usepackage{microtype}      

\usepackage{subfigure}
\usepackage{caption}
\usepackage[dvipsnames]{xcolor}
\usepackage{algorithm}
\usepackage{algorithmicx}
\usepackage{algpseudocode}

\usepackage{hyperref}
\hypersetup{
    colorlinks = true,
    linkcolor = blue,
    anchorcolor = blue,
    citecolor = blue,
    filecolor = blue,
    urlcolor = blue
    }
\usepackage{hyphenat}
\usepackage{amsfonts,amsmath,amssymb}
\usepackage{hhline}
\usepackage{bm}

\usepackage{colortbl}

\usepackage{pifont}

\usepackage{bbm,enumerate}

\usepackage{physics}
\usepackage{siunitx}
\usepackage{wrapfig}

\usepackage{microtype}
\usepackage{graphicx}
\usepackage{booktabs} 
\usepackage{makecell}
\usepackage{tabularx}

\usepackage{colortbl}

\usepackage{booktabs, multirow} 
\usepackage{soul}
\usepackage{changepage,threeparttable} 

\usepackage{hyperref}

\newcommand{\xmark}{\textcolor{red}{\ding{55}}}%
\newcommand{\cmark}{\textcolor{ForestGreen}{\ding{52}}}%

\usepackage{pgfplots}
\usepackage[framemethod=TikZ]{mdframed}
\usepackage{comment}
\usepackage{fontawesome}
\usepackage{pgfplotstable}
\usepackage{colortbl}

\mdfsetup{%
    middlelinecolor =   none,
    middlelinewidth =   1pt,
    backgroundcolor =   blue!5,
    roundcorner     =   5pt,
}

\usepackage{amsmath}
\usepackage{amssymb}

\usepackage{amsthm}
\theoremstyle{definition}
\newtheorem{definition}{Definition}[section]

\theoremstyle{property}

\theoremstyle{assumption}
\newtheorem{assumption}{Assumption}[section]
\theoremstyle{remark}

\theoremstyle{note}

\theoremstyle{proposition}

\newcommand{\Clower}{C_{\mathrm{low}}}
\newcommand{\Cupper}{C_{\mathrm{up}}}

\newcommand{\X}{\mathcal{X}}
\newcommand{\Y}{\mathcal{Y}}

\newcommand{\G}{\mathcal{G}}
\newcommand{\Dtrain}{\mathcal{D}_{\textrm{train}}}
\newcommand{\Dtest}{\mathcal{D}_{\textrm{test}}}

\newcommand{\Dcal}{\mathcal{D}_{\textrm{cal}}}

\newcommand{\D}{\mathcal{D}}

\newcommand{\R}{\mathbb{R}}
\newcommand{\N}{\mathbb{N}}

\newcommand{\set}[1]{\left\{ #1 \right\}}

\title{TRIAGE: Characterizing and auditing training data for improved regression}

\author{%
  Nabeel Seedat\\
  University of Cambridge\\
  \texttt{ns741@cam.ac.uk} \\
   \And
   Jonathan Crabb\'{e} \\
   University of Cambridge \\
   \texttt{jc2133@cam.ac.uk} \\
   \AND
   Zhaozhi Qian \\
   University of Cambridge \\
   \texttt{zq224@cam.ac.uk} \\
   \And
   Mihaela van der Schaar \\
   University of Cambridge \\
   \texttt{mv472@cam.ac.uk} \\
}

\begin{document}

\doparttoc
\faketableofcontents

\maketitle

\begin{abstract}
Data quality is crucial for robust machine learning algorithms, with the recent interest in data-centric AI emphasizing the importance of training data characterization. However, current data characterization methods are largely focused on classification settings, with \emph{regression settings} largely understudied. To address this, we introduce TRIAGE, a novel data characterization framework tailored to regression tasks and compatible with a broad class of regressors. TRIAGE utilizes conformal predictive distributions to provide a model-agnostic scoring method, the TRIAGE score. We operationalize the score to analyze individual samples' training dynamics and characterize samples as under-, over-, or well-estimated by the model. We show that TRIAGE's characterization is consistent and highlight its utility to improve performance via data sculpting/filtering, in multiple regression settings. Additionally, beyond sample level, we show TRIAGE enables new approaches to dataset selection and feature acquisition. Overall, TRIAGE highlights the value unlocked by data characterization in real-world regression applications. 
\end{abstract}

\section{Introduction} \label{sec:introduction}
\textbf{Data characterization, an important problem.}  
Over a decade ago, it was recognized that ``more data beats clever algorithms, but better data beats more data''\cite{norvig2011}, emphasizing the key role of training data in the performance and robustness of ML algorithms \cite{jain2020overview,gupta2021dataB,renggli2021data}. 
In particular, training data from the ``wild'' may have various errors or lack coverage, rendering it unsuitable for the ML task \cite{chen2021data}.

These observations motivate the need for \textit{data characterization}, which aims to score each training sample based on its impact on the performance of the ML task. The training data can then be partitioned or rank ordered according to the score to give a high-level description of the sample and its utility. Data characterization provides a natural way for \textit{data sculpting} \cite{liang2022advances, seedat2022dc}, which filters training samples that have a low or even detrimental effect on the ML task. The goal is that models built on the sculpted/filtered (and smaller) training data have improved performance.

In this work, we address the \emph{understudied} data characterization problem in \textbf{regression settings}.
In contrast, existing works on data characterization have primarily focused on the classification setting \cite{datamap,seedatdata,mainicharacterizing,vog,aum,gradn}. 
These methods leverage the probability distribution over discrete labels (softmax or logits) for scoring. 
However, in the equally important regression setting, the probability distribution over the continuous labels are not typically given by standard regression models. In fact, the majority of the regression models only output a point estimate, e.g. the predicted mean. 
Without access to probabilistic estimates, one cannot simply apply the existing classification-tailored methods for discrete labels to the regression setting. We provide further details why classification-tailored methods are inapplicable to the regression setting in Sec. \ref{sec:related} and Table \ref{related_work}. This non-applicability motivates the need for regression-tailored methods for data characterization, which are ``not only principled, but also practical for a larger collection (...) of ML models'' \cite{renggli2021data}.

To anchor the need for data characterization in regression, let's consider a few concrete examples. In predicting house prices, an outlier might be a house with an extremely high price due to a unique or under-represented feature in the data. This could affect the model's performance, causing it to overestimate prices for similar houses without the unique feature. Alternatively, imagine predicting patient length of stay based on clinical features. If the dataset contains some erroneous features, the regression predictions may be suboptimal, potentially leading to incorrect clinical decisions. 

\begin{figure*}[t]
    \centering
    \vspace{-3mm}
    \includegraphics[width=0.90\textwidth]{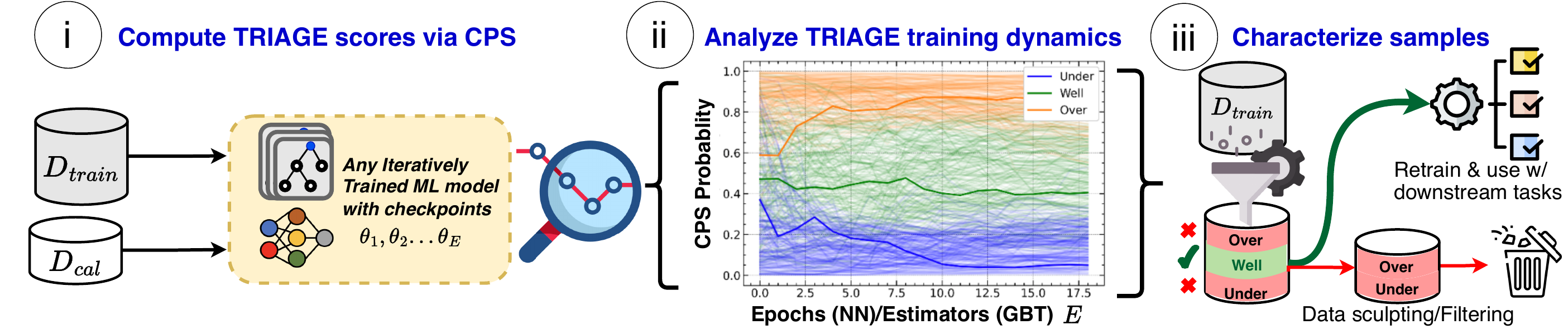}
    \vspace{-2mm}
    \caption{\footnotesize{TRIAGE systematically characterizes data in regression settings via a 3-step pipeline. \textcolor{blue}{(i)} Compute TRIAGE scores: of each sample in the dataset, for each regression checkpoint, via a Conformal Predictive System (CPS).  Over the training epochs/iterations, we \textcolor{blue}{(ii)} Analyze the per-sample training dynamics, studying the trajectory of CPS probability scores for each training sample, with the scoring method operationalized to characterize the samples. \textcolor{blue}{(iii)} Characterize samples, assigning them to different groups (over-, under- and well-estimated) based on the TRIAGE scores. These groups inform sculpting  \& other use-cases.}}
    \label{fig:overview}
     \vspace{-3mm}
    \rule{\linewidth}{.75pt}
    \vspace{-8mm}
\end{figure*}

In designing a data characterization framework for regression, we outline the following desired properties, drawing motivation from the capabilities of works in the classification setting \cite{mainicharacterizing, datamap,seedatdata,vog,aum,gradn}:
\begin{mdframed}[leftmargin=0pt, rightmargin=0pt, innerleftmargin=10pt, innerrightmargin=10pt, skipbelow=0pt]
\textbf{(P1) Consistency:} The characterization of data examples should be stable and consistent across runs and for similar performing models with different parametrizations.\\
\textbf{(P2) Improve performance:} The characterization should guide the improvements to model performance. For instance, via data sculpting/filtering.\\
\textbf{(P3) Informative for data collection:} The characterization score should be informative to enable selection between datasets, as well as, feature collection/acquisition.\\
\textbf{(P4) Regressor versatility:} The characterization should be principled and practical for a large collection of ML models (e.g. not just neural networks).
\end{mdframed} 
To fulfill \emph{P1-P4}, we propose a framework to characterize data in regression settings, called \textbf{TRIAGE} (s.f. \textbf{TR}aining \textbf{I}dentification \& \textbf{A}uditing of \textbf{G}ood \textbf{E}xamples) --- see Fig.\ref{fig:overview}. 
 TRIAGE introduces a \textbf{novel regression scoring method}---\emph{TRIAGE score},  building on the recent development of Conformal Predictive Systems (CPS) \cite{vovk2019nonparametric}. CPS is a model-agnostic method that can turn the output of \emph{any} regressor into a probability distribution for continuous labels, termed conformal predictive distribution (CPD). Unlike point estimates such as errors/residuals or conventional conformal prediction intervals, the CPD provides richer information through a full predictive distribution. This enables us to assess the probability of a prediction being over- or under- estimated. 

We then operationalize the CPD (and TRIAGE score) in a novel way for data characterization, computing the trajectory of TRIAGE scores over training checkpoints, as a measure of how samples are learned differently during training. Such \emph{training dynamics}, shown in Fig \ref{fig:overview} and have been shown to contain a salient signal about the value of the sample to a learner \cite{arpit2017closer,arora2019fine,li2020learning}. By analyzing the training dynamics, we then categorize training samples as: \emph{under-}, \emph{over-} or \emph{well-estimated} by the model. This is useful going back to our motivating example --- where erroneous values could cause the model to overestimate house price. Using the TRIAGE score, we can identify such a sample, as it would likely have a low probability of being correct based on the model's predictive distribution. Such identification unlocks a variety of use cases. For instance, sculpting/filtering such samples can improve model performance (even with less data), as we empirically show in  Sec. \ref{sec:experiments}. 

\textbf{Contributions.} \textbf{\textcolor{ForestGreen}{\textcircled{1}}} \textbf{\emph{Conceptually}}, we present TRIAGE, the first data characterization framework tailored to regression settings, addressing the need for an ML-aware data quality that is principled and practical across a large collection of ML regressors. \textbf{\textcolor{ForestGreen}{\textcircled{2}}} \textbf{\emph{Technically}},  we introduce a principled regression scoring method called TRIAGE score, that leverages conformal predictive distributions in a novel way for data characterization. We operationalize the score to analyze how different samples are learned during training, thereby permitting characterization of individual samples as: \emph{under-}, \emph{over-} or \emph{well-estimated} by the model. \textbf{\textcolor{ForestGreen}{\textcircled{3}}} \textbf{\emph{Empirically}}, we demonstrate the utility of TRIAGE across multiple use cases satisfying \emph{P1-P4}, including consistent characterization, sculpting to improve performance in a variety of settings, as well as, guiding dataset selection and feature acquisition.

\section{Related work} \label{sec:related}
\begin{table*}[!t]
\centering
\vspace{-4mm}
\caption{Comparing data characterization for classification to TRIAGE (a tailored regression method)}
\scalebox{0.6}{
\begin{tabular}{l|c|c|cc|c}
\toprule
        Type of method
      & \makecell{Approaches} 
      & \makecell{Scoring \\ method} 
      & \makecell{Can be repurposed \\ for regression}    
      &  \makecell{Applicable to \emph{any} ML \\ model trained iteratively} 
      & \makecell{Reason for \\ non-applicability }  
                        \\ \midrule
Tailored regression & TRIAGE (Ours) & CPS probability  & \cmark & \cmark  & -  \\ \hline
 \multirow{4}{*}{Tailored classification} & Data Maps \cite{datamap}  & Model uncertainty   & \xmark & \cmark   & Needs label probability distribution  \\
& Data-IQ \cite{seedatdata} & Data uncertainty  & \xmark & \cmark   & Needs label probability distribution   \\

& SSFT \cite{mainicharacterizing}          & Forgetting scores  & \xmark   & \cmark & Needs discrete class changes   \\
& AUM \cite{aum}          & Logit margin  & \xmark   & \xmark & Needs logit output     
\\ \hline
 \multirow{3}{*}{General-purpose} & Metadata Archaelogy \cite{metadata}            & Per-sample losses  & \cmark    & \cmark & -   \\
& VoG \cite{vog} & Variance of gradient  & \cmark   & \xmark & Needs differentiable model (e.g. NN)    \\
& GraNd \cite{gradn}            & Gradient Norm &  \cmark & \xmark   &  Needs differentiable model (e.g. NN)   \\
\bottomrule
\end{tabular}}
\label{related_work}
\vspace{-3mm}
\end{table*}

This work engages with recent efforts in data-centric AI \cite{seedat2022dc,liang2022advances,ng2021,seedat2022DS} on data characterization. However, prior methods have focused on classification settings. In contrast, TRIAGE addresses the \emph{regression setting} where tailored data characterization has not been explored.
We now discuss related works and outline the challenges of applying prior classification focussed methods to regression, for an overview see Table \ref{related_work}. Prior classification scoring methods can be divided into two groups:\\
$\blacksquare$ \textbf{Tailored methods}, assign scores based on aspects specific to classification, including logits \cite{aum}, probabilities over labels \cite{datamap,seedatdata} or changes to predictions on discrete labels \cite{toneva2018empirical,mainicharacterizing}. However, these methods are inapplicable to our regression setting because these aspects are \emph{not} present in regression. \\
$\blacksquare$ \textbf{General purpose methods}, while focussed on classification, they assign scores using generic measures and can be repurposed for regression. However, these general methods based on backpropagated gradients \cite{vog,gradn} or losses \cite{metadata} have two limitations. Firstly, scores like gradients can only be used with differentiable models (e.g. neural networks) and are not compatible with many widely used regressors, such as XGBoost --- i.e. does not satisfy \emph{(P4) Versatility}. Secondly, even if they can be used, general-purpose scores may not distinguish between samples that have similar magnitude scores (e.g. high loss scores), even if the underlying reasons for these scores may be different with respect to the task.  We show the value of such a differentiation between samples in Sec.\ref{p2-exp}.

\textbf{Data valuation.} A related area to data characterization is that of data valuation, for example: Shapley based valuation methods \cite{datashap,knnshap}. We first outline some conceptual differences: (a) TRIAGE unlocks other data-centric tasks beyond just data selection, e.g. feature-level acquisition/collection and selection between datasets. These are out of scope for data “sample” selection; (b) Unlike TRIAGE, these methods are not tailored to regression; (c) method differences: computationally Shapley-based methods need to assess multiple sample permutations and need to retrain the model many times. Hence, they struggle to scale to very high-samples sizes (e.g. 100k), unlike TRIAGE where the cost is cheap comparatively. Recently, a valuation method LAVA \cite{lava} has been introduced to address this which uses an additional embedding model to reduce dimensionality before the optimal transport step. Beyond conceptual differences, we compare TRIAGE experimentally and in terms of computation time in Appendix \ref{sec:appendixC}.

\section{Data Characterization Preliminaries} \label{sec:formulation}
In this section, we formulate the data characterization problem for regression settings. Consider the standard regression setting, where the aim is to assign an input $x \in \X \subseteq \R^{d_X}$ to a continuous valued label $y \in \R$ (often a point estimate for $y$). This is done using a regressor $f_\theta: \X \rightarrow \Y$, parameterized by $\theta \in \Theta$. 
We assume $f_\theta$ is trained on a dataset $\mathcal{D}$  with $M \in \mathbb{M}^*$ observations, i.e. $\mathcal{D} = \set{(x^m, y^m) \mid m \in [M]}$. For the purposes of later notation, we consider a sample space $\mathcal{Z}:= \X \times \R$, where each observation $z^m = (x^m,y^m) \in \mathcal{Z}$. Typically, parameters $\theta$ are learned in regression, via empirical risk minimization (ERM) with a mean squared error (MSE) loss, i.e. $\text{ERM}(\theta) = \arg \min_{\theta \in \Theta} \frac{1}{M} \sum_{m=1}^M \ell(x^m, y^m; \theta)$,
with loss function $\ell : \X \times \Y \times \Theta \to \R^+$.

ERM ensures regressor $f_\theta$ performs well \emph{on average}, however it goes without saying that the model is unlikely to perform equally well for all samples in the dataset. Consequently, we desire to characterize such types of samples. Formally, for each sample $x^m$, we aim to assign a \emph{group} label $g^m\,{\in}\,\mathcal{G}$, where $\mathcal{G}=\{$\text{\it Over-estimated, Well-Estimated, Under-estimated}$\}$. We then use these groups for downstream tasks, or to sculpt/filter samples in the dataset $\mathcal{D}$ based on their group assignment. Before giving a precise description of how the group labels are assigned, we provide some context.  

Prior works have shown the behavior of samples over the course of model training, called \emph{training dynamics} contains signal about the nature of a sample itself \cite{arpit2017closer,arora2019fine,li2020learning}.
For instance, some samples are easily learned and stable through training, whereas other samples might be challenging and have high variance. Many classification-based methods leverage this concept and assign $g^m$ by analyzing training dynamics e.g. Data-IQ, Data Maps, AUM etc.  However, as discussed in Secs. \ref{sec:introduction} and \ref{sec:related}, the aforementioned classification methods \emph{do not} apply to regression. Consequently, this requires us to formalize a new principled scoring method for regression. After that, we can operationalize the scoring method to analyze the training dynamics and assign group labels $g^m$ to each $x^m$. 

For regression, such a principled scoring method could evaluate the probability that the true target $y$ is less than or equal to observation $f(x)$, where the categories of samples are: \textcircled{1} \emph{Under-estimated}: $P(y\leq f(x))\gg~0.5$, \textcircled{2} \emph{Well-estimated}: $P(y\leq f(x))\sim~0.5$, \textcircled{3} \emph{Over-estimated}: $P(y\leq f(x))\ll~0.5$. In the next section, we discuss how we design TRIAGE to achieve this, with the final goal in mind --- assigning group labels $g^m$ to each sample $x^m$.

\section{TRIAGE: characterizing data for regression settings} \label{sec:method}
\vspace{-2mm}
We now introduce TRIAGE, a regression-focused data characterization framework, compatible with \emph{any} iteratively trained ML regressor (Sec. \ref{p4-formulation}). TRIAGE follows a three-step procedure:\\
1.~\textbf{Compute TRIAGE scores} (Sec \ref{cps}): Compute the TRIAGE score for every sample $x^m$ using conformal predictive distributions. Do this for every regression checkpoint to obtain a trajectory. \\
2. \textbf{Analyze TRIAGE score training dynamics} (Sec \ref{sec:training-dynamics}): 
Operationalize the TRIAGE score for data characterization by analyzing the TRIAGE score's trajectory for each training sample. \\
3. \textbf{Characterize \& assign groups} (Sec \ref{formulation:stratification}): Characterize samples based on their training dynamics, assigning each sample $x^m$ to a group $g^m$. The groups inform filtering/sculpting or other use cases.
\vspace{-2mm}

\subsection{Computing TRIAGE scores via Conformal Predictive Systems (Step 1).} \label{cps}

Recall, we want to compute $P(y \leq f(x))$ as a principled scoring method to characterize samples. We want to do so \emph{without} changing the chosen regressor type, i.e. \textbf{P4: Regressor versatility}. Of course, the full information of an uncertain label exists in its probability distribution, allowing us to compute probabilities of various outcomes.
Unfortunately, regular regressors (\emph{point predictions}) or conventional conformal regressors (\emph{prediction intervals}) do not provide this distribution. While Bayesian methods offer distributions, they require fundamentally changing the regressor.

These challenges motivate defining TRIAGE around Conformal Predictive Systems (CPS)\cite{vovk2019nonparametric}. CPS combines predictive distributions \cite{shen2018prediction,schweder2016confidence} with conformal prediction \cite{vovk2005conformal}, crucially enabling us to compute the required probability distribution of a continuous variable—called Conformal Predictive Distribution (CPD). CPS also fulfills our versatility goal (i.e. \emph{P4}) as it applies to \emph{any} regressor.  To clarify Conformal Predictive Systems output valid cumulative distribution functions, which are termed Conformal Predictive Distributions (CPD). This is the cumulative probability with respect to a label $y$
 , given some $x$
 and regressor $f$
. With CPDs denoted as $Q$
, the conformal p-values get arranged into a probability distribution which has the properties of a CDF — thus essentially becoming probabilities, see \cite{vovk2020computationally}. Since the CPD has the properties of a CDF, we use the CPD to estimate probabilities that the true target 
is less than or equal to a specified threshold/value. 

\textbf{Computing CPDs.}
TRIAGE formalizes the CPD computation in the framework of the computationally efficient \emph{Split Conformal Predictive Systems} \cite{vovk2020computationally}. A \emph{proper training set} $\mathcal{D}_{\mathrm{tr}} = \{(x^i_{\mathrm{tr}},y^i_{\mathrm{tr}})=z^i_{\mathrm{tr}} \mid i \in [M]\}$ is used to train regressor $f$. A separate, often smaller,  \emph{calibration set} $\mathcal{D}_{\mathrm{cal}}= \{(x^i_{\mathrm{cal}},y^i_{\mathrm{cal}}) = z^i_{\mathrm{cal}}\mid i \in [q]\}$, is used for conformal calibration. 
Finally, we define a \emph{conformity measure} $\mu$, where $\mu:\mathcal{Z}\to\R\cup\{-\infty,\infty\}$, quantifies the dataset's agreement with the observation. For each label $y^i_{\mathrm{cal}} \mid i \in [q]$, we compute $q$ conformity scores $\alpha_i=\mu(z^i_{\mathrm{cal}})$. We define the conformity measure ($\mu$) as per Eq. \ref{ncs}, which is a balanced isotonic conformity measure. We discuss this choice in Appendix \ref{cpd-extra}, a necessary condition s.t the CPD can be interpreted as a probability distribution \cite{vovk2020computationally} \footnote{Under the standard conformal exchangeability assumption (see Appendix \ref{cpd-extra}), the CPDs are valid.}.
\begin{equation}
    \mu(x,y) = \nicefrac{(y-f(x))}{\sigma(x)}
    \label{ncs}
\end{equation}
\begin{wrapfigure}{r}{0.55\linewidth}
\centering
\vspace{-3mm}
\scalebox{0.85}{%
\begin{minipage}{1.\linewidth}
\begin{algorithm}[H]
  \caption{Computing a CPD}
  \label{alg:SCPS}
  \begin{algorithmic}
    \Require
      Training set $(x^i_{\mathrm{tr}},y^i_{\mathrm{tr}})\in\Dtrain$, $i\in [M]$ and Calibration set $(x^i_{\mathrm{cal}},y^i_{\mathrm{cal}})\in\Dcal$, $i\in [q]$.
    \State 1: Regressor $f$ is trained on $\Dtrain$. \Comment{\colorbox{pink}{Training}}
    \State 2: Compute residuals on $\Dcal$: $\scriptstyle R_i=y^i_{\mathrm{cal}} - f(x^i_{\mathrm{cal}}), i\in [q]$
    \State 3: Define $\sigma(x)$; which finds the average residuals of the k-nearest neighbors $x$, where KNN is fit on $\Dcal$. 
    \State 4: Calculate conformity scores $\scriptstyle \alpha_1, \alpha_2 \dots \alpha_q$ over $\Dcal$ using $\mu$; i.e. $\scriptstyle \alpha_i = \mu (x^i_{\mathrm{cal}}, y^i_{\mathrm{cal}}), 
 i\in [q]$. \Comment{\colorbox{GreenYellow}{Calibration}}
    \State 5: Sort $\scriptstyle \alpha_{1},... \alpha_{q}$ in ascending order ; s.t $\scriptstyle \alpha_{(1)}<...<\alpha_{(q)}$
    
    \For{eval sample $x\in\mathbf{X}_{train}$} \Comment{\colorbox{SkyBlue}{Evaluation}}
    \State Let $S_{(i)}:= f(x) + \alpha_{i}\sigma(x)$ for $i\in [q]$.
    \State Let $S_{(0)}:=-\infty$ and $S_{(q+1)}:=\infty$.
    \EndFor
    \State Return the CPD as per Equation \eqref{pred_dist}.
  \end{algorithmic}
\end{algorithm}
\end{minipage}}
\vspace{-5mm}
\end{wrapfigure}
where $y$ is the label, $f(x)$ a predicted label, $\sigma$ an estimate of prediction quality, computed with a separate normalization model, enabling CPDs to reflect task difficulty. $\sigma$ is kNN fitted on the residuals of $\Dcal$, similar to \cite{johansson2014regression}.

We compute the CPD denoted $Q^m$ (Eq. \ref{pred_dist}) for each sample $x^m$ via Algorithm~\ref{alg:SCPS}, where the CPD is defined for $y^* \in \R$ \cite{vovk2020computationally}. We extract the TRIAGE score using the CPD. 
We repeat the CPD computation for all samples.  Next, we describe our novel approach to operationalize the TRIAGE scores, for the purpose of data characterization. 
\vspace{3mm}
  \begin{equation}\label{pred_dist}
    \footnotesize
      Q(y^*,\tau) :=
  \begin{cases}
    \frac{i+\tau}{q+1} & \text{if $y^*\in(S_{(i)},S_{(i+1)})$; $i\in[0:q]$}\\
    \frac{i'-1+(i''-i'+2)\tau}{q+1} & \text{if $y^*=S_{(i)}$; $i\in[1:q]$}
  \end{cases}
    \end{equation}
    \label{eq:equation_label}
        where $\tau \in [0,1]$, $i':=\min\{j\mid S_{(j)}=S_{(i)}\}$ and \\ $i'':=\max\{j\mid S_{(j)}=S_{(i)}\}$ to account for ties.

\subsection{Analyzing TRIAGE score trajectories --- training dynamics (Step 2)}  \label{sec:training-dynamics}

\textbf{Novel use of CPD's for data characterization.}  The \textbf{\emph{TRIAGE score}}, denoted as $T(x,y,\theta) = P(y \leq f_{\theta}(x))$, represents a novel use of CPDs for data characterization, diverging from their original predictive uncertainty use. Additionally, while CPDs are normally calculated post-model fitting, we observe from Fig.\ref{fig:overview}(ii) that samples' learning varies, leading to different convergence rates of scores. As discussed in Sec. \ref{sec:formulation}, such dynamics reveal sample characteristics \cite{arpit2017closer,arora2019fine,li2020learning}. Hence, we propose to examine TRIAGE score trajectories to capture the learning dynamics of different samples.

To enable this analysis, we utilize the model checkpoints of \emph{any} ML regressor trained over iterations  (e.g. neural networks, XGBoost etc). Specifically, during iterative training, the model parameters $\theta$ vary over $E \in \N^*$ iterations, taking $E$ different 
parameter values at each checkpoint, i.e. $\theta_1 , \theta_2,..., \theta_E$. We wish to account for the impact of these changing parameters.  At each checkpoint $\theta_e$, we compute the CPD to obtain the TRIAGE score for each sample $x^m$. We repeat the computation for every checkpoint to obtain a TRIAGE score trajectory over checkpoints per-sample of 
 dimensionality $E$. i.e. we evaluate the function $Q$ for a specific $f_{\theta_{e}}(x)$ to get the estimated probability $P(y \leq f_{\theta_{e}}(x))$. 
 We repeat for all $f_{\theta_{e}}(x)$ checkpoints where $e \in E$ to get the trajectory of TRIAGE scores for sample.

To quantify and contrast the behavior of the different sample trajectories, we compute two metrics to summarize the per-sample trajectory (see Fig. \ref{fig:characteristic curve}), namely (1) \textbf{\emph{Confidence}}: reflects the degree of confidence wrt to $P(y \leq f_{\theta}(x))$. It is computed as the mean of the TRIAGE score over training, i.e. $C(x^i, y^i)=\frac{1}{E} \sum_{e=1}^E T(x^i,y^i,\theta_e)$. (2) \textbf{\emph{Variability}}: reflects the consistency or fluctuation of the score for a specific sample. It is computed as the standard deviation of the TRIAGE score over training, i.e. $V(x^i,y^i)= \sqrt{\frac{\sum_{e=1}^E (T(x^i,y^i,\theta_e)-C(x^i, y^i))^2}{E}}$. We describe next how these two metrics form the basis of sample characterization and group assignment.

\subsection{Characterization \& group assignment (Step 3)}\label{formulation:stratification}

\begin{wrapfigure}{r}{0.375\textwidth}
  \centering
    \centering
    \vspace{-2mm}
    \includegraphics[width=0.375\textwidth]{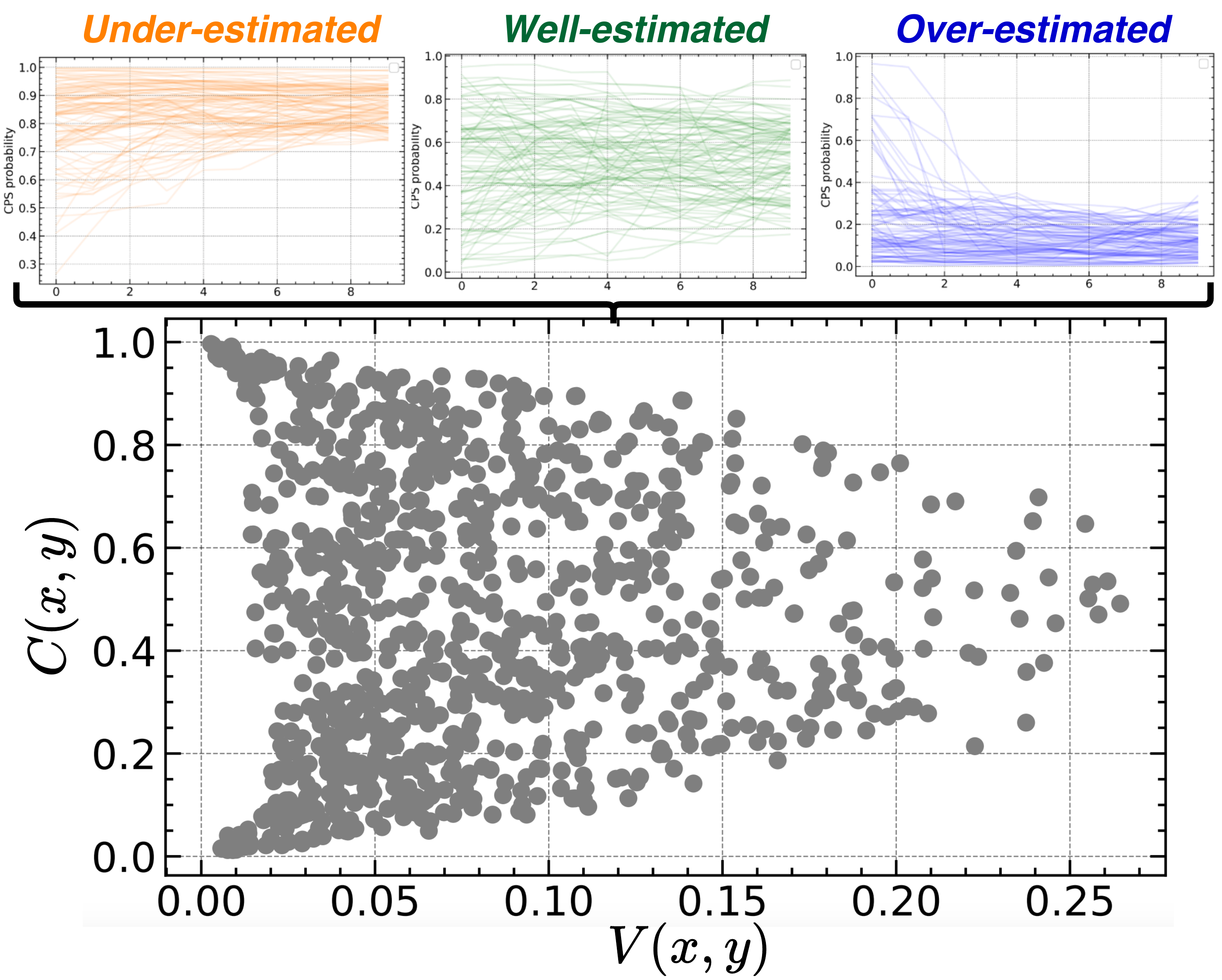}
    \vspace{-4mm}
    \caption{\footnotesize{Training dynamics of samples (above) are summarized in a characteristic curve (below-grey)}}
    \vspace{-2mm}
    \rule{\linewidth}{.4pt}
    \label{fig:characteristic curve}
    \vspace{-15mm}
\end{wrapfigure}

We now discuss how the two characterizing metrics, $C$ and $V$, are used to assign a group label $g \in \G$ to each training sample $x^m$. Let's intuitively define each group. \textcircled{1} \emph{Under-estimated (UE)}: low variability $V$ (i.e. decisive about) and HIGH average TRIAGE score $C$, i.e. $P(y \leq f(x)) \gg 0.5$, \textcircled{2} \emph{Over-estimated (OE)}: low variability $V$ (i.e. decisive about) and LOW average TRIAGE score $C$, i.e. $P(y \leq f(x)) \ll 0.5$,  and \textcircled{3} \emph{Well-estimated (WE)}:  average TRIAGE score $C$ is uncertain $P(y \leq f(x)) \sim 0.5$. In addition, these samples are often linked to high variability, implying uncertainty about $P(y \leq f(x))$. We empirically observe samples with these different traits per group, depicted in the characteristic curve, see Fig. \ref{fig:characteristic curve}.

We take a threshold-based approach to assign $g^m$, similar to previous characterization works \cite{datamap,seedatdata}:
\vspace{-1mm}
\begin{align} \label{eq:groups_training}
\small
    g^m = 
    \begin{cases} 
\mathrm{UE} & \text{if} \ C(x^m ,y^m) \ge \Cupper \wedge  V(x^m,y^m) < P_{n}^{V(\mathrm{\Dtrain})}   \\
\mathrm{OE} & \text{if} \ C(x^m,y^m) \le \Clower \wedge V(x^m,y^m) < P_{n}^{V(\mathrm{\Dtrain})}  \\
\mathrm{WE} & $\text{otherwise}$  \\
\end{cases}
\vspace{-9mm}
\end{align}
where $\Cupper$ and $\Clower$ are upper and lower confidence threshold resp. and $P_{n}$ the n-th percentile. We discuss the selection of $\Cupper$ and $\Clower$ in Appendix \ref{sec:appendixA}.

In Sec. \ref{sec:experiments}, we empirically show how the groups assigned to each sample can be used to sculpt/filter the training dataset or for other downstream tasks such as dataset selection or feature acquisition.

\subsection{Using TRIAGE with a variety of regressors (P4) }\label{p4-formulation}

The \emph{general-purpose} characterization methods discussed are largely compatible with differentiable models (neural networks). However, in real-world settings (e.g., healthcare, finance), practitioners often employ other iterative algorithms, like XGBoost/GBDTs \cite{borisov2021deep}. TRIAGE addresses this limitation, offering a model-agnostic approach using CPS --- as the conformity scores can be computed for any regressor. Of course, our use of training dynamics assumes regressor $f$ can be checkpointed. This still confers compatibility with a broader class of models: neural nets, XGBoost, GBDTs, linear regression ---  \emph{satisfying P4}. We underscore TRIAGE's versatility beyond neural networks, allowing practitioners to use TRIAGE with their preferred model for a specific application. Appendix \ref{adapt} details the ease of usage and space and time considerations.

\section{Empirical investigation} \label{sec:experiments}
This section presents an empirical evaluation demonstrating TRIAGE  
\footnote{Code: https://github.com/seedatnabeel/TRIAGE or https://github.com/vanderschaarlab/TRIAGE}
satisfies \textbf{(P1)} Consistency, 
\textbf{(P2)} Improved regression performance and 
\textbf{(P3)} Informative for data collection. Recall \textbf{(P4)} Regressor versatility is satisfied by TRIAGE's design.  Additional use-cases and results are in Appendix \ref{sec:appendixC}.

\textbf{Datasets.} 
We conduct experiments on \textbf{10 real-world regression datasets} with varying characteristics. i.e. different sample sizes (500-100k), dimensionality (8-85), and task difficulty. The datasets are drawn from diverse domains, including safety-critical medical regression: (i)~Prostate cancer from the US \cite{duggan2016surveillance} and UK \cite{prostate}, (ii)~Hospital Length of Stay \cite{microsoft_2017} and (iii)~MIMIC Antibiotics \cite{johnson2016mimic}. Additionally, we analyze general UCI regression datasets \cite{uci}, including Bike, Boston Housing, Bio, Concrete, Protein and Star. The datasets are detailed in Appendix \ref{sec:appendixB}, along with further experimental details. We observe similar performance across different datasets. However, due to space limitations, we sometimes highlight results for a subset and include results for the remainder in Appendix \ref{sec:appendixC}.

\subsection{\textbf{(P1)} Consistent characterization}\label{p1-exp}

\begin{wrapfigure}{r}{0.33\textwidth}
\captionsetup{font=footnotesize}
\vspace{-12mm}
  \centering
    \includegraphics[width=0.33\textwidth]{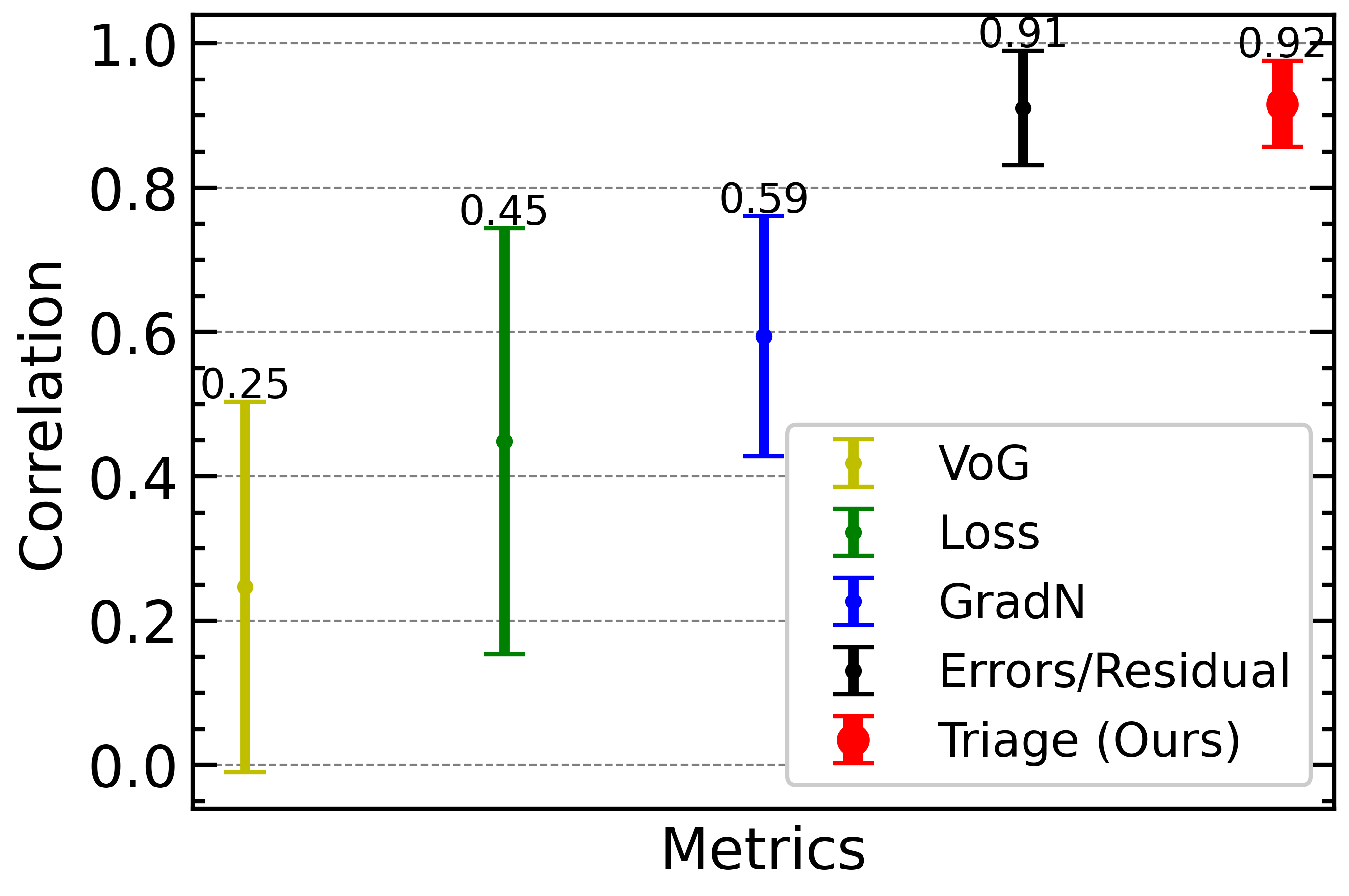}
    \vspace{-5mm}
   \caption{\footnotesize{Stability of scoring methods. TRIAGE is most consistent (highest Spearman correlation of \emph{0.91} \& lowest variation) averaged across datasets. Results on all 10 datasets are in Appendix \ref{consistency-indiv}. }}
   \vspace{-2mm}
    \rule{\linewidth}{.5pt}
    \label{fig:comparison}
    \vspace{-10mm}
\end{wrapfigure}
A key aspect of data characterization is: the ordering of samples matter \cite{mainicharacterizing,seedatdata}.  This is important when the scores are utilized for data sculpting/filtering. For example, if we rank sort samples by their scores and filter a subset of samples. Consequently, it is crucial to ensure the scoring method used for data characterization is stable and consistent. This consistency is desired across different seeds, but also different practitioners may adopt diverse model architectures/parameterizations for data characterization. 

\textbf{Consistency of TRIAGE.}  We compare the consistency and robustness to variation of TRIAGE vs baseline \emph{general purpose} methods: VoG \cite{vog}, GradN \cite{gradn}, Loss \cite{metadata}. We also consider simply computing the end of training Residual Errors (used in the TRIAGE conformity score).  We assess the consistency of the scores across different model architectures/parameterizations (described in Appendix \ref{extra-details}).  All models are trained to convergence with early stopping. Similar to prior classification-based characterization methods \cite{mainicharacterizing,seedatdata}, we assess consistency based on the Spearman rank correlation of the scores evaluated for all model combinations.

Fig.~\ref{fig:comparison}, shows TRIAGE is the most consistent compared to baselines, with the highest Spearman correlation (0.91), \emph{satisfying P1}. This consistency means insights derived using TRIAGE scores will also remain consistent. 
Analyzing each of the 10 datasets (see Appendix \ref{consistency-indiv}), further shows that  TRIAGE is stable in magnitude and rank order, whereas  the baselines \emph{do not} have such consistency both in magnitude or ordering across datasets, making the baselines challenging to use in practice.

\textbf{\textcolor{ForestGreen}{Takeaway 1.}}~TRIAGE has the most consistent scoring method compared to baselines over different datasets and model combinations, satisfying P1. 

\begin{wrapfigure}{r}{0.4\textwidth}
\captionsetup{font=footnotesize}
\vspace{-5mm}
    \centering
    \includegraphics[width=0.4\textwidth]{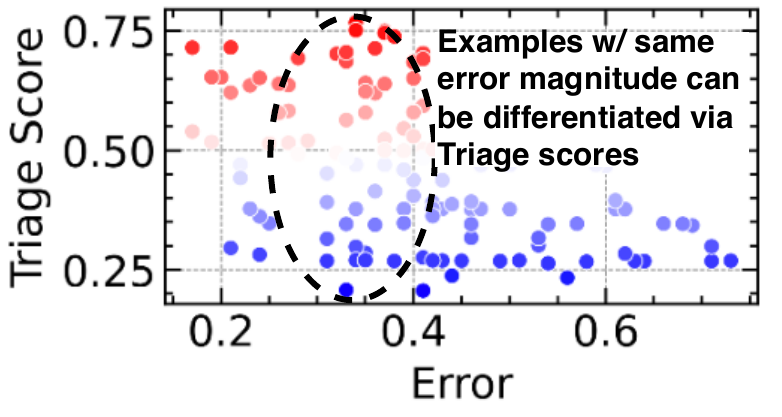}
    \vspace{-5mm}
   \caption{\footnotesize{Samples with the same error, have different Triage scores, highlighting the potential to differentiate samples}}
     \label{fig:complement}
    \vspace{-2mm}
    \rule{\linewidth}{.5pt}
    \vspace{-7mm}
\end{wrapfigure}
\textbf{TRIAGE v Errors/Residuals.} We note that TRIAGE and Errors/Residuals (considered an ablation of TRIAGE's dynamics component) have similar correlations.  This raises two questions: \textit{(i) are the samples identified the same?} We stratify the samples in $\Dtrain$ into three equal sized groups ordered by TRIAGE or Error scores. The overlap is $\sim$0.77, highlighting a difference. This leads us to ask \textit{(ii) what added value does TRIAGE offer?}
Consider the case of different samples that have the same error magnitude. 
Errors/Residuals treat these samples the same, based solely on magnitude. Can TRIAGE offer greater precision to further differentiate samples with the same magnitude?
As shown in Fig.\ref{fig:complement}, TRIAGE scores evidently offer a way to handle samples differently, showing that samples with similar residuals/errors are associated with different TRIAGE scores. We show the phenomena holds across other \emph{general-purpose} scoring rules in Appendix \ref{diff_other_scores}. This result demonstrates the added value of the \emph{training dynamics} aspect, to differentiate samples with the same error magnitude at a more granular level. This contrasts, errors/residuals which are an ablation of TRIAGE, computing the conformity score \emph{only} at the final training checkpoint, instead of for \emph{all} checkpoints.
We explore the value of this differentiation to improve regressor performance next.

\subsection{\textbf{(P2)} Improve regression performance}\label{p2-exp}
We now evaluate the ability of TRIAGE to improve regression performance by sculpting/filtering the data --- \emph{by keeping only well-estimated samples}. We study two setups using TRIAGE: (1) fine-grained filter (as described above) and (2) sculpting the data based on a deployment purpose.
\vspace{-0mm}

\subsubsection{Fine-grained filtering}

\begin{wrapfigure}{r}{0.65\textwidth}
\vspace{-14mm}
  \centering
  \subfigure[
\scriptsize{Errors}]{\includegraphics[width=0.31\textwidth]{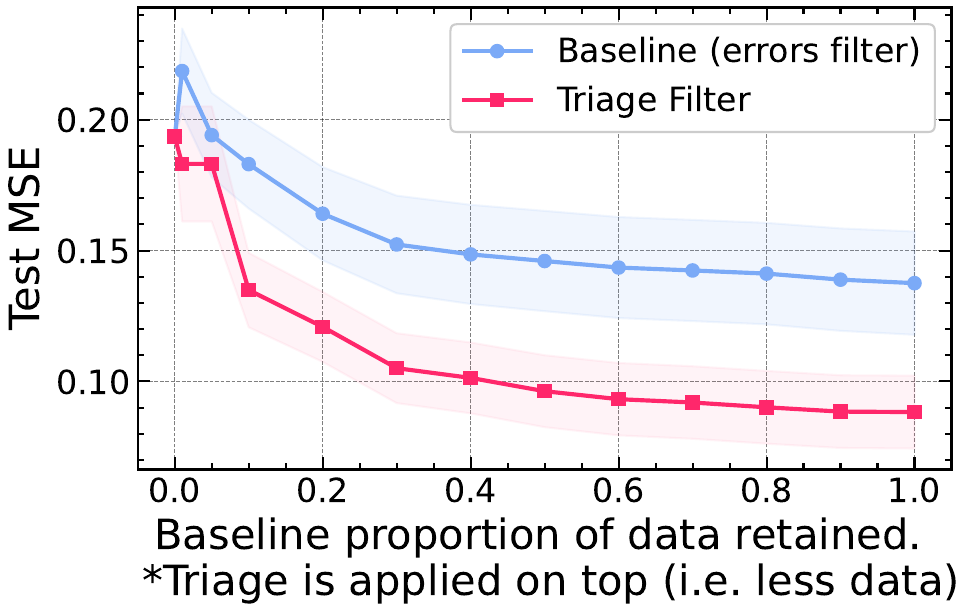}}\quad
  \subfigure[\scriptsize{Loss} ]{\includegraphics[width=0.31\textwidth]{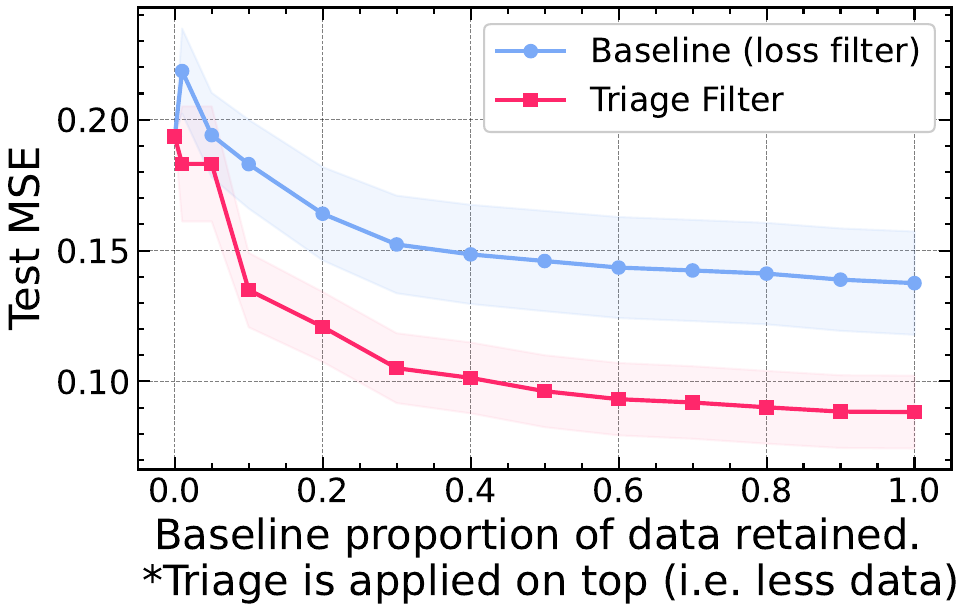}}\quad\\
  \vspace{-2mm}
  \subfigure[\scriptsize{Grad}]{\includegraphics[width=0.31\textwidth]{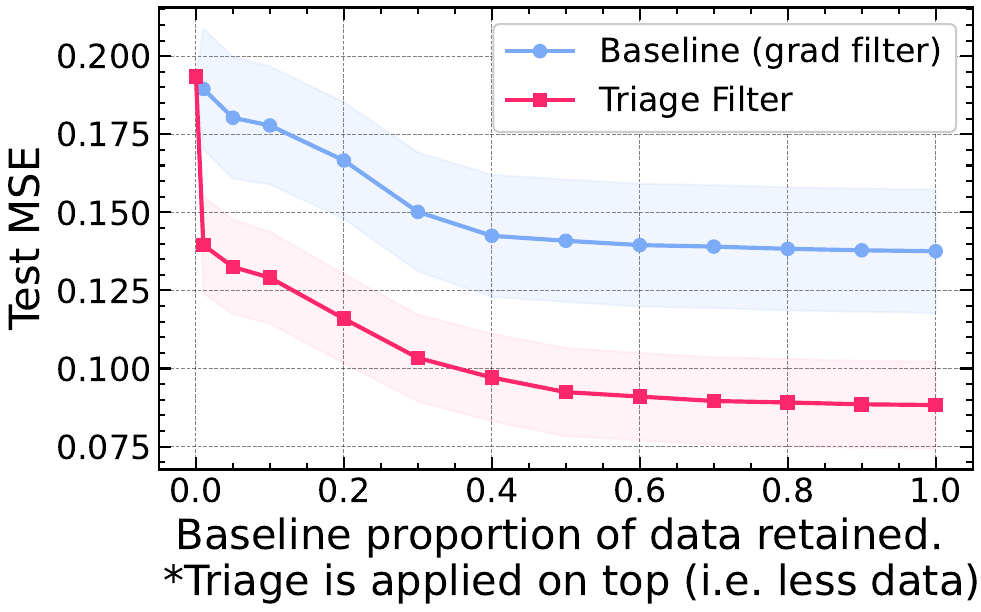}}\quad
  \subfigure[\scriptsize{VoG} ]{\includegraphics[width=0.31\textwidth]{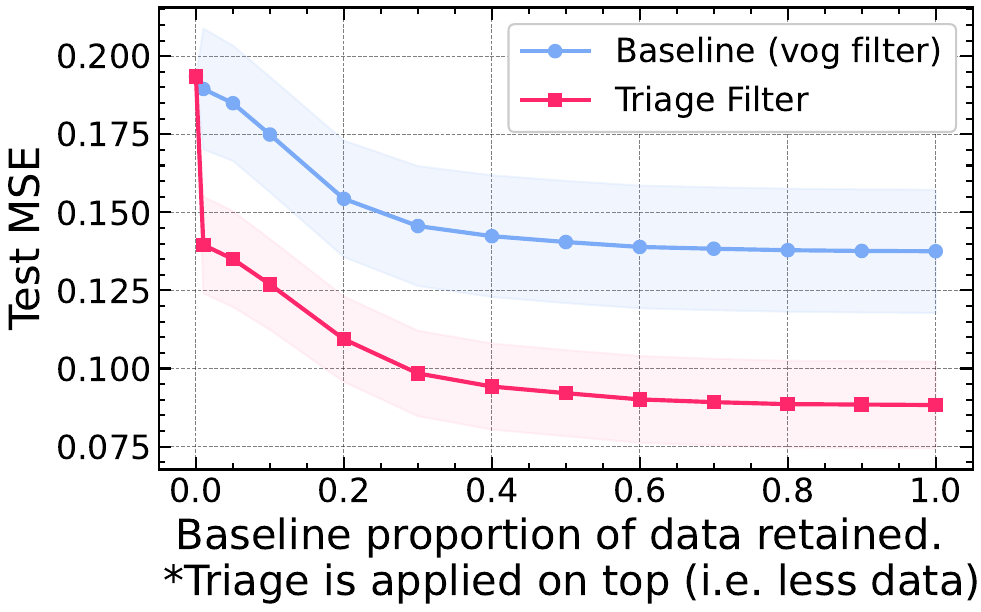}}\quad
  \vspace{-4mm}
  \caption{\footnotesize TRIAGE improves test MSE, as a fine-grained filter. For each baseline proportion retained, TRIAGE is applied on top --- such that TRIAGE has less data ($D_{Triage} \subset D_{Baseline}$), only keeping well-estimated samples for each proportion $p$.}
   \label{fig:filter_exp_main}
\vspace{-3mm}
    \rule{\linewidth}{.45pt}
\vspace{-9mm}
\end{wrapfigure}

In the previous section, we showed TRIAGE's ability to differentiate samples with more precision. Does this differentiation with TRIAGE scores benefit model performance? To find out, we sort the training samples from least to most challenging based on the baseline computed scores: Residuals/Errors, Loss, VoG, GradN. We then evaluate test MSE performance, as we increase the proportion $p$ of samples retained for training. i.e. increasingly retain more challenging samples with higher loss/VoG etc. As expected, the baselines (shown in Fig \ref{fig:filter_exp_main}, blue) have lower MSE with more samples retained for training. However, the elbow around $p=0.5$, after which MSE does not decrease, shows that the added high-magnitude samples are not beneficial.

We now assess the benefit of differentiating these scores using TRIAGE. To do so, we apply TRIAGE as an additional filter on top of the baseline retained samples --- \emph{only keeping the well-estimated subset of samples}. i.e. at each baseline proportion $p$, TRIAGE \textbf{reduces the number of samples retained} compared to the baseline s.t. $D_{Triage} \subset D_{Baseline}$. Fig \ref{fig:filter_exp_main} shows, for all proportions $p$, we improve MSE performance over baselines simply by differentiating between samples using TRIAGE and filtering to keep only well-estimated samples. This performance benefit holds for all 4 \emph{general-purpose} scoring methods, illustrating the value of TRIAGE as a fine-grained filter, where even with less data than baselines, the careful selection of samples by TRIAGE leads to improved regression performance. Results on more datasets mirroring the above are in Appendix \ref{filter-additional}.

\textbf{\textcolor{ForestGreen}{Takeaway 2.}} TRIAGE's differentiation of samples can be used to augment other approaches as a fine-grained filter; improving regressor test performance, simply by keeping well-estimated samples.

\subsubsection{Data sculpting to fit a deployment purpose.}\label{p3-exp}

In many situations, we might want to take a regressor trained on a large $\Dtrain$ and deploy it in an area with limited labeled data. For instance, repurposing a regressor in a low-resource healthcare setting. This raises the question, instead of acquiring new data -  which in some cases might be \emph{impossible} -  could we sculpt the existing large $\Dtrain$ to be more ``fit for purpose''. The goal is to enable better regressor performance. Our setup assumes limited access to a small amount of data from the deployment environment, which is \emph{insufficient} in sample size to directly train a highly performant regression model, yet we could use this data for calibration $\Dcal$ purposes. Can TRIAGE help?

As a case-study, we consider a regression task similar to \cite{stamey1989prostate,hastie2009elements,li2021linear}, using patient covariates to predict PSA blood levels, a known predictor of prostate cancer mortality. This task has utility in low-resource areas with limited medical funds \cite{li2021linear}.
We use multi-country prostate cancer data in which their is a distribution shift and train a regressor on data from the US (SEER)\cite{duggan2016surveillance}. We evaluate performance on data from the UK (CUTRACT)\cite{prostate}, where the smaller $\Dcal$ is also drawn from. In addition to the normal train-large, test-small baseline, we evaluate the utility of sculpting $\Dtrain$ from the US using TRIAGE --- \emph{only keeping well-estimated samples} ($D_{WE}$=$D_{Triage}$), s.t $D_{Triage}\subset\Dtrain$.  Conformal calibration uses $\Dcal$ from the UK.

We compare TRIAGE to several methods, as detailed in Appendix \ref{extra-details}, including: (i) \emph{data-driven} training: training a model directly on the $\Dcal$ or combining the two datasets ($\Dtrain \cup \Dcal$). (ii) \emph{prediction-based} sculpting: train on $\Dcal$ (UK data) and then subsequently filter $\Dtrain$ (US data) based on predictive uncertainty scores \footnote{Model-driven methods are fit on $\Dcal$, such that we assess how uncertain samples in $\Dtrain$ are with respect to $\Dcal$}. We assess NGBoost, Bayesian Neural Network, Gaussian Process, Bayesian ridge regression, conformal intervals and residuals/errors. We only consider general 
purpose methods whose criteria allow us to leverage $\Dcal$ and are compatible beyond neural nets for flexibility.

We then evaluate MSE test performance of regressors trained on the different ``sculpted'' $\Dtrain$. We test sensitivity to varying sizes of $\Dcal$, as in many settings, we might not have access to many calibration samples.  Table \ref{fit_for_purpose} shows that TRIAGE consistently outperforms \emph{model-driven} methods across the board. This suggests that predictive uncertainty scores are insufficient to select the samples with the highest utility in $\Dtrain$. This contrast implies TRIAGE's scoring method better selects samples in $\Dtrain$ with greater utility, leading to improved performance on $\Dtest$. 

Compared to \emph{data-driven} approaches, we find that \textbf{TRIAGE has the best MSE}, specifically, in the \emph{small sample regime}, where $\Dcal$ only contains a limited number of samples $n$=10-200. Of course, when sufficient data is available ($\ge$300 samples), sculpting has a reduced benefit. Rather, training directly on the sufficient $\Dcal$ is preferable. The result highlights the viability of sculpting a larger $\Dtrain$ with respect to a small $\Dcal$ to improve regression in settings with limited data access --- with TRIAGE offering the best performance benefits. 

In addition, sculpting especially in healthcare settings should account for both (i) imbalanced feature subgroups and (ii) long-tailed outcomes. We assess this in Appendix \ref{fit_for_purpose_more} and show that TRIAGE does retains strong performance on minority subgroups and long-tailed outcomes by virtue of the calibration dataset.

\begingroup

\setlength{\tabcolsep}{1.5pt} 
\renewcommand{\arraystretch}{1} 
\begin{table*}[!t]
\vspace{-0mm}
 \centering
 \captionsetup{font=footnotesize}
\caption{Comparison of test MSE for different approaches, changing the size of $\Dcal$. TRIAGE's approach to data sculpting of the large $\Dtrain$ with respect to a small $\Dcal$ in shown to outperform alternatives with lower test MSE, especially when $\Dcal$ from the target domain has small sample sizes. ($\downarrow$ better).  Note $|\Dtrain|\gg|\Dcal|$}.
\vspace{-2mm}
\scalebox{0.51}{
\begin{tabular}{l|l|cccccccccc}

\hline

& $\Dcal$ sample size & 10 & 20 & 30 & 40 & 50 & 100 & 200 & 300 & 400 & 500 \\ \hline\hline
 \multirow{2}{*}{\makecell{Ours (Sculpting)}} & \cellcolor{green!15} \textbf{TRIAGE} ($WE$) & \cellcolor{green!15} \bf 0.051+-0.003 & \cellcolor{green!15} \bf 0.050+-0.003 & \cellcolor{green!15} \bf 0.047+-0.002 & \cellcolor{green!15} \bf 0.046+-0.002 & \cellcolor{green!15} \bf 0.046+-0.001 & \cellcolor{green!15} \bf 0.046+-0.002 & \cellcolor{green!15} \bf 0.046+-0.001 & \cellcolor{green!15} 0.045+-0.001 & \cellcolor{green!15} 0.045+-0.002 & \cellcolor{green!15} 0.046+-0.002  \\ 
& Not TRIAGE ($OE \cup UE$) & 0.088+-0.014 & 0.068+-0.005 & 0.066+-0.003 & 0.066+-0.005 & 0.068+-0.006 & 0.080+-0.011 & 0.087+-0.015 & 0.093+-0.015 & 0.082+-0.012 & 0.078+-0.009  \\ \hline\hline
 \multirow{3}{*}{\makecell{Data-driven}} & Baseline ($\Dtrain$) & 0.064+-0.002 & 0.064+-0.002 & 0.064+-0.002 & 0.064+-0.002 & 0.064+-0.002 & 0.064+-0.002 & 0.064+-0.002 & 0.064+-0.002 & 0.064+-0.002 & 0.064+-0.002  \\
& Baseline ($\Dcal$) & 0.092+-0.023 & 0.070+-0.012 & 0.070+-0.010 & 0.064+-0.003 & 0.066+-0.009 & 0.052+-0.004 & 0.047+-0.001 & \bf 0.043+-0.001 & \bf 0.043+-0.001 & \bf 0.041+-0.001  \\
& Baseline ($\Dtrain \cup \Dcal$) & 0.060+-0.002 & 0.059+-0.002 & 0.058+-0.002 & 0.056+-0.002 & 0.055+-0.001 & 0.049+-0.001 & 0.047+-0.001 & 0.044+-0.001 & 0.044+-0.001 & 0.042+-0.001 \\  \hline\hline
 \multirow{6}{*}{\makecell{Prediction based \\ sculpting of $\Dtrain$}} & Error Sculpt & 0.066+-0.010 & 0.058+-0.010 & 0.060+-0.010 & 0.060+-0.010 & 0.057+-0.010 & 0.058+-0.010 & 0.055+-0.011 & 0.055+-0.011 & 0.057+-0.010 & 0.057+-0.010 \\
& CP Intervals Sculpt & 0.064+-0.006 & 0.081+-0.023 & 0.082+-0.008 & 0.106+-0.043 & 0.073+-0.013 & 0.059+-0.005 & 0.055+-0.002 & 0.063+-0.011 & 0.054+-0.002 & 0.051+-0.003  \\
& NGBoost & 0.066+-0.006 & 0.121+-0.041 & 0.085+-0.015 & 0.080+-0.008 & 0.094+-0.008 & 0.196+-0.030 & 0.132+-0.010 & 0.096+-0.011 & 0.095+-0.006 & 0.103+-0.008  \\
& Bayesian ridge & 0.080+-0.008 & 0.118+-0.023 & 0.111+-0.012 & 0.129+-0.011 & 0.129+-0.016 & 0.111+-0.017 & 0.106+-0.006 & 0.110+-0.010 & 0.114+-0.009 & 0.103+-0.009  \\
& BNN & 0.068+-0.005 & 0.063+-0.006 & 0.066+-0.004 & 0.066+-0.004 & 0.064+-0.005 & 0.057+-0.006 & 0.072+-0.010 & 0.056+-0.006 & 0.064+-0.003 & 0.068+-0.008  \\
& GP & 0.051+-0.006 & 0.066+-0.008 & 0.069+-0.005 & 0.077+-0.008 & 0.072+-0.006 & 0.071+-0.009 & 0.085+-0.011 & 0.085+-0.003 & 0.095+-0.004 & 0.089+-0.006  \\ \hline\hline

\end{tabular}}
\label{fit_for_purpose}
\vspace{-3mm}
\end{table*}
\endgroup

\textbf{Data insights from sculpting.}  We seek to understand what types of samples are ``sculpted''. Such insights are especially useful in clinical settings. We see that US patients ($\Dtrain$) typically have \emph{higher} cancer stages and \emph{higher} PSA scores than their UK counterparts ($\Dcal/\Dtest$). It is precisely these dichotomous samples that are filtered by TRIAGE to improve predictive performance. These insights can be illustrated to stakeholders via a radial diagram, as in Appendix \ref{insights_exp}.

\vspace{-0mm}

Additionally, given the potential data shift (around exchangeability), Appendix \ref{validity}  analyzes CPS calibration, along with continuous ranked probability score (CRPS) quantifying the quality of the predictive distribution.

\vspace{-2mm}
\paragraph{Beyond predictive performance to fairness.} We also show that sculpting the data has potential beyond simply improving predictive performance. In Appendix \ref{fairness-exp}, we illustrate how TRIAGE can be used to sculpt data to improve fairness of regressors based on access to a limited set of samples.

\textbf{\textcolor{ForestGreen}{Takeaway 3.}} TRIAGE can help to improve regression performance by sculpting a larger dataset to be fit-for-purpose, simply by leveraging a small number of calibration samples.

\vspace{-1mm}
\subsection{\textbf{(P3)} Informative for data collection}\label{p4-exp}
\vspace{-2mm}
We now demonstrate the value of TRIAGE for dataset selection and feature collection/acquisition.
\vspace{-2mm}
\subsubsection{Dataset selection: From sample-level to dataset level.}
\vspace{-1mm}
In real-world applications, we might want entire dataset characterization, rather than just sample-level characterization. An understudied scenario is how to compare and select between different datasets. This need arises in organizations where data is stored in isolated silos and it is difficult to access; with synthetic data a common solution \cite{el2020practical, goncalves2020generation} or alternatively when purchasing data via data marketplaces \cite{spiekermann2019data}. We study the former scenario, where we have multiple 'synthetic' data versions, produced by different ML models or data vendors. However, our assessment could easily apply to the latter of data marketplaces. 

We explore the setting where the real $\Dtrain$ is \emph{not accessible} to calculate statistical similarity measures, yet we still wish to compare synthetic datasets and select a version to train on. The underlying assumption is that the synthetic datasets reflect the true underlying distribution. However, depending on the generation process, some synthetic datasets might prove superior to others. 

The question we address is: \emph{Can TRIAGE's data characterization allow us to compare and select the synthetic training dataset that yields the best test performance on real data?}

\begin{wraptable}{r}{0.49\textwidth}
\vspace{-4mm}
 \centering
\captionsetup{font=footnotesize}
\caption{Comparing performance \& (\textcolor{ForestGreen}{Triage retention}). Training on synthetic data w/ better utility ($\uparrow$ retained) has the lowest real test data MAE across datasets. }
\vspace{-0mm}
\scalebox{0.8}{
\begin{tabular}{ccc}
\toprule
Dataset &  (V1) CTGAN  & (V2) TVAE    \\ \hline
\midrule

Bike & 0.13 (45\% Retained)   & \bf 0.09 (\textcolor{ForestGreen}{58\% Retained}):  \\ \hline

Bio & 0.24 (45\% Retained)   & \bf   0.21(\textcolor{ForestGreen}{52\% Retained}) \\ \hline

Boston &  0.16 (38\% Retained)   & \bf  0.13(\textcolor{ForestGreen}{60\% Retained})  \\ \hline

Concrete & 0.19 (46\% Retained)   & \bf  0.11 (\textcolor{ForestGreen}{48\% Retained}) \\ \hline

LOS &  0.16 (47\% Retained)   & \bf   0.09(\textcolor{ForestGreen}{56\% Retained}) \\ \hline

MIMIC & \textbf{0.05} (\textcolor{ForestGreen}{41\% Retained})   &  0.08 (\textbf{48\% Retained}) \\ \hline

Prostate & 0.21 (46\% Retained)   & \bf  0.18 (\textcolor{ForestGreen}{52\% Retained})  \\ \hline

Protein &  0.22 (52\% Retained)   & \bf  0.20 (\textcolor{ForestGreen}{54\% Retained})  \\ \hline

Star &  0.18  (38\% Retained)  & \bf  0.14 (\textcolor{ForestGreen}{51\% Retained})  \\ \hline

\bottomrule
\end{tabular}}
\label{tab:datasets_quality_synth}
\vspace{-3mm}
\end{wraptable}

We simulate this scenario by generating synthetic training data using (V1) CTGAN and (V2) TVAE \cite{xu2019modeling}, representing 2 synthetic data vendors.
We employ TRIAGE to analyze and compare them, based on the percentage of well-estimated examples recommended for retention. As is standard, we train a regressor on synthetic data and test it on real data (TSTR) \cite{esteban2017real,platzer2021holdout}. The best synthetic data should yield the lowest MAE on real test data.
Table \ref{tab:datasets_quality_synth} shows that where TRIAGE's recommendation has a \emph{higher retention percentage} (i.e. more well-estimated samples), that training with that dataset, leads to lower MAE on real test data. This illustrates that even if the real data is unavailable, TRIAGE's characterization can be used by practitioners to select the synthetic training dataset that will give the best real test performance.

\textbf{\textcolor{ForestGreen}{Takeaway 4.}} TRIAGE scores allow us to assess utility between different synthetic datasets in a principled manner. We show that synthetic datasets that TRIAGE suggests retaining \emph{more} samples (i.e. more well-estimated), allow for better generalization when testing on real-data (i.e. MAE).

\subsubsection{Value of feature collection/acquisition.}\label{p5-exp}
Previous settings considered assume a fixed dataset. What if we can actively improve the data's utility by acquiring features? e.g., a doctor runs extra  tests to improve a regressor. Understanding the value of features one might acquire is useful, especially if acquisition is costly.  We note feature acquisition differs from feature selection (e.g. \cite{knockoffgan}), where all features are present and we select useful ones. It also differs from active learning, which quantifies sample acquisition value, not features. 

\begin{wrapfigure}{r}{0.4\textwidth}
\captionsetup{font=footnotesize}
\vspace{-5mm}
  \centering
    \includegraphics[width=0.4\textwidth]{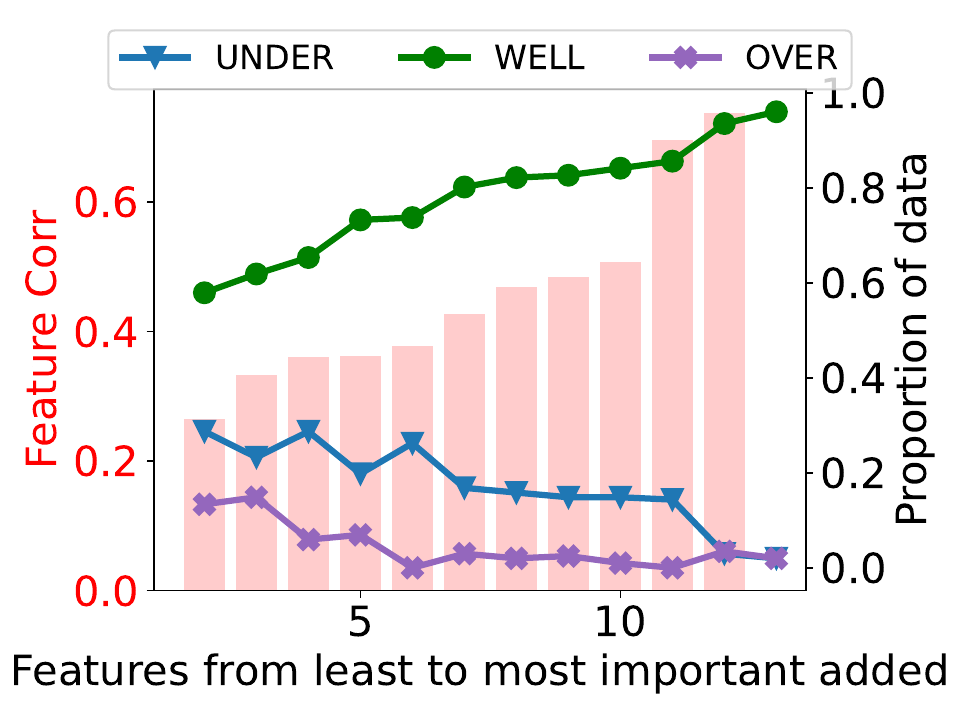}
   \vspace{-5mm}
   \caption{\footnotesize{The value of acquiring a feature can be quantified by its ability to increase the proportion of well-estimated samples.}}
    \vspace{-2mm}
    \rule{\linewidth}{.45pt}
   \vspace{-7mm}
    \label{fig:acquire}
\end{wrapfigure}

We now show that TRIAGE enables a principled approach to assess the benefit of acquiring a specific feature. We posit that acquiring a valuable feature wrt. the task should increase the proportion of well-estimated samples (which we've shown is linked to better performance). 

To illustrate the potential of TRIAGE scores, we set up an experiment where we rank the features based on their correlation to the target (i.e. increasing utility). We then simulate the sequential "acquisition" of features with increasing utility (based on correlation). Concurrently, we observe how the groups change as features are acquired. 

Fig.~\ref{fig:acquire} shows that 
as we acquire ``valuable'' features, the proportion of the well-estimated samples increases, with the over- and under-estimated proportions dropping.  This result, repeated on other datasets in Appendix \ref{appx:acquire-extra}, shows TRIAGE scores are sensitive to the value of a feature, quantified by the increase in the proportion of well-estimated samples.

\textbf{\textcolor{ForestGreen}{Takeaway 5.}} TRIAGE offers a principled way to assess active dataset improvement when we acquire new features. The feature's value links to the ability to increase the well-estimated proportion.

\vspace{-2mm}
\section{Discussion} \label{sec:discussion}

We present TRIAGE, a regression-focused data characterization framework, an \emph{understudied} area in data-centric AI. TRIAGE offers a new scoring method to overcome the limitations of classification-centric scores. We show TRIAGE's characterization can unlock a variety of use cases to enhance regression, with minimal effort, beyond standard training on a dataset and exchangeability assumptions. We highlight that a key aspect of ensuring good performance of TRIAGE is careful construction of the calibration set --- which should be as representative as possible. Rather than automating or replacing a data scientist's role, TRIAGE as a ``data-centric AI'' tool aims to empower data scientists to perform the ``data'' work in a more principled manner.  

\textbf{Limitations and future opportunities}: (1) understanding the attributes contributing to characterization, (2) provide theoretical guarantees taking into account the challenging interaction between the CPD and Triage trajectory (for more see Appendix \ref{huber-appendix}),  (3) in high-stakes settings, to prevent harm we could leverage domain expertise, where a human-in-the-loop could audit the samples surfaced by TRIAGE, before sculpting.

\section*{Acknowledgements}
The authors are grateful to Fergus Imrie, Boris van Breugel, Paulius Rauba and the anonymous NeurIPS
reviewers for their useful comments \& feedback. The authors would also like to thank Richard Samworth, Marco Scutari, Qingyuan Zhao and Yao Zhang for their insightful discussion on robust statistics linked to TRIAGE. Nabeel Seedat is supported by the Cystic Fibrosis
Trust, Jonathan Crabbe by Aviva, Zhaozhi Qian by Cancer Research UK.


\clearpage
\bibliographystyle{unsrt}
\bibliography{main}

\clearpage
\onecolumn
\appendix

\addcontentsline{toc}{section}{Appendix}
\part{Appendix - TRIAGE: Characterizing and auditing training data for improved regression}
\mtcsetdepth{parttoc}{2} 
\parttoc
\newpage
\section{TRIAGE further details \& related work} \label{sec:appendixA}
\subsection{Extended related work}\label{related-extended}

This paper primarily engages with the literature on data characterization and contributes to the nascent area of data-centric AI. We provide specific details of related methods in both these areas below.

\textbf{Data characterization.} 
Existing scoring methods to characterize data examples are primarily for classification tasks.  In contrast, our method TRIAGE is to the best of our knowledge the first specifically designed method for regression.
As discussed in the main paper, prior classification focussed methods can be divided into two groups. We provide specific details of each below.

 \textbf{Methods tailored to classification}, these methods rely on aspects only found in classification settings. 
 \begin{itemize}
     \item Forgetting scores \cite{toneva2018empirical} and Split-Second Forgetting Scores \cite{mainicharacterizing} surface difficult and easy examples by analyzing the time when examples change from correct to incorrect discrete class
     \item AUM \cite{aum} identifies mislabeled examples based on logits
     \item  Data Maps \cite{datamap} and Data-IQ \cite{seedatdata} require probabilities of the true class label to distinguish easy, ambiguous, and hard examples. 
 \end{itemize}

\textbf{General purpose methods}, these methods were built for classification, but could be repurposed for regression.
 \begin{itemize}
 \item Variance of Gradients (VoG) ranks examples by difficulty \cite{vog}
 \item GradN uses gradient norms to identify "important examples" for pruning during training \cite{gradn}
 \item Loss values to identify different data subsets \cite{metadata}
 \item Residual (Error) analysis, a common statistical approach identifying examples with large model errors.

 \end{itemize}

We argue that methods tailored to classification are not directly related to our regression setting, as they rely on properties specific to classification that are not present in regression. General purpose methods (besides errors \& losses) based on their scoring method (e.g. gradients), are limited to differentiable methods like neural networks. Hence, these methods are often inapplicable for example in tabular settings (e.g. healthcare or finance), where practitioners often use iterative learning algorithms like XGBoost. In addition, the general scoring methods are not sufficiently fine-grained to distinguish between examples with the same scores. 

Finally, we highlight two specific differences of our approach to all the aforementioned data characterization approaches. 

Firstly, all these approaches look to characterize data purely to improve predictive performance. Whilst we also aim to improve predictive performance, we also highlight the value of characterizing data beyond predictive performance with regard to fairness (See Appendix \ref{fairness-exp}). 

Secondly, other methods assume that the model deployment environment is the exactly same as the training environment when characterizing the training data. In contrast, our approach allows for characterization of the training data in a more nuanced manner, with respect to a potential deployment environment with minimal access to a limited amount of data (see Sec \ref{p3-exp}).

\textbf{Data-Centric AI.}
Assessing data quality is a crucial to improve ML system performance, yet is often overlooked in favor of algorithmic development \cite{Sambasivan}. However, systematic characterization of data could offer guidelines for data enhancement as well as, enable ML system performance improvements \cite{chen2021data, jain2020overview}. While current approaches when quality is considered is often adhoc or artisinal \cite{Sambasivan,ng2021,chug2021statistical,seedat2022dc}, there has been a recent shift to giving data "center-stage" termed data-centric AI. The goal being to build ``systematic methods to evaluate, synthesize, clean and annotate the data used to train and test the AI model'' \cite{seedat2022dc,polyzotis2021can}. 

Our work on characterizing TRIAGE contributes to the best of our knowledge the first method to evaluate data for regression settings, providing an ML-aware data quality monitoring \cite{renggli2021data}.

\subsection{Adapting TRIAGE to any model}\label{adapt}

\subsubsection{Overview.}
TRIAGE's formulation allows us to use it with \emph{any} ML model trained in stages (i.e. iterative learning). Of course, neural networks are easy since we just make use of the model checkpoints. How can we adapt to gradient boosting decision trees (GBDTs) methods - such as XGBoost, LightGBM etc, and statistical methods such as linear regression? As a rule of thumb, we simply need a set of checkpoints through training in order to apply TRIAGE.

This property is incredibly important for practical utility, opening up avenues for more flexible selection of regressors. 
We look at how TRIAGE can be used with XGBoost and Linear models next.

\subsubsection{XGBoost/GBDTs}

\paragraph{XGB - Overview}
Methods such as GBDTs or XGBoost are widely used by practitioners due to their performant nature on tabular data (often outperforming neural networks). By iteratively combininb weak models, it has shown to have great success. Hence, having our method applicable for GBDTs in addition to neural networks adds to the broad utility.  

Before outlining why methods such as GBDT's fit the TRIAGE paradigm, we provide a brief overview of GBDTs.
Formally, given a dataset $\Dtrain$, GBDT iteratively constructs a model $F:X \rightarrow \mathbb{R}$ to minimize the empirical risk. At each iteration $e$ the model is updated as: 
\begin{equation}\label{eq:update}
F^{(e)}(\mathbf{}{x}) = F^{(e-1)}(\mathbf{x}) + \epsilon h^{(e)}(\mathbf{x}),
\end{equation}
where $F^{(e-1)}$ is a model constructed in the previous iteration, $h^{(e)}(\bm{x}) \in \mathcal{H}$ is a weak learner, and $\epsilon$ is the learning rate.  We can consider the model at every $F^{(e)}$ as a checkpoint.

\begin{figure}[!h]
    \centering
    \includegraphics[width=0.65\textwidth]{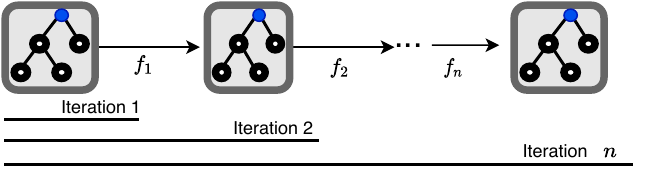}
    \caption{Example illustrating how TRIAGE can be adapted to XGBoost or Gradient Boosting methods by using a pseduo-ensemble. i.e. each sub-model is a checkpoint}
    \label{fig:xgboost_fig}
\end{figure}

\paragraph{Space \& Time Complexity.} We now provide guidance on how to apply TRIAGE to such methods. Naively, we could construct the checkpoints as an ensemble of multiple independent GBDT's. However, this is inefficient as the space and time complexity scales with $N$ models. To avoid increasing the overhead of a single model (from a practitioner perspective), we create a pseudo-ensemble using a single GBDT, see Figure \ref{fig:xgboost_fig}. 

\paragraph{Implementation.}
Similar to neural networks, the iterative nature of GBDTs means that the sequential submodels can be considered as checkpoints. Formally, each sub-model has parameters $\mathbf{\theta}^{(i)}$, hence the ensemble of checkpoints can be described as $\Theta = \{\mathbf{\theta}^{(i)}, 1 \le i \le N\}$. 

We then apply TRIAGE as normal to the checkpoints ($\theta_{1}, \theta_{2}...\theta_{E}$). The flexibility of this approach is that it applies both to training a new model, but interestingly, \textbf{we can also apply this to an ALREADY trained model by looping through the structure to create the pseudo-ensemble.}

\subsubsection{Linear models}

\paragraph{Overview.} Models like linear regression and ridge regression are commonly used statistical models widely used in regression settings. While they are often solved in closed form, we can also learn parameters iteratively. For example, using gradient descent.

\paragraph{Implementation.} Hence, to use methods like linear regression with TRIAGE, we assume we iteratively learn the parameters via something like gradient descent and store parameters $\theta$ at each iteration, i.e. checkpoint. We can then run TRIAGE as normal.
\clearpage

\subsection{Further details on Conformal Predictive Distributions}\label{cpd-extra}

This section provides additional details about Conformal Predictive Systems (CPS) \citep{vovk2019nonparametric, vovk2020computationally} that we use in TRIAGE.

Recall CPS combines work from parametric statistics of predictive distributions \citep{shen2018prediction,schweder2016confidence} with the method of conformal prediction \citep{vovk2005conformal}. It produces what are called Conformal Predictive Distributions (CPD), estimating the probability distribution of a continuous variable. 

We note that this predictive distribution could be transformed to prediction intervals with the corresponding quantiles. 

However, in TRIAGE we specifically study the predictive distribution to model the probability of the events related to the data's labels. We provide further details next.

\textbf{Split Conformal Predictive System (SCPS).}
We provide a formal definition of an SCPS. An SCPS is a function that is BOTH: (i) conformal transducer and (ii) randomized predictive system (RPS). 

\textit{(i) Conformal transducer:}
As per \cite{vovk2019nonparametric}, we thus define a function $Q$ as being a \emph{split conformal transducer}, if $Q$ is determined by a \emph{conformity measure} $\mu$ (the same as defined in the main text)., where $\mu$  measures the degree of agreement between the data set and the observation. Then for each possible label $y_i \in \R$ as part of $\mathcal{D}_{\mathrm{cal}} \in [q]$, we compute $q$ conformity scores $\alpha$. This is the same as described in the main paper.

\textit{(ii) RPS:}
The question is, how can a conformal transducer satisfy the properties of an RPS? For more details on this definition of an RPS we refer the reader to \cite{vovk2019nonparametric, vovk2020computationally}.

\paragraph{Motivating the choice of conformity measure.}
The implication that affects TRIAGE is that unlike traditional conformal prediction, where the choice of conformity measure does not matter, in the case of CPS not all conformity measures result in valid predictive distributions.  

As shown in \cite{vovk2019nonparametric}, in the context of SCPS, the conformity
measure $\mu$ must be a balanced isotonic function. See Definition \ref{def_isotonic}.

\begin{definition}\label{def_isotonic} 
A conformity measure $\mu$ is \emph{isotonic} - order preserving if, $y \leq y' \Rightarrow \mu(y) \leq \mu(y')$  
\end{definition}

We wish to satisfy this property and apply Proposition 3.1 from \cite{vovk2020computationally} that states: ``a split conformal transducer based on a
balanced isotonic split conformity measure is an RPS''.

Hence, this motivates our choice of conformity score, which we define as per Eq. \ref{ncs} in the main paper.

If $Q$ satisfies (1) a conformal transducer and (2) RPS, then we can relate $Q$ to a CDF of a given $y$, since $Q$ is monotonically increasing in $y \in \R$ and uniformly distributed on [0,1] \cite{vovk2019nonparametric}.

\textbf{\it While not critical to the performance of TRIAGE, we wish to highlight a nice property of validity that is provided in TRIAGE through the CPD.}

\paragraph{Remarks on validity.}\label{app:conf_guarantees}

The validity of the CPD is guaranteed if the data is exchangeable between $\Dcal$ and $\Dtest$. By validity, this refers to well-calibrated probabilities. (see Assumption \ref{exchangeability}). This means that we aren't required to impose any additional requirements for the validity of the CPD, since the aforementioned assumptions on the underlying data are typically made for any ML model.

We further examine the empirical effects on validity under a variety of settings in Appendix \ref{validity}.

\begin{assumption}[Exchangeability] 
In a dataset of $n$ observations, the data points do not follow any particular order, i.e., all $n$ permutations are equiprobable. Exchangeability is weaker than IID observations; however, IID observations satisfy exchangeability.
\label{exchangeability}
\end{assumption}

\subsection{Comparison of Triage to vanilla Conformal Prediction for regression}

We wish to highlight some key similarities between Triage and the usage of conventional conformal prediction for regression, e.g. \cite{johansson2014regression, seedat2023improving}.

\textbf{Differences.}

\begin{itemize}
    \item Objective: TRIAGE performs data characterization, scoring samples on their impact on a regressor, enabling data-centric tasks like data sculpting, dataset selection and feature acquisition. In contrast, \cite{johansson2014regression} is conventional conformal prediction for predictive uncertainty estimation.

   \item  Algorithm: (1) TRIAGE uses conformal predictive distributions (CPDs) providing a full predictive distribution. This contrasts conformal regression's prediction intervals, which are less informative than a predictive distribution. (2) TRIAGE’s novelty is studying the training dynamics of scores. In contrast, conformal regression computes prediction intervals once after training, not reflecting dynamic changes, which we show are vital to characterize differences between samples.

   \item  Experiments: TRIAGE tackles data-centric tasks like data sculpting (Sec 5.2), dataset selection (Sec 5.3.1) and feature collection/acquisition (Sec 5.3.2). In contrast, conformal regression like in \cite{johansson2014regression} evaluates the prediction intervals for predictive uncertainty (coverage and efficiency).

   \item  Results: While conformal regression tackles different tasks from TRIAGE, we adapt the CP intervals for sculpting (Sec 5.2.2). Table 2's baseline "CP Intervals Sculpt" corresponds to a conformal regressor.  Table 2 shows that porting CP intervals for sculpting are less effective than TRIAGE tailored for this task.

\end{itemize}

\textbf{Similarities:}  Similarities are on conformity score and calibration sets usage are general elements fundamental to conformal prediction itself and not unique to any method. This is akin to loss functions (conformity scores) and validation sets (calibration sets).

\subsection{TRIAGE training dynamics}\label{training_dynamics}

One might wonder why we need to evaluate training dynamics and how it can help us to differentiate between samples.

Of course, at the end of training samples can converge to the same score --- however samples are learned differently and at different rates during training. For example: one sample might converge to a CPS probability of 0.9 quickly during training, while another might take longer and oscillate more. Clearly, these samples are not the same.

\begin{figure}[!h]

  \centering
 \subfigure[Over-estimated]{\includegraphics[width=0.3\textwidth]{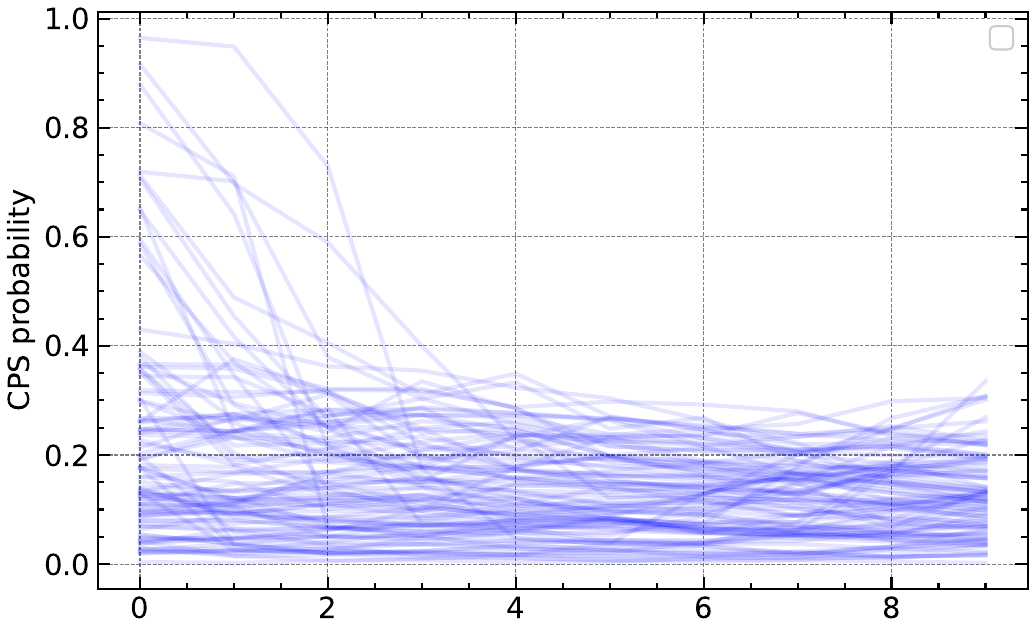}}\quad
  \subfigure[Well-estimated]{\includegraphics[width=0.3\textwidth]{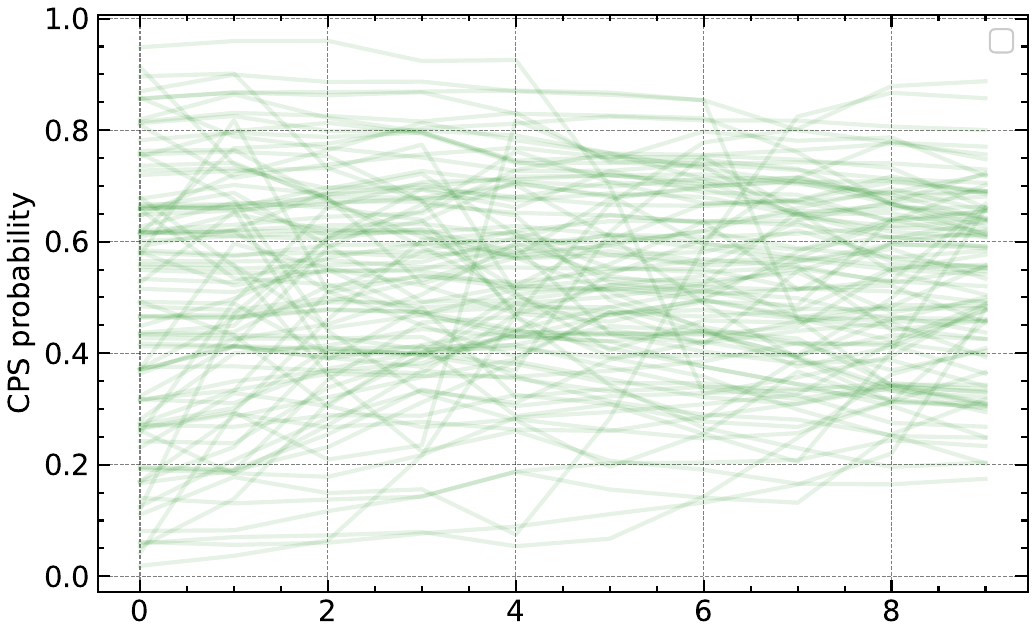}}\quad
  \subfigure[Under-estimated]{\includegraphics[width=0.3\textwidth]{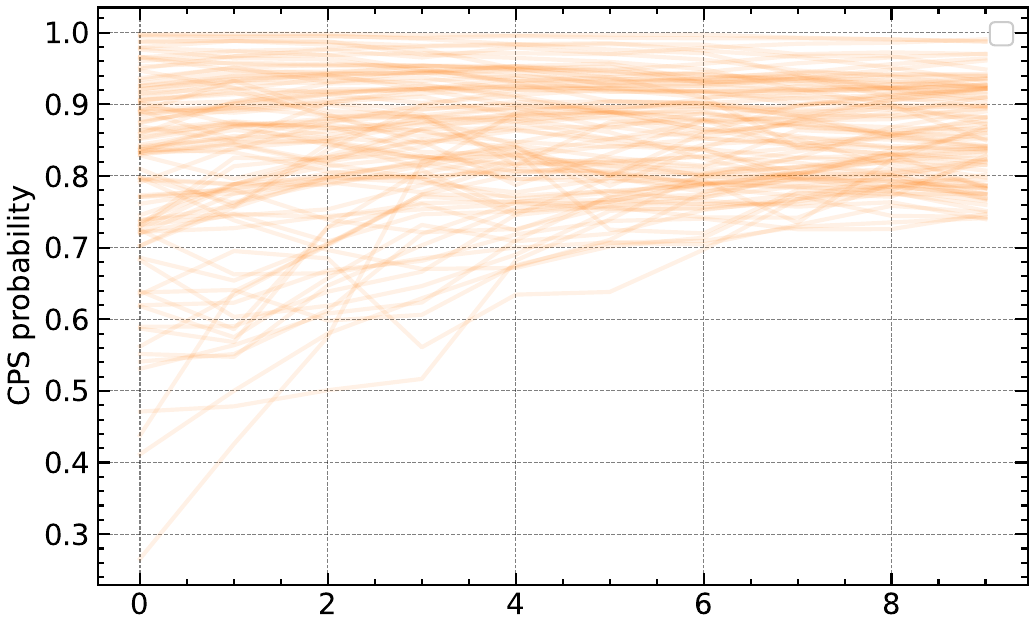}}\quad

  \caption{Training dynamics of the different groups -- clearly illustrates the necessity and value of training dynamics in capturing the differences between samples. The oscillations also help motivate our two metrics, confidence and variability.}
   \label{fig:training_groups}
    \rule{\linewidth}{.45pt}
\end{figure}

We show in Figure \ref{fig:training_groups} examples of training dynamics where we have already bucketed the samples as over-, under- and well-estimated. \textbf{Each line on the curve represents the training dynamics of a single sample.}

We can clearly see certain samples have much higher varying CPS probabilities over training, whilst others are way more stable.

This motivates our two metrics - Confidence and Variability. Where confidence helps to capture the average trend, but the variability is also important to delineate these oscillating samples.

\subsection{TRIAGE thresholds}
As outlined in the main text, we stratify samples in a threshold style applied to $C$ and $V$. In particular, the practitioner is required to set $\Cupper$ and $\Clower$. We deem this as a parameter the practitioner is free to set based on their tolerance levels. For example, someone might consider $>0.9$ to be confident, whereas another might consider $>0.7$. These could hence vary from application to application and across problem domains. We take a middle ground such that $\Cupper=0.75$ and $\Clower=0.25$. 

We however, provide a practical method to guide users in how they might select a threshold for $\Cupper$ and $\Clower$. We define a threshold $thresh \in \{0,0.5\}$ such that $\Cupper=1-thresh$ and $\Clower=thresh$. 

Now assume, for any dataset, we train a model and apply TRIAGE. We can then sweep $thresh \in \{0,0.5\}$ and assess the proportion of examples of well-estimated examples.

\begin{figure}[!h]
    \centering
    \includegraphics[width=0.5\textwidth]{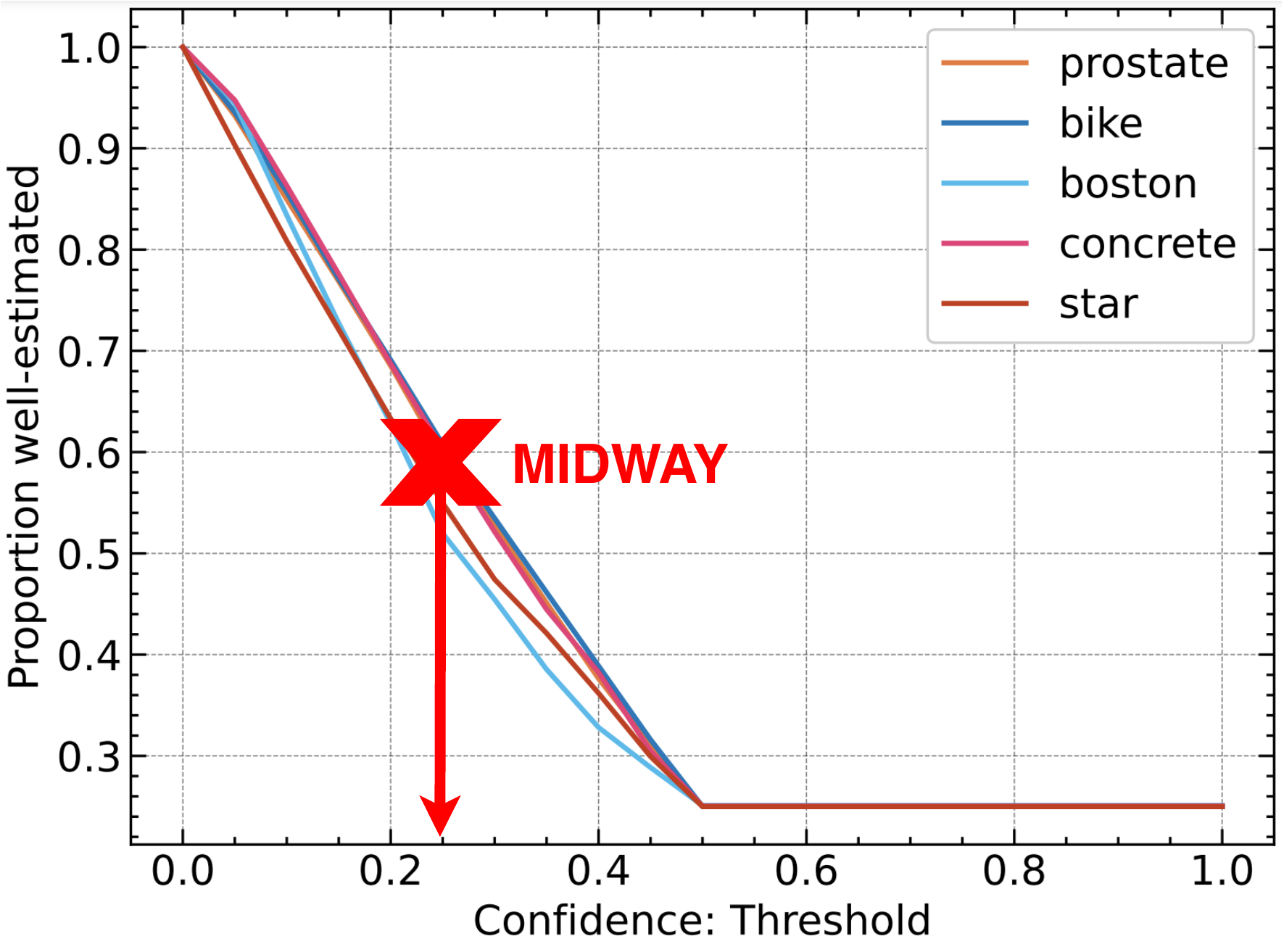}
    \caption{Threshold curve}
    \label{fig:thresh}
\end{figure}

We show results in Figure \ref{fig:thresh} below as we sweep the threshold. For low threshold values (e.g. thresh=0.1, $\Cupper=0.9$ and $\Clower=0.1$, where we want to be very certain to filter values. This results in a high proportion of well-estimated examples, which then decreases as the threshold increases.

We observe that our heuristic $\Cupper=0.75$ and $\Clower=0.25$, matches the midpoint of this curve, highlighting that this is indeed a reasonable choice

\newpage

\subsection{TRIAGE subgroup characterization}

With TRIAGE, we can obtain a characteristic curve as a summary of the data where the x-axis represents variability and the y-axis confidence --- both metrics computed wrt the TRIAGE score (i.e. CPS probability).  We now illustrate where example's lie on the plot.

\begin{figure}[!h]
    \centering
    \vspace{-3mm}
    \includegraphics[width=0.5\textwidth]{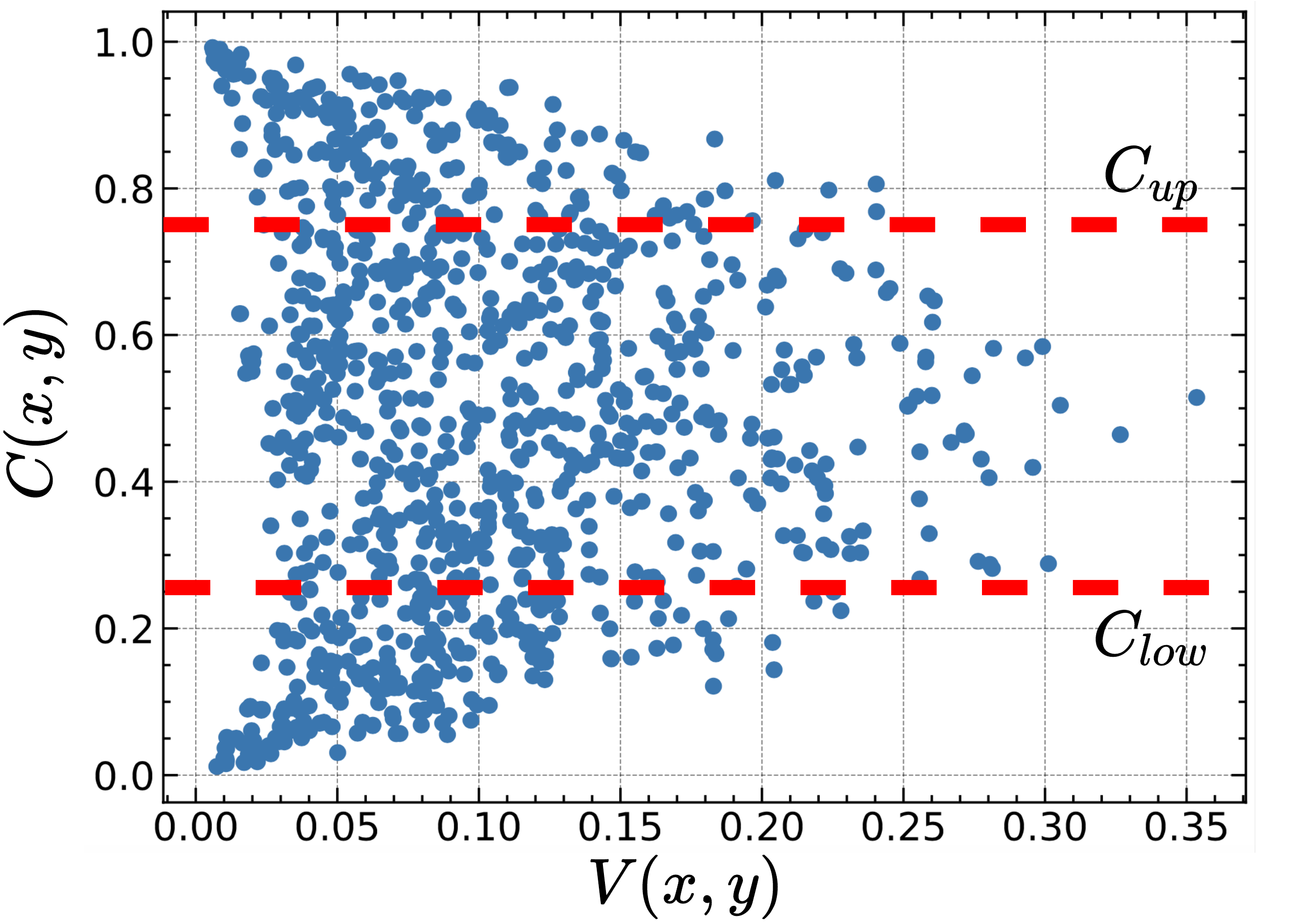}
    \caption{TRIAGE characteristic curve}
    \label{fig:triage_shape}
\end{figure}

We highlight in Figure \ref{fig:triage_shape}, that well-estimated samples are in the middle as they oscillate around 0.5.  

We also wish to highlight two types of samples that we DO NOT find in practice.  Types of samples we DO NOT find:
\begin{itemize}
    \item  $C>>0.5$ - very high and also have very high variability $V$
    \item $C<<0.5$ - very low and also have very high variability $V$
\end{itemize}
This is likely due to the nature of training dynamics, where samples with very high and very low CPS probabilities are quite clear-cut and hence do not oscillate much over training.

Such phenomena that we observe in practice motivate our rules, which are used to assign samples to the three categories.

\newpage

\newpage
\section{Benchmarks \& Experimental Details} \label{sec:appendixB}
\subsection{Benchmarks}\label{appendix:benchmarks}

\subsubsection{TRIAGE}
The description of TRIAGE is detailed in the main paper, where we compute Conformal Predictive Distributions at each checkpoint $E$ for each training example. We then study the evolution of the CPD through training.

\paragraph{Implementation details.} We implement TRIAGE in a versatile manner such that it can plug into regressors built either with Pytorch or Scikit-learn.  Our rationale for supporting both types of model API's is that since TRIAGE is applicable to any regressor which we can train iteratively, our functionality should be equally flexible from a usability perspective to foster easy usage and integration.

\paragraph{Code.} We will release the code upon acceptance such that TRIAGE can be used by practitioners and researchers alike.

\subsubsection{General-purpose scoring methods}

\paragraph{GraNd}

The gradient norm at epoch $t$ for an input $x,y$ is computed as :\\
$\chi_{t}(x, y)=\mathbb{E}\left\|\sum_{k=1}^{K} \grad_{f^{(k)}} \ell\left(f_{t}(x), y\right)^{T} \psi_{t}^{(k)}(x)\right\|_{2}$\\
where $\psi_{t}^{(k)}(x)=\grad_{\mathbf{w}_{t}} f_{t}^{(k)}(x)$.

\textbf{Implementation details.}  The benchmark is based on \citep{paul2021deep}. 
We adapt the Jax implementation from \footnote{https://github.com/mansheej/data\_diet}
to Pytorch with the help of \footnote{https://github.com/cybertronai/autograd-lib}.




\paragraph{Losses}

Losses to assess training dynamics evaluate the loss at each checkpoint $E$. Such that the score is evaluated on the training trajectory:\\

$s_{i}^e = (l(x_i, y_i, \theta_{1}),l(x_i, y_i, \theta_{2}),\ldots,l(x_i, y_i, \theta_{E})| (x_i, y_i) \in \Dtrain)$

\textbf{Implementation details.}  The benchmark is based on \cite{metadata} and we use the implementation from \footnote{https://github.com/shoaibahmed/metadata\_archaeology}. 

\paragraph{VoG}

The Variance of Gradient (VoG) is computed across all $E$ checkpoints. At each checkpoint the gradient $G_{x}$ with respect to input $x$ is computed.

For a given set of $E$ checkpoints, we generate the gradients for each sample $\mathbf{S}$ for all individual checkpoints, i.e., $\{\mathbf{G}_{1}, \dots, \mathbf{S}_{E}\}$. We then calculate the mean gradient $\mu$ by taking the average of the $E$ gradients. The variance of gradients is then: \\

$   \text{VoG}_{x_i} = \sqrt {\frac{1}{E}\sum_{i=1}^{E}(\mathbf{S}_{i} -\mu)^{2}.}$

\textbf{Implementation details.}  The benchmark is based on \citep{vog} and we use the implementation from \footnote{https://github.com/chirag126/VOG}. 

\newpage
\subsection{Datasets}
We provide a summary of the different datasets we use in this paper in Table \ref{tab:dataset}. The datasets vary in number of samples, number of features and domain.

\begin{table}[h]
\centering
\caption{Summary of the datasets used in TRIAGE}
\scalebox{0.9}{
\begin{tabular}{llll}
\toprule

Name &  $n$ sample magnitude & $n$ features & Domain \\ 
\midrule
Bike \cite{uci} & 11k & 18 & General-UCI \\ 
Bio \cite{uci} & 10k & 9 & General-UCI \\ 
Boston Housing \cite{uci} & 500 & 13 & General-UCI \\ 
Concrete \cite{uci} & 1k & 8 & General-UCI \\ 
CUTRACT Prostate \cite{prostate} & 2k & 12 & Medical \\ 
Microsoft Length of Stay \cite{microsoft_2017} & 100k & 23 & Medical \\ 
MIMIC Antibiotics \cite{johnson2016mimic}  & 5k & 85 & Medical \\
Protein \cite{uci} & 46k & 9 & General-UCI \\ 
SEER Prostate \cite{duggan2016surveillance} & 20k & 12 & Medical \\ 
Star \cite{uci} & 20k & 12 & General-UCI \\ 
\bottomrule
\end{tabular}}
\vspace{.5cm}
\label{tab:dataset}
\end{table}

\subsection{Additional experiment details}\label{extra-details}
We note that all experiments were performed using a single Nvidia Tesla P100 GPU.

\subsubsection{Section 3.1. Robustness to variation experiment details.}
We assess the robustness to the variation of different scoring methods when using different models and/or parameterizations. 

We parameterize these models differently based on the following changes: (1) Model type, (2) number of layers, (3) number of hidden units, (4) type of optimizer used. We train the models to similar levels of performance for each variant.

We motivate these four changes as these are common changes to model architectures or parameterizations made by practitioners.

We then evaluate all different combinations of these changes over all datasets. We compute both the Spearman correlation for all combinations for each of the 10 datasets.

(1) Model type: Standard MLP and residual MLP (ResMLP)
(2) number of layers --- we evaluate a 3 layer, 4 layer and 5 layer MLP\\
(3) number of hidden units --- we either reduce the units per layer by 1/2 as we go from the input dimension or by 1/4.\\
(4) type of optimizer --- we assess both Adam and SGD.

Finally, through training, we compute the TRIAGE score and the general purpose scores and then compute the pairwise Spearman correlation of these metrics.

\subsubsection{Section 3.2. Fine-grained filter experiment details.}
We train a 5-layer MLP for every regression task. Running TRIAGE and the other general-purpose methods at the end of training to characterize the data (i.e. assign scores to the data). We then take all the general-purpose scores and rank sort them from lowest-highest. Which represents easiest to hardest samples. Thereafter, as a baseline for each method we sweep and retrain a model with an increasing proportion of data included from lowest to highest. As a fine-grained filter, we apply TRIAGE on top of these other methods and show we can improve performance, simply by virtue of differentiating samples. What is interesting is that at each proportion, the TRIAGE version contains less samples.

\subsubsection{Section 3.3. Fit for purpose experiment details.}
For this experiment we train a baseline XGBoost regressor model on SEER (US) - $\Dtrain$. The XGBoost implementation uses the Python version from \footnote{https://github.com/dmlc/xgboost}.

We then draw $\Dcal$ from CUTRACT (UK), as is $\Dtest$ from the UK but a disjoint set. We then assess the following performances evaluated on the test set wrt. MSE performance. Besides the sculpting and training directly on different subsets of the data, we assess the following model-driven approaches described next.

The following models are trained on $\Dcal$, they are then applied to $\Dtrain$, where $\Dtrain$ is then sculpted based on the predictive uncertainty. The following models are assessed: 
NGBoost \footnote{https://github.com/stanfordmlgroup/ngboost}, Bayesian Neural Network \footnote{https://github.com/IBM/UQ360}, Gaussian Process \footnote{https://scikit-learn.org/}, Bayesian Ridge Regression \footnote{https://scikit-learn.org/}.

\subsubsection{Section 3.4. Beyond example-level to dataset level experiment details.}
We compare two synthetic data models as representations of comparing two sources producing synthetic data (e.g. two companies producing synthetic data). We compare CTGAN and TVAE and use their implementations from 
\footnote{https://github.com/sdv-dev/SDV}.

\subsubsection{Section 3.5. Active improvement via acquisition experiment details.}
We assess the viability of using TRIAGE to quantify the value of acquiring features based on the increase in well-estimated samples. The value in the simulated setup is quantified by the Pearson correlation of the feature to the label. We rank sort these correlations from low to high and sequentially acquire them, running TRIAGE at each instantiation. Our underlying model in which to do the quantification is an XGBoost with 20 estimators.

\newpage
\section{Additional experiments} \label{sec:appendixC}
\subsection{Consistency across datasets}\label{consistency-indiv}

\begin{figure}[h]
    \centering
    \includegraphics[width=0.8\textwidth]{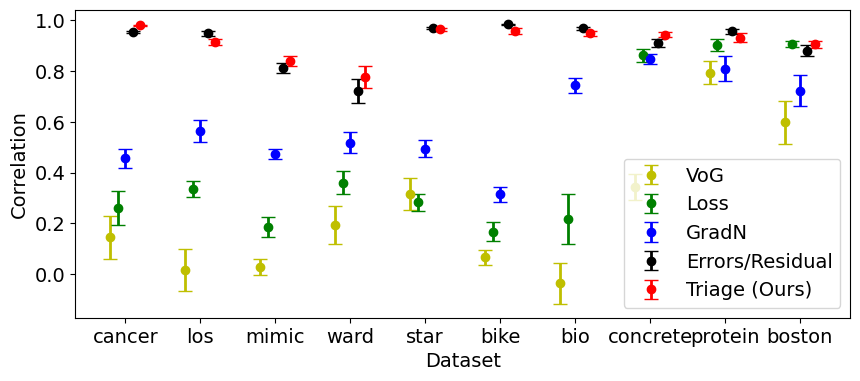}
    \caption{Consistency all datasets}
    \label{fig:consistency_indiv}
    \vspace{-5mm}
\end{figure}

\paragraph{Goal.} In the main paper we looked at consistency and stability of the different scoring methods when averaged across the different datasets. We now assess the individual correlations for each dataset in Figure \ref{fig:consistency_indiv}.

\paragraph{Takeaway.} TRIAGE has the highest consistency across all the different datasets compared to the other methods. We also see the baselines do not have 
consistent performance ordering across datasets, making it challenging to
select an appropriate scoring rule.

\subsection{Computation time}

\paragraph{Goal.} An important question is how does TRIAGE scale to the size of the dataset from a computation time persepctive. To assess this we vary the dataset size from 2000-100k samples and assess the TRIAGE computation time. 

\begin{figure}[!h]    
\centering
\includegraphics[width=0.4\textwidth]{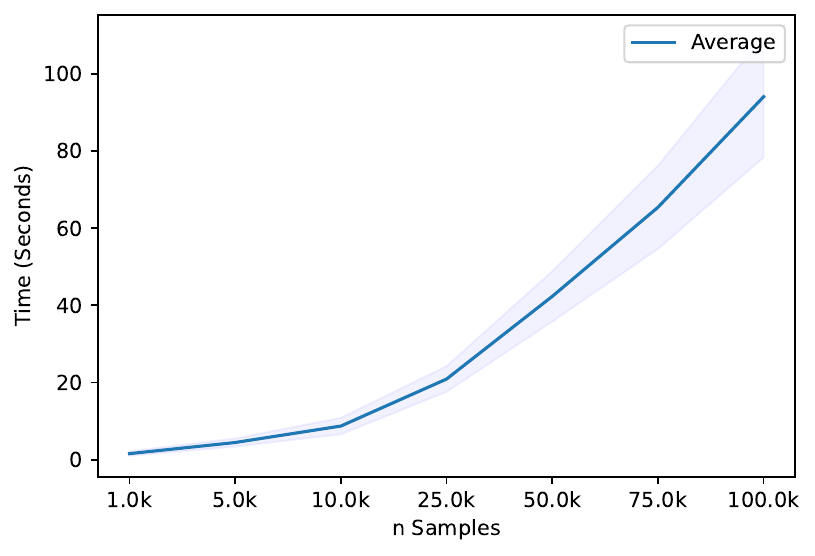}
     \caption{Overall computation time with data sizes varying between 2000-100k samples}
     \label{computation}
     \vspace{-5mm}
\end{figure}%

\paragraph{Takeaway.}  As shown in Figure \ref{computation}, naturally as the data size increases, so does computational time. However, even at 100k samples computing the TRIAGE scores for all 100k takes less than 2 min highlighting TRIAGE’s time efficiency to scale to larger data sizes. This suggests that TRIAGE can be used efficiently with larger datasets.

\subsection{TRIAGE w/ other iterative algorithms: XGBoost \& CatBoost}

\paragraph{Goal.} As discussed in the main paper, TRIAGE can be used with any regressor which can be trained iteratively.  Methods such XGBoost and CatBoost methods are widely used regressors by practitioners on tabular data, often more so than neural networks \cite{borisov2021deep}.  As mentioned, we desire consistency of data characterization, such that samples are consistently characterized across similar performing models irrespective of the model.

\begin{figure}[!h]

  \centering
 \subfigure[Prostate ]{\includegraphics[width=0.3\textwidth]{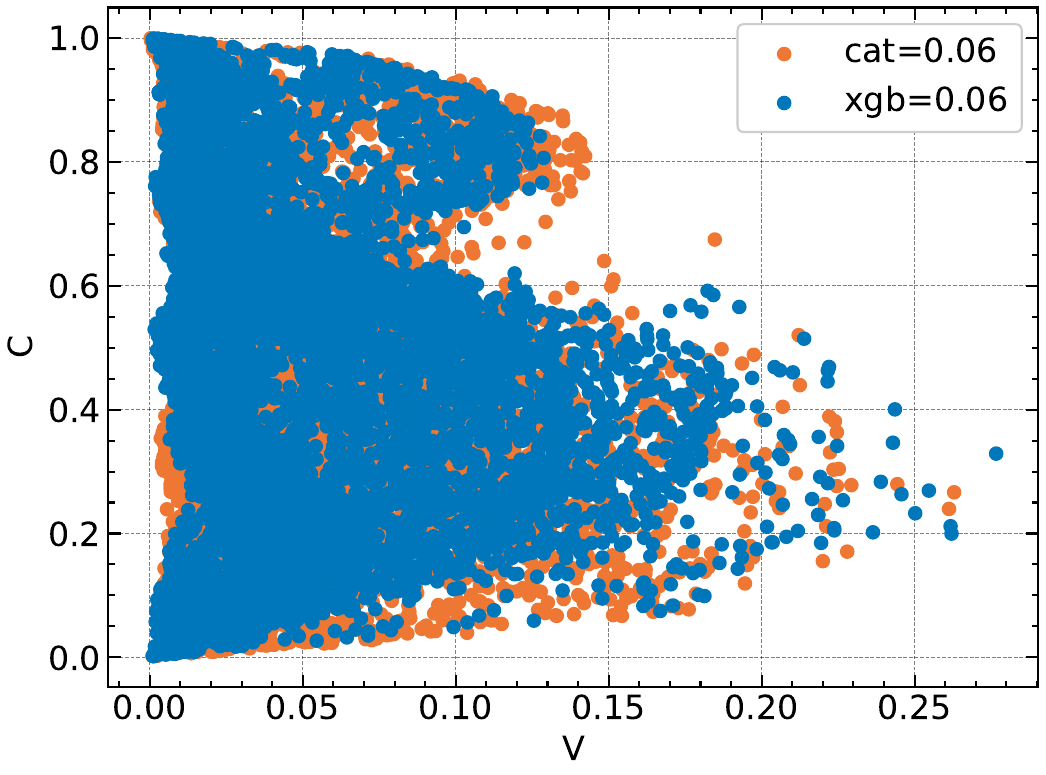}}\quad
  \subfigure[{Concrete} ]{\includegraphics[width=0.3\textwidth]{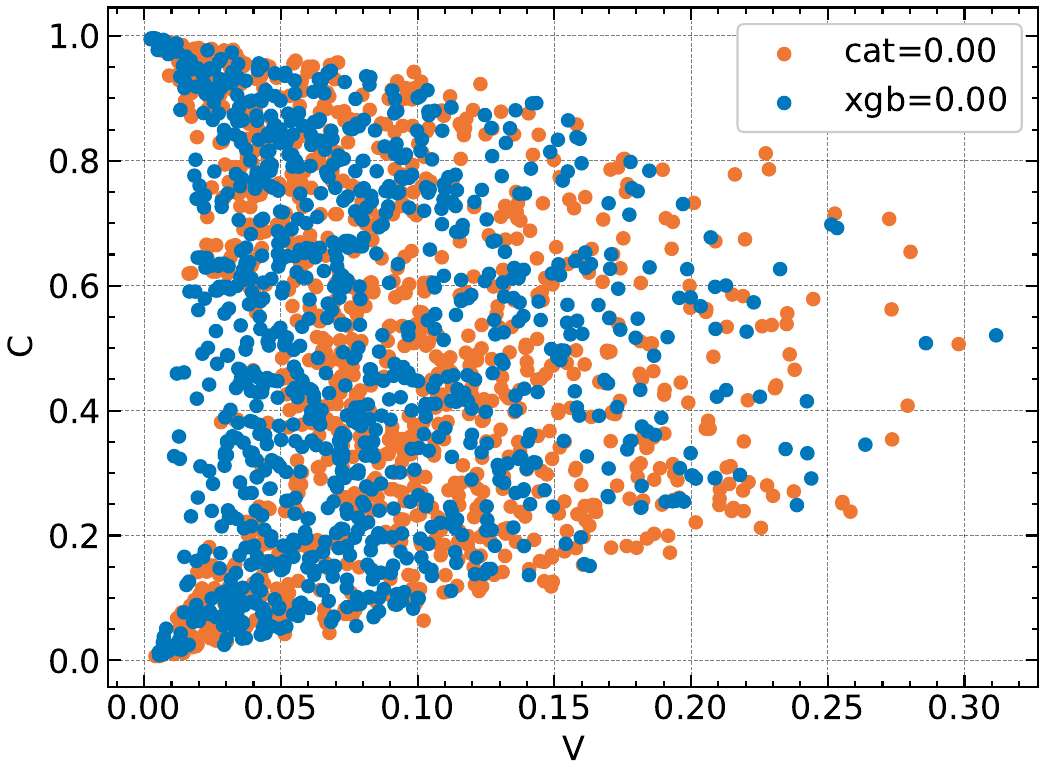}}\quad
  \subfigure[{Bio} ]{\includegraphics[width=0.3\textwidth]{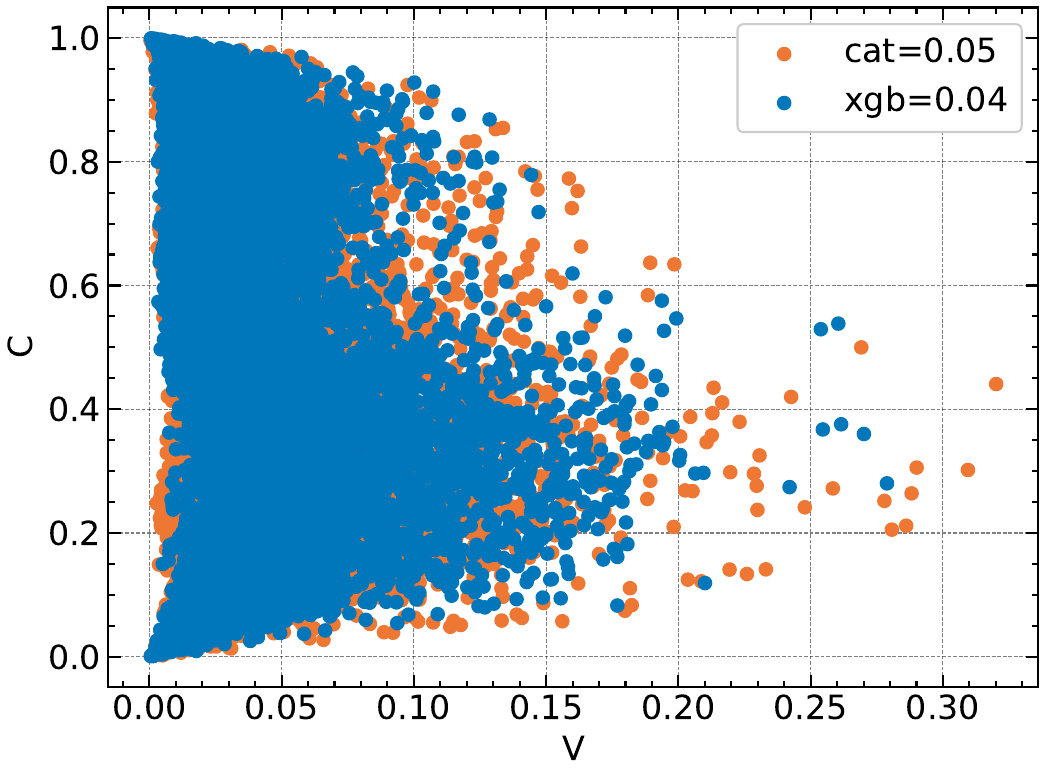}}\quad
  \subfigure[{Protein} ]{\includegraphics[width=0.3\textwidth]{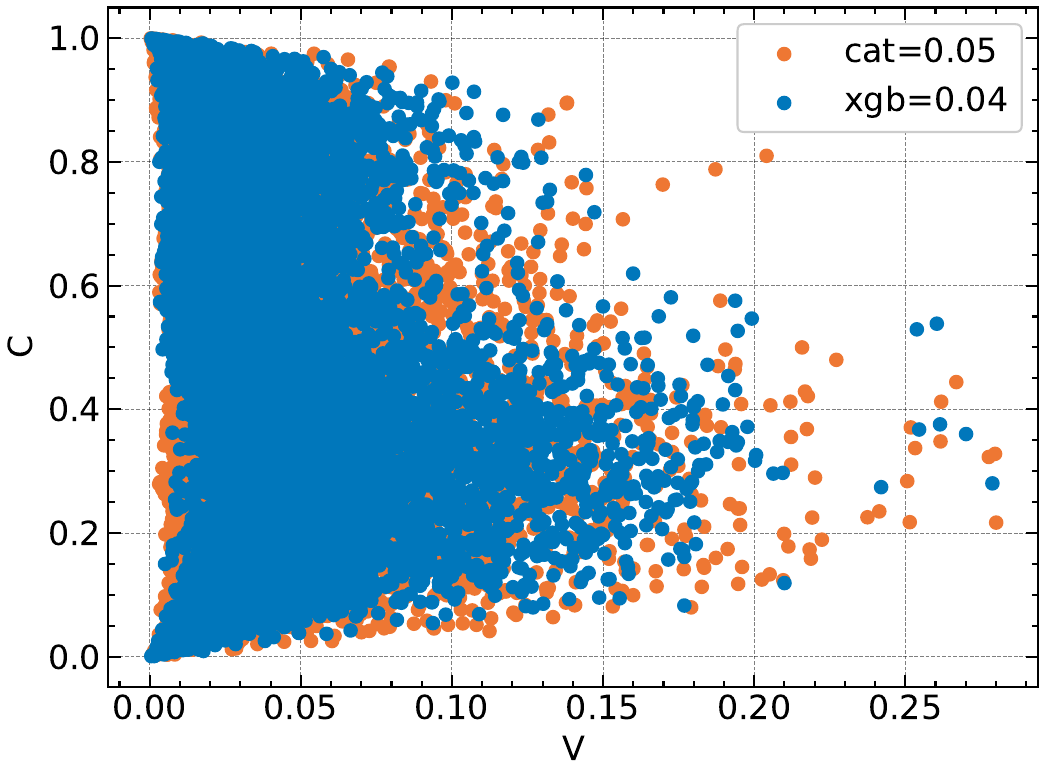}}\quad
  \subfigure[{Star} ]{\includegraphics[width=0.3\textwidth]{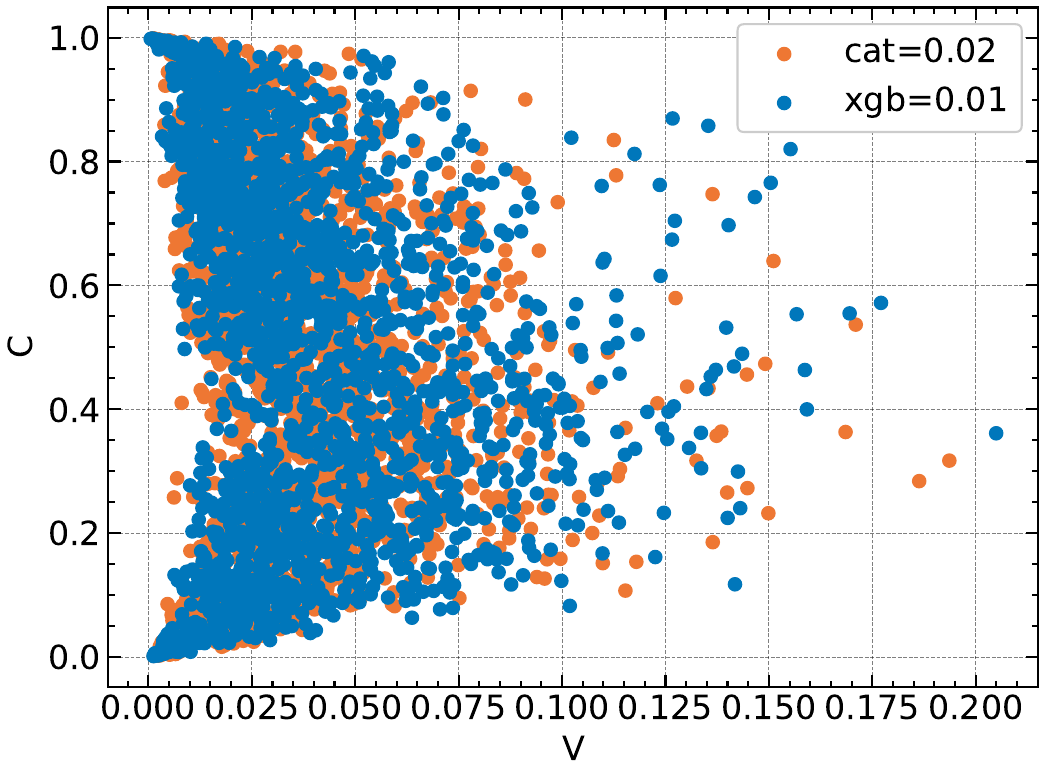}}\quad
  \subfigure[{Bike} ]{\includegraphics[width=0.3\textwidth]{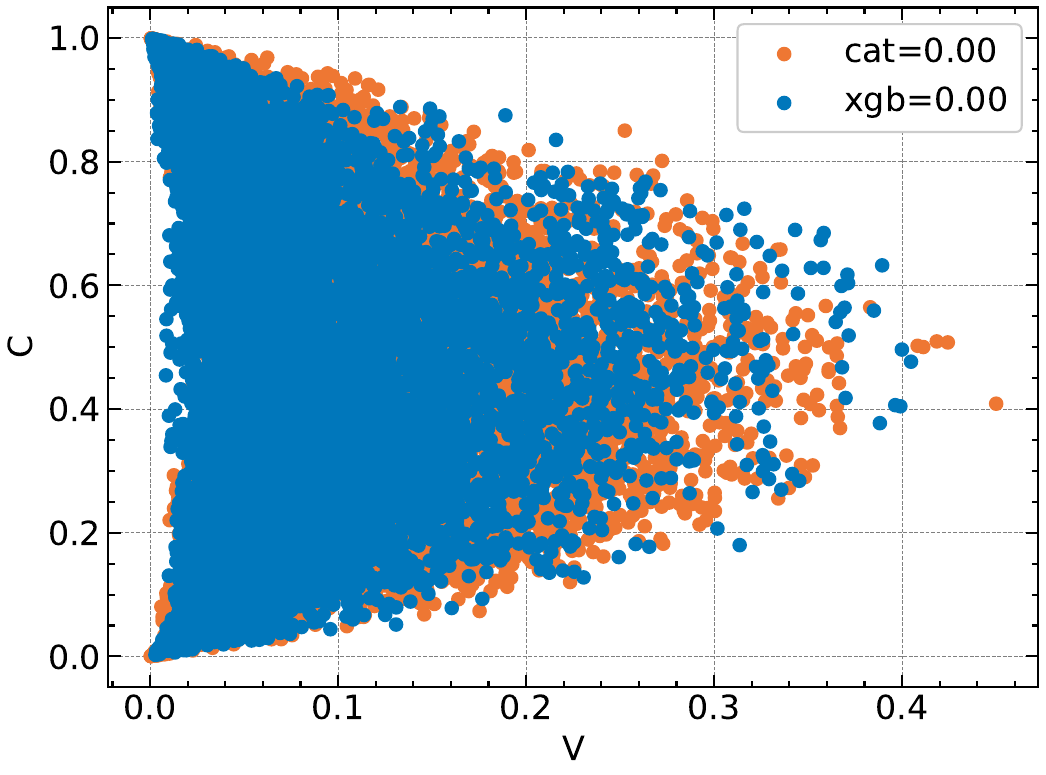}}\quad
  \subfigure[{Boston} ]{\includegraphics[width=0.3\textwidth]{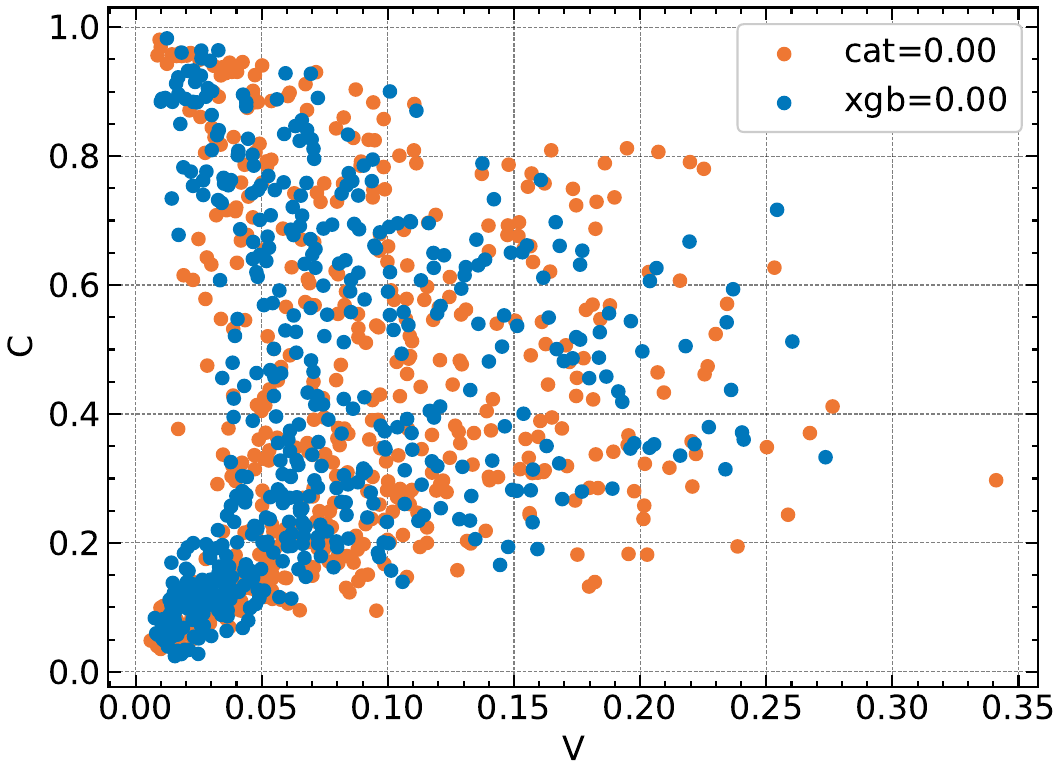}}\quad\\
  
  \caption{Assessing the consistency \& similarity of the TRIAGE characteristic curve using both XGBoost and CatBoost. The characterization by TRIAGE for both methods is similar.}
   \label{fig:gbt_stability}
    \rule{\linewidth}{.45pt}
\end{figure}

We have assessed the neural network consistency, but not assessed this across models such as XGBoost or CatBoost. We train both to achieve a similar level of performance and then apply TRIAGE as an assessment for these methods. The evaluation of both on different datasets is in Figure \ref{fig:gbt_stability}, where we show characteristic curves.

\textit{Note, the other baselines could not be easily assessed since the models under evaluation are not differentiable.}

\paragraph{Takeaway.} We can see TRIAGE characterizes samples consistently across the two models, as can be seen across the characteristic curves. We also compute the Spearman correlation of scores across models in Table \ref{tab:gbt_corr}. The mean Spearman correlation across datasets is \textbf{$0.88+-0.04$}.

\begin{table}[h]
\centering
\caption{Spearman correlation xgboost vs catboost}
\scalebox{0.9}{
\begin{tabular}{lll}
\toprule

Name &  Scores corr \\ 
\midrule
Bike \cite{uci} & 0.77  \\ 
Bio \cite{uci} & 0.94  \\ 
Boston Housing \cite{uci} & 0.8  \\ 
Concrete \cite{uci} & 0.83  \\ 
Prostate \cite{duggan2016surveillance} & 0.97  \\ 
Protein \cite{uci} & 0.94  \\ 
Star \cite{uci} & 0.97  \\ 
\bottomrule
\end{tabular}}
\vspace{.5cm}
\label{tab:gbt_corr}
\end{table}

\textbf{PTO - SEE NEXT PAGE FOR CONTINUATION}

\clearpage

\subsection{Fairness - extending sculpting beyond predictive performance} \label{fairness-exp}

\paragraph{Goal.}
Data sculpting can be used to not only improve predictive performance but also address issues of fairness. For instance, when a dataset contains biases that may harm certain groups at deployment, a smaller, more reflective dataset can guide the sculpting of the larger dataset, improving the fairness of the model trained on it.

\paragraph{Experiment.}
We demonstrate this using a similar calibration approach as in previous experiments, using the Communities and Crime dataset \cite{redmond2002data, uci}, where the regression task is predicting the ViolentCrimesPerPop variable (total number of violent crimes per 100K population) based on a set of demographic variables. Similar to \cite{steinberg2020fairness}, we identify race as a sensitive attribute, often a source of bias. The protected group is considered as those where more than 50\% of the community identifies as Black.

We compare TRIAGE based sculpting to a baseline XGBoost (trained without sculpting). We also use the same data-driven and model-driven baselines as the previous experiment. Additionally, we compare to a method that directly optimizes the model for fairness (as opposed to TRIAGE sculpting based on a set of examples) i.e. Fair Regression with Bounded Group Loss \cite{agarwal2019fair} \cite{agarwal2019fair} --- BGL Model. 

We evaluate the methods using three fairness metrics for regression from \citet{steinberg2020fairness}; estimated via Direct Density Ratio Estimation, i.e. \\
(i) \textit{Independence}: $S\bot A$; \\
(ii) \textit{Separation}: $S\bot A \mid Y$ \\
(iii)
\textit{Sufficiency}: $S\bot A \mid R$. \\ 
Where: A - protected group, Y - true target and S - model's prediction.  We use the common 80\% rule \cite{feldman2015certifying,equal1990uniform} and set $\epsilon=0.8$ as the threshold for fair assessment, comparing the ratio of metrics between protected and privileged groups, which should fall between ($\epsilon$, 1/$\epsilon$), i.e. (0.8, 1.25).

\paragraph{Results.}
We show that TRIAGE-based sculpting has potential beyond predictive tasks, as it can improve fairness metrics without sacrificing performance, as shown in Table \ref{fair_table}. Unlike other methods that require direct access to the protected attribute or direct model optimization, TRIAGE uses a set of calibration examples to sculpt the data. We compare TRIAGE to BGL models and Baseline $\Dcal$, which are the only methods to meet the criteria on all three fairness metrics, and show these methods often sacrifice predictive performance to achieve fairness. TRIAGE offers a flexible alternative to improve models for fairness simply based on access to a limited set of examples.

\begin{table}[H]
\vspace{-1mm}
 \centering
 \captionsetup{font=footnotesize}
\caption{Assessment of fairness metrics, highlighting the potential of data sculpting via TRIAGE}
\vspace{-0mm}
\scalebox{0.9}{
\begin{tabular}{lcccc}
\hline
& \multicolumn{3}{c}{\emph{Fairness}} & \\
\cmidrule(lr){2-4}
Method & Independence & Separation & Sufficiency  & MAE  \\ \midrule

TRIAGE & \cellcolor{green!10} 1.21 (\cmark)& \cellcolor{green!10} 1.11 (\cmark)& \cellcolor{green!10} 1.00 (\cmark) & 0.0528 \\ \hline

Baseline ($\Dtrain$) & \cellcolor{red!10} 1.29 (\xmark)& \cellcolor{green!10} 1.18 (\cmark)& \cellcolor{green!10} 1.01 (\cmark) & 0.0554 \\

Baseline ($\Dcal$) & \cellcolor{green!10} 1.14 (\cmark)& \cellcolor{green!10} 1.09 (\cmark)& \cellcolor{green!10} 0.99 (\cmark) & 0.0689 \\

Baseline ($\Dtrain \cup \Dcal$) & \cellcolor{red!10} 1.27 (\xmark)& \cellcolor{green!10} 1.13 (\cmark)& \cellcolor{green!10} 1.02 (\cmark) & 0.0467 \\ \hline

Error Sculpt & \cellcolor{red!10} 1.41 (\xmark)& \cellcolor{green!10} 1.19 (\cmark)& \cellcolor{green!10} 1.03 (\cmark) & 0.0502 \\

CP Intervals Sculpt & \cellcolor{red!10} 1.26 (\xmark)& \cellcolor{green!10} 1.15 (\cmark)& \cellcolor{green!10} 1.01 (\cmark) & 0.0467 \\

NGBoost sculpt & \cellcolor{red!10} 1.47 (\xmark)& \cellcolor{red!10} 1.26 (\xmark)& \cellcolor{green!10} 1.02 (\cmark) & 0.0594 \\

Bayesian ridge sculpt & \cellcolor{red!10} 1.39 (\xmark)& \cellcolor{green!10} 1.21 (\cmark)& \cellcolor{green!10} 1.00 (\cmark) & 0.0578 \\

BNN sculpt & \cellcolor{red!10} 1.27 (\xmark)& \cellcolor{green!10} 1.15 (\cmark)& \cellcolor{green!10} 1.00 (\cmark) & 0.0507 \\

GP sculpt & \cellcolor{red!10} 1.43 (\xmark)& \cellcolor{red!10} 1.28 (\xmark)& \cellcolor{green!10} 1.02 (\cmark) & 0.0559 \\ \hline
BGL Model 1 & \cellcolor{green!10} 1.01 (\cmark)& \cellcolor{green!10} 1.00 (\cmark)& \cellcolor{green!10} 1.02 (\cmark) & 0.1254 \\
BGL Model 2 & \cellcolor{green!10} 1.21 (\cmark)& \cellcolor{green!10} 1.12 (\cmark)& \cellcolor{green!10} 1.00 (\cmark) & 0.0602 \\
BGL Model 3 & \cellcolor{red!10} 1.31 (\xmark)& \cellcolor{green!10} 1.16 (\cmark)& \cellcolor{green!10} 1.0 (\cmark) & 0.1101 \\

\end{tabular}}
\label{fair_table}
\vspace{-4mm}
\end{table}

\paragraph{Takeaway.} TRIAGE demonstrates the value of sculpting larger datasets by leveraging a limited set of examples. This emphasizes the importance of fit-for-purpose data and demonstrates how it can not only improve predictive performance but also address important issues such as fairness.

\clearpage

\subsection{Data insights}\label{insights_exp}

\paragraph{Goal.} Besides characterizing examples, it is often useful to try to understand which types of samples are in each group. Specifically, for the fit for purpose experiment, we are trying to understand which samples in $\Dtrain$ are sculpted as a consequence of $\Dcal$. Recall $\Dtrain$ are patients from the US and $\Dcal$ and $\Dtest$ are patients from the UK. 

\begin{figure}[!h]
    \centering
    \includegraphics[width=0.7\textwidth]{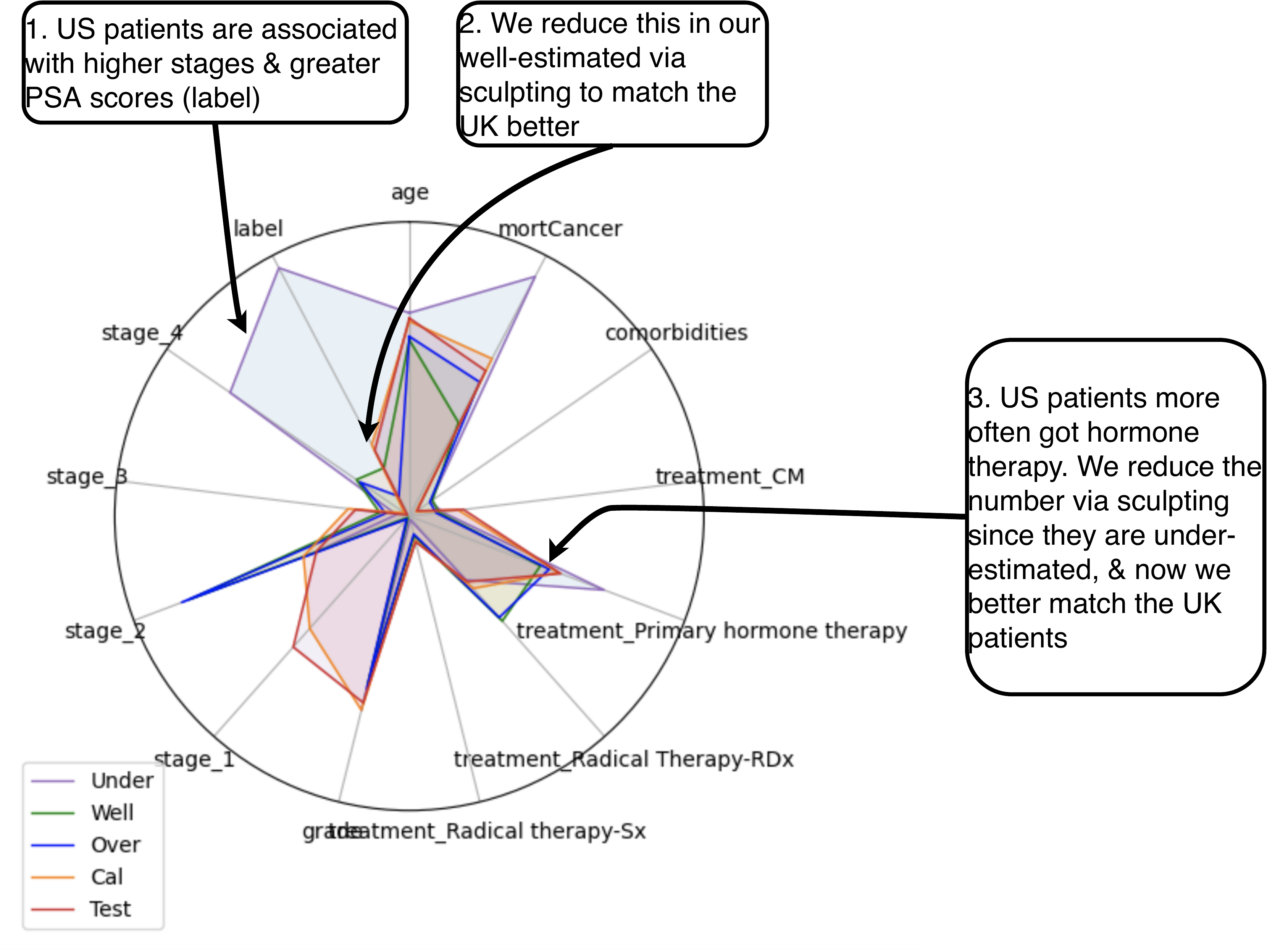}
    \caption{Insights into the different groups resulting from TRIAGE based sculpting. We see we make $\Dtrain$ look more similar to the UK for many features by sculpting.}
    \label{fig:insights}
\end{figure}

We visualize the output of TRIAGE based sculpting using a radial plot in Figure \ref{fig:insights}.

\paragraph{Takeaway.}   US patients ($\Dtrain$) typically have \emph{higher} cancer stages and \emph{higher} PSA scores than their UK counterparts ($\Dcal/\Dtest$). It is precisely these uncommon samples that are filtered by TRIAGE to improve predictive performance. i.e. we can see this based on the samples classed as well-estimated. This now makes the sculpted US data look more similar to the UK data, hence improving the performance of a model trained on said data. In addition, the US patients more often get hormone therapy, in the sculpted well-estimated set, we reduce these by filtering to match the UK better.

\subsection{Validity assessment: Calibration curves \& CRPS}\label{validity}

\paragraph{Goal.} 
We also wish to evaluate the CPDs and their quality. Typically, CPDs are valid, meaning they are well-calibrated. There are three typical ways in which Conformal Predictive Distributions (CPDs) are evaluated for such properties like validity and quality.

\begin{enumerate}
    \item Calibration curves: where the desired curve is diagonal. 
    \item Computing the Continuous Rank Probability (CRPS) score \cite{hersbach2000decomposition}: measures the quality of the predictive distribution, where CRPS=0 is perfect accuracy, and CRPS=1 is fully inaccurate, given by Equation \ref{crps}. 
    \item Assigning quantiles to the CPD and then computing coverage of the resulting intervals. i.e. if we assign quantiles of lower = 0.05 and upper = 0.95, then we desire coverage = 0.9.
\end{enumerate}

\begin{equation}\label{crps}
    CRPS(P,x) = \int_{-\infty}^{\infty} ||P(x^*) - H(x^*-x)||_{2} dx
\end{equation}
where $x$ is the true value of $x$, $P(x^*)$ is the proposed predictive distribution, $H(x)$ the Heaviside step function ($H(x)=1$ if $x=0$, $H(x)=0$ if $x \leq 0$).

We evaluate these metrics for two settings:
\begin{itemize}
    \item \textbf{Fine-grained filter}: in this experiment we are IID, hence we satisfy the exchangeability assumption naturally. Consequently, the CPDs have guaranteed validity.
    \item \textbf{Sculpting for a deployment purpose}: in this experiment $\Dtrain$ and $\Dcal$ might not be exchangeable, hence we wish to assess the potential impact empirically using the different metrics.
\end{itemize}

\subsubsection{Fine-grained filter}

\begin{figure}[!h]
  \centering
 \subfigure[Prostate ]{\includegraphics[width=0.25\textwidth]{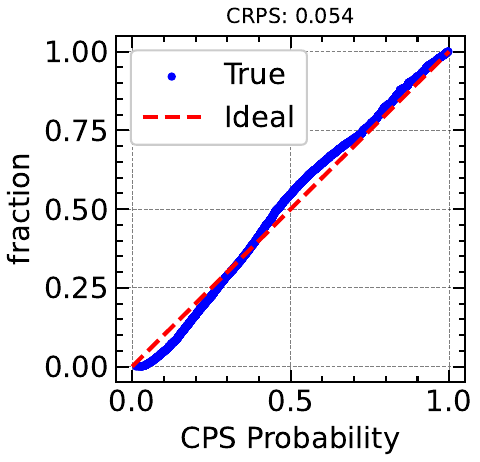}}\quad
  \subfigure[{MIMIC} ]{\includegraphics[width=0.25\textwidth]{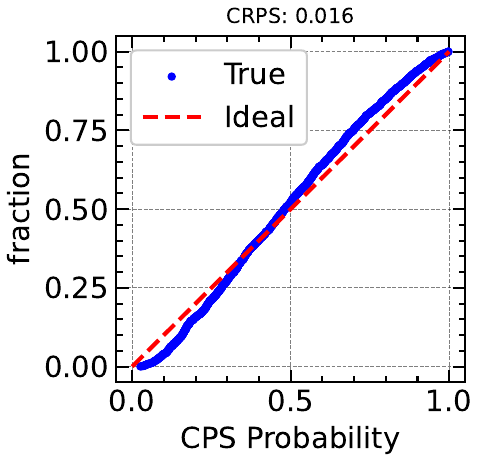}}\quad
  \subfigure[{LOS} ]{\includegraphics[width=0.25\textwidth]{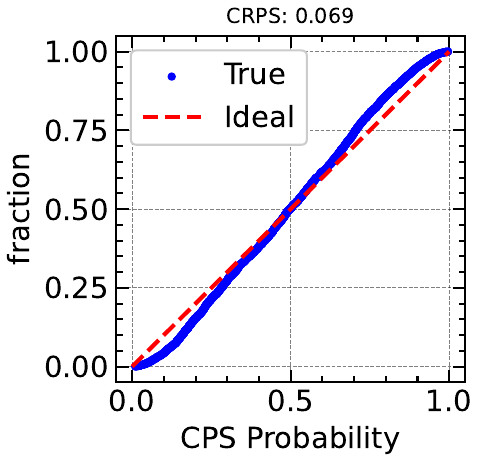}}\quad
  \subfigure[{Concrete} ]{\includegraphics[width=0.25\textwidth]{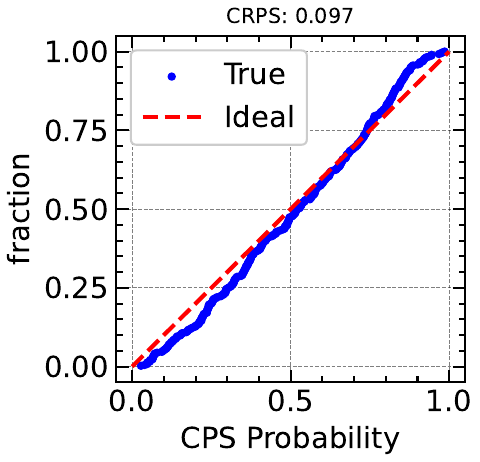}}\quad
  \subfigure[{Bio} ]{\includegraphics[width=0.25\textwidth]{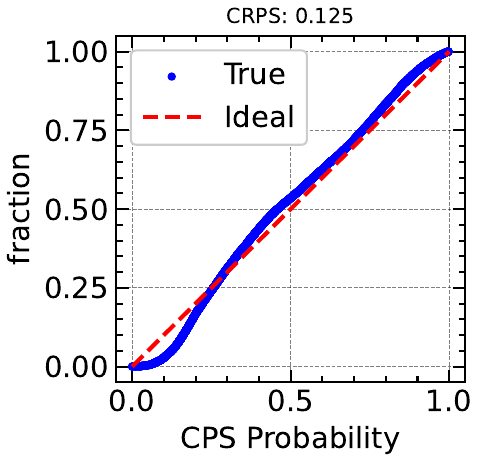}}\quad
  \subfigure[{Protein} ]{\includegraphics[width=0.25\textwidth]{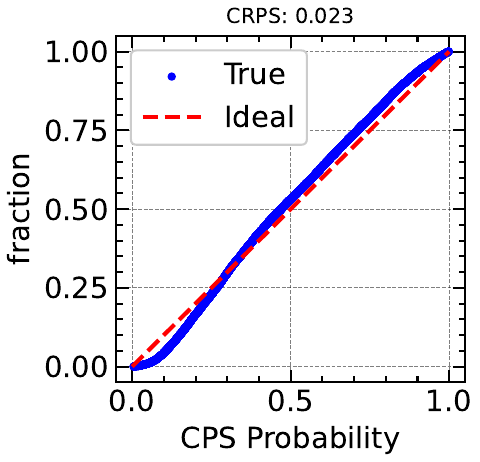}}\quad
  \subfigure[{Star} ]{\includegraphics[width=0.25\textwidth]{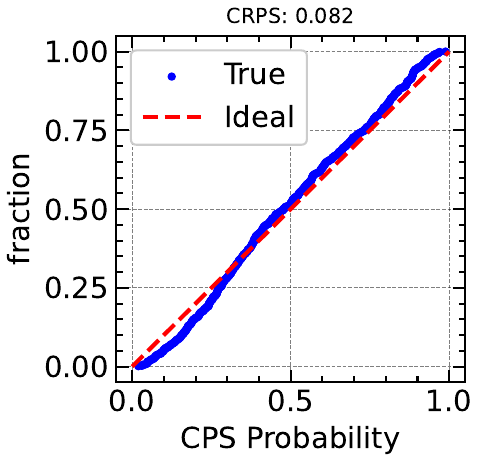}}\quad
  \subfigure[{Bike} ]{\includegraphics[width=0.25\textwidth]{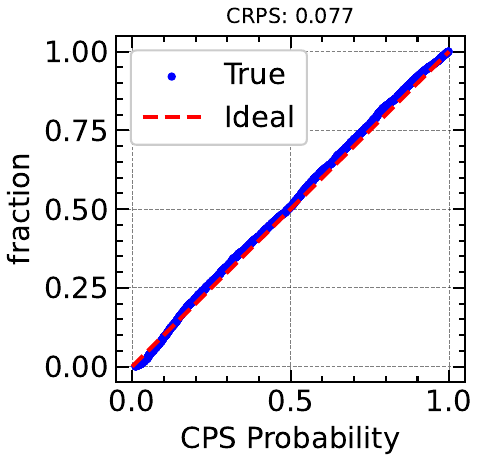}}\quad
  \subfigure[{Boston} ]{\includegraphics[width=0.25\textwidth]{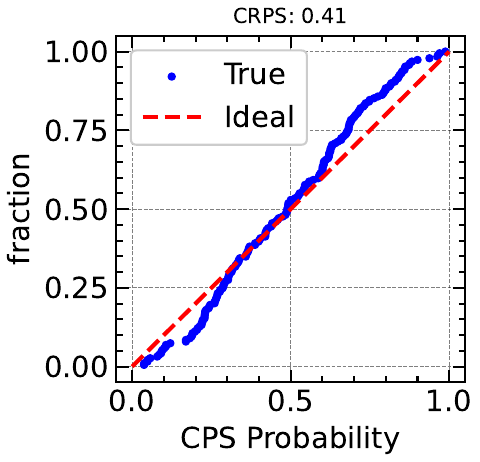}}\quad\\
  
  \caption{Calibration curves for the different datasets, showing that the CPDs are well calibrated, as they match the ideal}
   \label{fig:cpds_iid}
    \rule{\linewidth}{.45pt}
\end{figure}

This is the ideal setting since we are IID hence, the data is naturally exchangeable. Hence, we have guarantees about the validity of our CPDs. However, we still empirically assess the CPDs computing the calibration curves for all datasets, shown in Figure \ref{fig:cpds_iid}. 

\paragraph{Takeaway.} We see that all the datasets have calibration curves matching the ideal diagonal line. The CRPS scores are also very low, indicating high-quality predictive distributions. The takeaway is that indeed the CPDs are well-calibrated.

\newpage
\subsubsection{Sculpting for a deployment purpose}\label{fit_for_purpose_cpd}

\paragraph{Goal.} We now assess the sculpting for a deployment purpose case where we sculpt $\Dtrain$ (US patients) with respect to $\Dcal$ (UK patients). We evaluate this experiment for varying sizes of $\Dcal$ and wish to assess how the potential violation of exchangeability harms calibration, as well as the quality of the CPDs.

\begin{figure}[!h]
  \centering
 \subfigure[Coverage ($\alpha=0.1$)]{\includegraphics[width=0.4\textwidth]{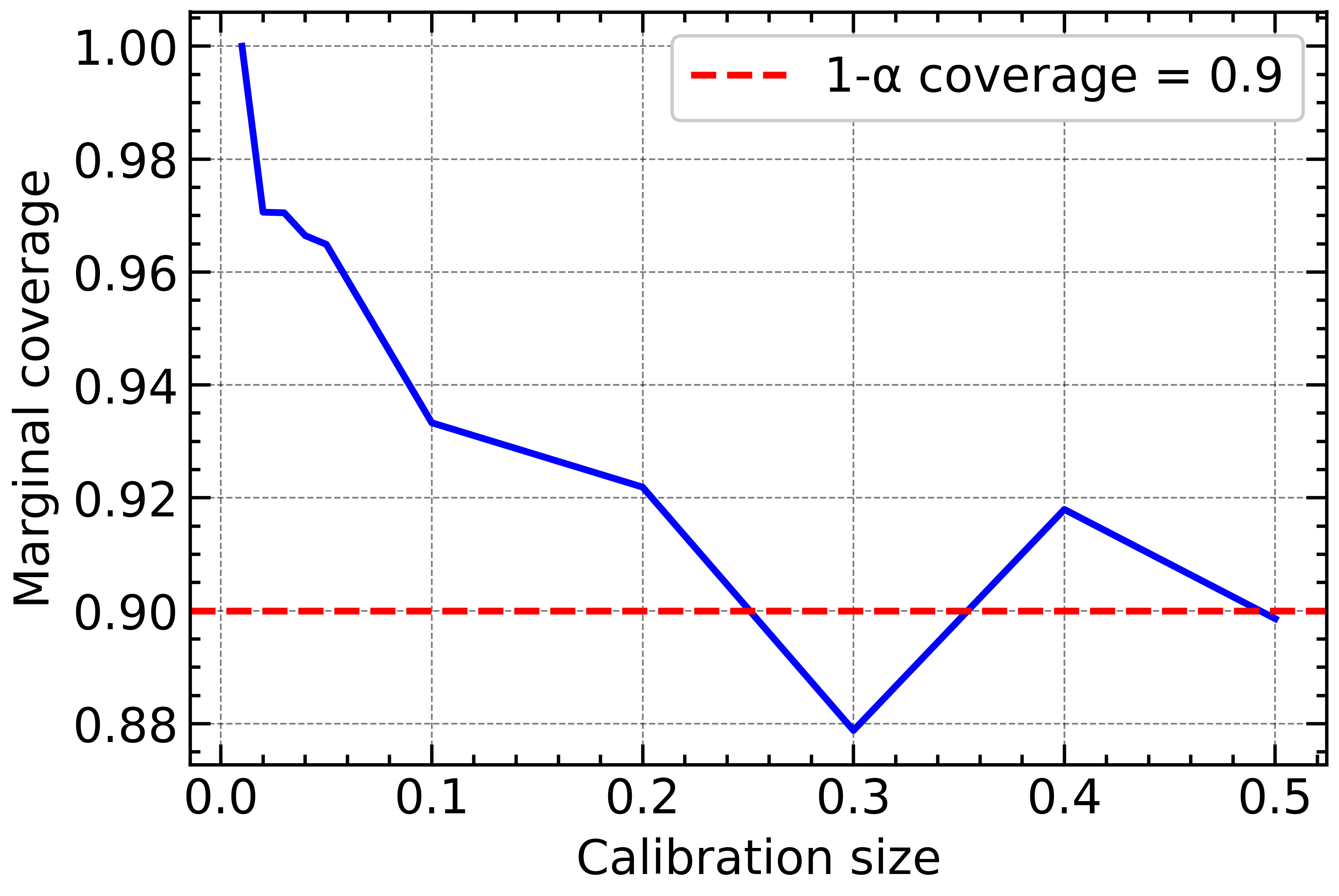}}\quad
\subfigure[CRPS]{\includegraphics[width=0.4\textwidth]{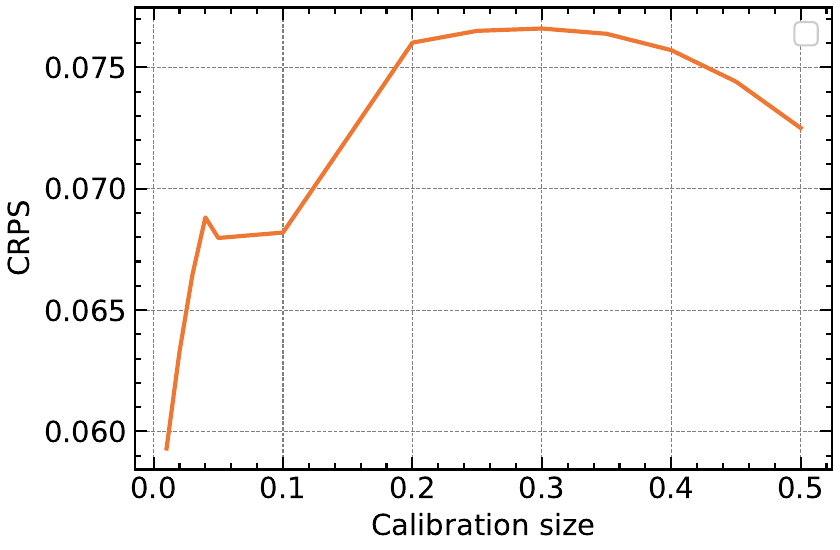}}\quad
  
  \caption{Coverage and CRPS as we increase the size of $\Dcal$}
   \label{fig:scores_cov_crps}
    \rule{\linewidth}{.45pt}
\end{figure}

\begin{figure}[!h]

  \centering
 
\subfigure[0.02]{\includegraphics[width=0.25\textwidth]{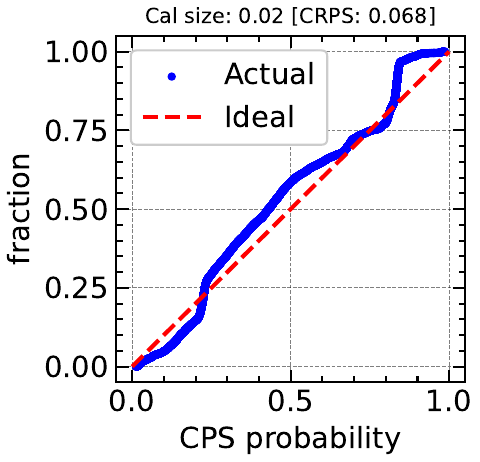}}\quad

\subfigure[0.1]{\includegraphics[width=0.25\textwidth]{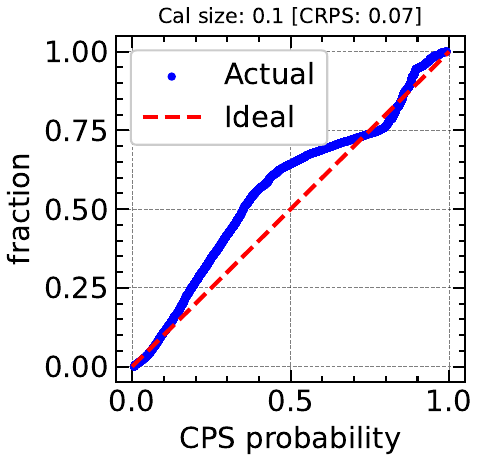}}\quad

\subfigure[0.3]{\includegraphics[width=0.25\textwidth]{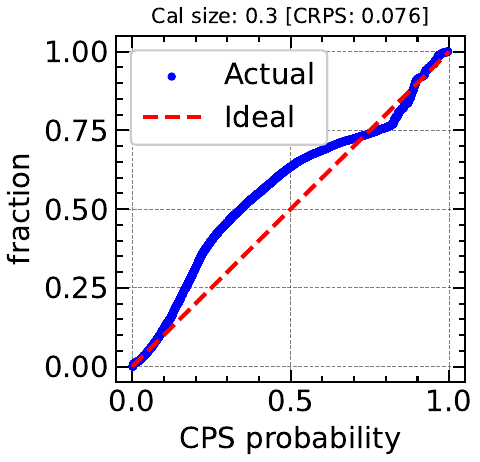}}\quad

  \caption{Calibration curves for different sizes of $\Dcal$}
   \label{fig:cal_ffp_check}
    \rule{\linewidth}{.45pt}
\end{figure}

\paragraph{Results.}
Figure \ref{fig:scores_cov_crps} shows in (a) how coverage varies as the size of $\Dcal$ increases, while (b) looks at how CRPS changes as a function of the size of $\Dcal$. We also check the calibration curves in Figure \ref{fig:cal_ffp_check}. 

\paragraph{Takeaway.} We see the following

\begin{enumerate}
    \item When the calibration set $\Dcal$ is small, $<0.3$ in proportion ($<200/300$ samples) then we still \textbf{achieve marginal coverage} $>0.9$ --- this is also reflected in the low CRPS and good calibration curves.
    \item After $0.3$ ($>300$ samples) we have sufficient samples that violate exchangeability. This reduces the marginal coverage below 0.9.  Interestingly, this \textbf{matches the change-over point we refer to in the main paper - Table \ref{fit_for_purpose}}, where training directly on $\Dcal$ will lead to better performance.
\end{enumerate}

We do, however, highlight that the harm is not significant--- hence our CPDs are still of high quality, as can be seen by the low CRPS score.

\textit{Consequently, we show empirically, even if this may seem an issue --- the resulting CPDs are not significantly harmed by this. Moreover, coverage does not have as dramatic a drop empirically as one might expect}

\clearpage

\subsection{Sculpting for deployment purpose: imbalance and long tails}\label{fit_for_purpose_more}

In the main text, in Section \ref{p3-exp} we looked at sculpting $\Dtrain$ guided by $\Dcal$, in order to make the data more ``fit for purpose''. In this setup, we looked at the data on average. We now take a deeper dive, looking at: (i) Data imbalance (Majority and minority groups) and (ii) Long tails of outcomes

\subsubsection{Data Imbalance }

\paragraph{Goal.} 
The dataset is heavily skewed towards older patients $>65$ years old, with very few younger patients by virtue of the nature of prostate cancer. We wish to understand, beyond average, what is the impact of sculpting on performance of different subsets. In particular, the majority (older) and minority (younger) groups.

We report the results in Table \ref{fit_for_purpose_imbalance} and see the following results. Note: TRIAGE (REST) refers to over- and under-estimated samples used, whilst TRIAGE (OURS) refers to only using well-estimated samples:

\paragraph{Takeaway.} In the case of data imbalance, TRIAGE demonstrates the utility of sculpting the data to be more fit for purpose --- especially in the small sample regime. After around 200 examples, there is reduced benefit to sculpting a pre-existing larger dataset to be more fit for purpose, but rather to directly train on the given examples. We see this benefit also under data imbalance (via attributes) across both majority and minority groups.

Overall, the behavior we see here is similar to the main experiment on average behavior.

\begingroup

\setlength{\tabcolsep}{1.5pt} 
\renewcommand{\arraystretch}{1} 
\begin{table*}[h]
\vspace{-1mm}
 \centering
 \captionsetup{font=footnotesize}
\caption{Data Imbalance: Comparison of MSE for different approaches as we change the number of prior examples we have can access.}
\vspace{-0mm}
\scalebox{0.55}{
\begin{tabular}{l|l|cccccccccc}

\hline

 & $\Dcal$ examples & 10 & 20 & 30 & 40 & 50 & 100 & 200 & 300 & 400 & 500 \\ \hline\hline
\parbox[t]{2mm}{\multirow{11}{*}{\rotatebox[origin=c]{90}{Overall - all samples}}}  & \cellcolor{green!15} TRIAGE (Ours) - $WE$  & \cellcolor{green!15} \bf 0.038+-0.002 & \cellcolor{green!15} \bf 0.037+-0.002 & \cellcolor{green!15} \bf 0.038+-0.002 & \cellcolor{green!15} \bf 0.04+-0.002 & \cellcolor{green!15} \bf 0.04+-0.002 & \cellcolor{green!15} \bf 0.041+-0.002 & \cellcolor{green!15} 0.039+-0.002 & \cellcolor{green!15} 0.039+-0.001 & \cellcolor{green!15} 0.039+-0.001 & \cellcolor{green!15} 0.04+-0.001 \\
 & TRIAGE (REST) $OE \cup UE$ & 0.104+-0.019 & 0.104+-0.018 & 0.107+-0.023 & 0.109+-0.016 & 0.101+-0.019 & 0.124+-0.021 & 0.113+-0.023 & 0.109+-0.021 & 0.115+-0.02 & 0.111+-0.02 \\
  & Baseline ($\Dtrain$) & 0.085+-0.011 & 0.085+-0.011 & 0.085+-0.011 & 0.085+-0.011 & 0.085+-0.011 & 0.085+-0.011 & 0.085+-0.011 & 0.085+-0.011 & 0.085+-0.011 & 0.085+-0.011 \\
 & Baseline ($\Dcal$) & 0.05+-0.01 & 0.047+-0.011 & 0.045+-0.004 & 0.040+-0.003 & 0.041+-0.002 & 0.041+-0.003 & \bf 0.034+-0.001 & \bf 0.033+-0.001 & \bf 0.033+-0.001 & 0.032+-0.001 \\
 & Baseline ($\Dtrain \cup \Dcal$) & 0.071+-0.01 & 0.062+-0.006 & 0.059+-0.006 & 0.054+-0.005 & 0.051+-0.005 & 0.044+-0.004 & 0.039+-0.003 & 0.035+-0.002 & \bf 0.033+-0.002 & \bf 0.031+-0.001 \\
 & Error Sculpt & 0.078+-0.02 & 0.077+-0.019 & 0.077+-0.019 & 0.062+-0.019 & 0.076+-0.019 & 0.078+-0.018 & 0.076+-0.019 & 0.073+-0.019 & 0.076+-0.018 & 0.077+-0.019 \\
 & CP Intervals Sculpt & 0.09+-0.016 & 0.061+-0.022 & 0.045+-0.004 & 0.059+-0.015 & 0.052+-0.006 & 0.054+-0.013 & 0.059+-0.02 & 0.037+-0.001 & 0.034+-0.001 & 0.039+-0.002 \\
 & NGBoost & 0.076+-0.012 & 0.086+-0.011 & 0.083+-0.021 & 0.085+-0.01 & 0.102+-0.017 & 0.109+-0.038 & 0.089+-0.018 & 0.096+-0.015 & 0.084+-0.011 & 0.097+-0.019 \\
 & Bayesian ridge & 0.092+-0.015 & 0.102+-0.023 & 0.098+-0.018 & 0.103+-0.018 & 0.117+-0.025 & 0.094+-0.017 & 0.102+-0.015 & 0.114+-0.028 & 0.115+-0.034 & 0.099+-0.02 \\
 & BNN & 0.081+-0.008 & 0.088+-0.013 & 0.062+-0.007 & 0.101+-0.012 & 0.095+-0.017 & 0.11+-0.015 & 0.092+-0.023 & 0.078+-0.011 & 0.078+-0.009 & 0.077+-0.01 \\
 & GP & 0.122+-0.023 & 0.117+-0.016 & 0.12+-0.019 & 0.113+-0.018 & 0.108+-0.013 & 0.125+-0.017 & 0.131+-0.016 & 0.13+-0.021 & 0.145+-0.024 & 0.144+-0.024 \\ \hline \hline
  & $\Dcal$ examples & 10 & 20 & 30 & 40 & 50 & 100 & 200 & 300 & 400 & 500 \\ \hline\hline
\parbox[t]{2mm}{\multirow{11}{*}{\rotatebox[origin=c]{90}{Majority group (75\%)}}} &  \cellcolor{green!15} TRIAGE (Ours) - $WE$  & \cellcolor{green!15} \bf 0.042+-0.004 & \cellcolor{green!15} \bf 0.041+-0.002 & \cellcolor{green!15} \bf 0.042+-0.003 & \cellcolor{green!15} \bf 0.042+-0.002 & \cellcolor{green!15} \bf 0.044+-0.003 & \cellcolor{green!15} 0.046+-0.002 & \cellcolor{green!15} 0.043+-0.003 & \cellcolor{green!15} 0.043+-0.002 & \cellcolor{green!15} 0.044+-0.002 & \cellcolor{green!15} 0.045+-0.001 \\
 & TRIAGE (Rest) - $OE \cup UE$ & 0.09+-0.018 & 0.093+-0.018 & 0.095+-0.024 & 0.099+-0.018 & 0.086+-0.018 & 0.106+-0.022 & 0.101+-0.023 & 0.093+-0.021 & 0.099+-0.021 & 0.097+-0.02 \\
  & Baseline ($\Dtrain$) & 0.075+-0.013 & 0.075+-0.013 & 0.075+-0.013 & 0.075+-0.013 & 0.075+-0.013 & 0.075+-0.013 & 0.075+-0.013 & 0.075+-0.013 & 0.075+-0.013 & 0.075+-0.013 \\
 & Baseline ($\Dcal$) & 0.055+-0.012 & 0.054+-0.014 & 0.052+-0.004 & 0.044+-0.003 & 0.047+-0.002 & 0.045+-0.003 & 0.037+-0.001 & 0.036+-0.001 & 0.036+-0.001 & 0.035+-0.001 \\
 & Baseline ($\Dtrain \cup \Dcal$) & 0.056+-0.012 & 0.048+-0.006 & 0.045+-0.005 & 0.042+-0.003 & \bf 0.044+-0.003 & \bf 0.037+-0.002 & \bf 0.034+-0.001 & \bf 0.032+-0.001 & \bf 0.032+-0.001 & \bf 0.031+-0.001 \\

 & Error Sculpt & 0.038+-0.002 & 0.039+-0.002 & 0.039+-0.002 & 0.039+-0.002 & 0.039+-0.002 & 0.041+-0.004 & 0.038+-0.002 & 0.038+-0.002 & 0.038+-0.002 & 0.038+-0.002 \\
 & CP Intervals Sculpt & 0.073+-0.012 & 0.06+-0.022 & 0.044+-0.005 & 0.055+-0.013 & 0.053+-0.004 & 0.056+-0.017 & 0.046+-0.005 & 0.039+-0.002 & 0.037+-0.001 & 0.042+-0.002 \\
 & NGBoost & 0.06+-0.01 & 0.073+-0.01 & 0.071+-0.02 & 0.071+-0.009 & 0.097+-0.021 & 0.098+-0.039 & 0.077+-0.02 & 0.08+-0.012 & 0.073+-0.012 & 0.086+-0.02 \\
 & Bayesian ridge & 0.083+-0.017 & 0.099+-0.027 & 0.088+-0.022 & 0.096+-0.02 & 0.114+-0.032 & 0.085+-0.018 & 0.092+-0.014 & 0.109+-0.036 & 0.111+-0.042 & 0.092+-0.024 \\
 & BNN & 0.074+-0.009 & 0.075+-0.018 & 0.05+-0.006 & 0.086+-0.012 & 0.086+-0.019 & 0.106+-0.019 & 0.081+-0.025 & 0.067+-0.009 & 0.066+-0.008 & 0.062+-0.007 \\
 & GP & 0.102+-0.025 & 0.094+-0.014 & 0.095+-0.012 & 0.098+-0.016 & 0.094+-0.009 & 0.105+-0.021 & 0.111+-0.019 & 0.113+-0.02 & 0.129+-0.024 & 0.132+-0.026 \\ \hline \hline
 & $\Dcal$ examples & 10 & 20 & 30 & 40 & 50 & 100 & 200 & 300 & 400 & 500 \\ \hline\hline
\parbox[t]{2mm}{\multirow{11}{*}{\rotatebox[origin=c]{90}{Minority group (25\%)}}} &  \cellcolor{green!15}
TRIAGE (Ours) - $WE$  & \cellcolor{green!15} \bf 0.028+-0.004 & \cellcolor{green!15} \bf 0.026+-0.002 & \cellcolor{green!15} \bf 0.026+-0.002 & \cellcolor{green!15} \bf 0.027+-0.003 & \cellcolor{green!15} \bf 0.026+-0.002 & \cellcolor{green!15} \bf 0.026+-0.002 & \cellcolor{green!15} \bf 0.026+-0.003 & \cellcolor{green!15} 0.026+-0.002 & \cellcolor{green!15} 0.026+-0.003 & \cellcolor{green!15} 0.027+-0.003 \\
 & TRIAGE (Rest) - $OE \cup UE$ & 0.143+-0.027 & 0.135+-0.021 & 0.14+-0.023 & 0.138+-0.014 & 0.141+-0.026 & 0.175+-0.022 & 0.149+-0.027 & 0.155+-0.024 & 0.161+-0.024 & 0.153+-0.027 \\
  & Baseline ($\Dtrain$) & 0.115+-0.016 & 0.115+-0.016 & 0.115+-0.016 & 0.115+-0.016 & 0.115+-0.016 & 0.115+-0.016 & 0.115+-0.016 & 0.115+-0.016 & 0.115+-0.016 & 0.115+-0.016 \\
 & Baseline ($\Dcal$) & 0.036+-0.008 & 0.027+-0.003 & 0.027+-0.004 & 0.025+-0.002 & 0.026+-0.003 & 0.028+-0.005 & \bf 0.026+-0.001 & \bf 0.025+-0.001 & \bf 0.024+-0.001 & \bf 0.024+-0.002 \\
 & Baseline ($\Dtrain \cup \Dcal$) & 0.114+-0.015 & 0.103+-0.015 & 0.1+-0.017 & 0.089+-0.017 & 0.082+-0.015 & 0.066+-0.015 & 0.054+-0.009 & 0.044+-0.007 & 0.037+-0.004 & 0.033+-0.003 \\

 & Error Sculpt & 0.192+-0.074 & 0.185+-0.073 & 0.187+-0.073 & 0.131+-0.071 & 0.18+-0.071 & 0.183+-0.07 & 0.184+-0.071 & 0.173+-0.071 & 0.182+-0.069 & 0.189+-0.071 \\
 & CP Intervals Sculpt & 0.138+-0.034 & 0.063+-0.023 & 0.046+-0.009 & 0.07+-0.022 & 0.048+-0.01 & 0.047+-0.017 & 0.098+-0.069 & 0.029+-0.001 & 0.026+-0.002 & 0.028+-0.003 \\
 & NGBoost & 0.122+-0.027 & 0.124+-0.023 & 0.117+-0.029 & 0.123+-0.018 & 0.115+-0.013 & 0.142+-0.037 & 0.121+-0.016 & 0.142+-0.026 & 0.116+-0.02 & 0.128+-0.022 \\
 & Bayesian ridge & 0.114+-0.017 & 0.11+-0.015 & 0.123+-0.016 & 0.122+-0.017 & 0.124+-0.016 & 0.122+-0.022 & 0.129+-0.022 & 0.127+-0.022 & 0.125+-0.021 & 0.119+-0.02 \\
 & BNN & 0.103+-0.014 & 0.123+-0.023 & 0.095+-0.014 & 0.147+-0.023 & 0.118+-0.016 & 0.122+-0.022 & 0.122+-0.021 & 0.108+-0.022 & 0.109+-0.017 & 0.118+-0.029 \\
 & GP & 0.18+-0.033 & 0.183+-0.034 & 0.193+-0.043 & 0.154+-0.034 & 0.149+-0.035 & 0.182+-0.023 & 0.191+-0.023 & 0.18+-0.035 & 0.19+-0.036 & 0.179+-0.031 \\
\end{tabular}}
\label{fit_for_purpose_imbalance}

\end{table*}
\endgroup
We look into the issue of long tail next

\clearpage
\subsubsection{Long tail of outcomes}

\paragraph{Goal.} Going beyond imbalance on the feature-level, there are also long-tails of outcomes. In this case, the long tails are few samples are associated with high prostate cancer scores. So we partition based on the outcome where the tail greater than 0.75 normed score represents only 3\% of the data, whilst the remainder is 97\%.

We seek to understand the performance via sculpting on both the head and tail of the distributions to assses the impact.

The results in Table \ref{fit_for_purpose_longtail} show the following results. Note: TRIAGE (REST) refers to over- and under-estimated samples used, whilst TRIAGE (OURS) refers to only using well-estimated samples:

\paragraph{Takeaway.}  In the case of having long tails on the outcome, TRIAGE demonstrates the utility of sculpting the data to be more fit for purpose, when assessed overall (averaged across all samples) --- especially in the small sample regime. After around 200 examples, there is reduced benefit to sculpting a pre-existing larger dataset to be more fit for purpose, but rather to directly train on the given examples.

Head: For the head of the distribution which represents the majority of the data this is also the case and we see the best performance for TRIAGE ($WE$). 

Tails: On the long tails of the outcome distribution, we see that in fact using the TRIAGE (REST) - consisting of the ``filtered'' examples ($OE \cup UE$) has the best performance. This result highlights an alternative example selection enabled by TRIAGE. i.e. we should train a specific and better performing model for the  long tails specifically using $OE \cup UE$.

\begingroup

\setlength{\tabcolsep}{1.5pt} 
\renewcommand{\arraystretch}{1} 

\begin{table*}[h]
\vspace{-1mm}
 \centering
 \captionsetup{font=footnotesize}
\caption{Long tail: Comparison of MSE for different approaches as we change the number of prior examples we have can access.}
\vspace{-0mm}
\scalebox{0.55}{
\begin{tabular}{l|l|ccccccccccc}

\hline

 & $\Dcal$ examples & 10 & 20 & 30 & 40 & 50 & 100 & 200 & 300 & 400 & 500 \\ \hline\hline
\parbox[t]{2mm}{\multirow{11}{*}{\rotatebox[origin=c]{90}{Overall - all samples}}}  & \cellcolor{green!15} TRIAGE (Ours) - $WE$ & \cellcolor{green!15} \bf 0.04+-0.002 & \cellcolor{green!15} \bf 0.037+-0.002 & \cellcolor{green!15} \bf 0.037+-0.002 & \cellcolor{green!15} \bf 0.04+-0.002 & \cellcolor{green!15} \bf 0.04+-0.002 & \cellcolor{green!15} \bf 0.04+-0.002 & \cellcolor{green!15} 0.039+-0.002 & \cellcolor{green!15} 0.039+-0.001 & \cellcolor{green!15} 0.039+-0.001 & \cellcolor{green!15} 0.04+-0.0 \\
 & TRIAGE (Rest) - $OE \cup UE$ & 0.103+-0.02 & 0.102+-0.019 & 0.104+-0.017 & 0.104+-0.018 & 0.105+-0.022 & 0.125+-0.022 & 0.114+-0.023 & 0.112+-0.022 & 0.115+-0.021 & 0.112+-0.021 \\
  & Baseline ($\Dtrain$) & 0.085+-0.011 & 0.085+-0.011 & 0.085+-0.011 & 0.085+-0.011 & 0.085+-0.011 & 0.085+-0.011 & 0.085+-0.011 & 0.085+-0.011 & 0.085+-0.011 & 0.085+-0.011 \\
 & Baseline ($\Dcal$) & 0.05+-0.01 & 0.047+-0.011 & 0.045+-0.004 & 0.039+-0.003 & 0.041+-0.002 & 0.041+-0.003 & \bf 0.034+-0.001 & \bf 0.033+-0.001 & \bf 0.033+-0.001 & 0.032+-0.001 \\
 & Baseline ($\Dtrain \cup \Dcal$) & 0.071+-0.01 & 0.062+-0.006 & 0.059+-0.006 & 0.054+-0.005 & 0.051+-0.005 & 0.044+-0.004 & 0.039+-0.003 & 0.035+-0.002 & \bf 0.033+-0.002 & \bf 0.031+-0.001 \\

 & Error Sculpt & 0.08+-0.019 & 0.078+-0.019 & 0.078+-0.019 & 0.062+-0.019 & 0.075+-0.019 & 0.078+-0.018 & 0.077+-0.019 & 0.073+-0.018 & 0.076+-0.018 & 0.077+-0.019 \\
 & CP Intervals Sculpt & 0.091+-0.017 & 0.063+-0.025 & 0.044+-0.004 & 0.058+-0.015 & 0.05+-0.006 & 0.053+-0.012 & 0.058+-0.019 & 0.036+-0.001 & 0.034+-0.001 & 0.039+-0.002 \\
 & NGBoost & 0.075+-0.011 & 0.082+-0.009 & 0.082+-0.022 & 0.087+-0.009 & 0.106+-0.021 & 0.109+-0.038 & 0.089+-0.018 & 0.098+-0.016 & 0.084+-0.011 & 0.098+-0.019 \\
 & Bayesian ridge & 0.091+-0.015 & 0.093+-0.023 & 0.097+-0.018 & 0.104+-0.018 & 0.119+-0.026 & 0.094+-0.017 & 0.104+-0.013 & 0.112+-0.029 & 0.113+-0.032 & 0.099+-0.02 \\
 & BNN & 0.091+-0.015 & 0.114+-0.027 & 0.099+-0.025 & 0.082+-0.014 & 0.075+-0.01 & 0.072+-0.006 & 0.079+-0.014 & 0.065+-0.009 & 0.073+-0.009 & 0.095+-0.018 \\
 & GP & 0.111+-0.024 & 0.117+-0.018 & 0.12+-0.018 & 0.113+-0.018 & 0.108+-0.013 & 0.125+-0.016 & 0.135+-0.014 & 0.129+-0.021 & 0.146+-0.024 & 0.142+-0.024 \\ \hline \hline
 &  $\Dcal$ examples & 10 & 20 & 30 & 40 & 50 & 100 & 200 & 300 & 400 & 500 \\ \hline\hline
\parbox[t]{2mm}{\multirow{11}{*}{\rotatebox[origin=c]{90}{Tail of distribution (3\%)}}}  & TRIAGE (Ours) - $WE$  & 0.574+-0.028 & 0.57+-0.019 & 0.582+-0.025 & 0.604+-0.026 & 0.61+-0.019 & 0.616+-0.016 & 0.597+-0.023 & 0.603+-0.02 & 0.611+-0.018 & 0.621+-0.016 \\
 & \cellcolor{green!15} TRIAGE (Rest) - $OE \cup UE$ & \cellcolor{green!15} \bf 0.272+-0.017 & \cellcolor{green!15} \bf 0.289+-0.036 & \cellcolor{green!15} \bf 0.264+-0.025 & \cellcolor{green!15} \bf 0.256+-0.032 & \cellcolor{green!15} \bf 0.224+-0.021 & \cellcolor{green!15} \bf 0.25+-0.035 & \cellcolor{green!15} \bf 0.246+-0.033 & \cellcolor{green!15} \bf 0.268+-0.028 & \cellcolor{green!15} \bf 0.262+-0.037 & \cellcolor{green!15} \bf 0.268+-0.038 \\
  & Baseline ($\Dtrain$) & 0.315+-0.029 & 0.315+-0.029 & 0.315+-0.029 & 0.315+-0.029 & 0.315+-0.029 & 0.315+-0.029 & 0.315+-0.029 & 0.315+-0.029 & 0.315+-0.029 & 0.315+-0.029 \\
 & Baseline ($\Dcal$) & 0.475+-0.05 & 0.456+-0.042 & 0.424+-0.047 & 0.426+-0.033 & 0.399+-0.026 & 0.382+-0.018 & 0.374+-0.014 & 0.386+-0.024 & 0.39+-0.02 & 0.378+-0.023 \\
 & Baseline ($\Dtrain \cup \Dcal$) & 0.366+-0.032 & 0.364+-0.024 & 0.377+-0.035 & 0.388+-0.028 & 0.389+-0.026 & 0.395+-0.025 & 0.408+-0.018 & 0.407+-0.016 & 0.402+-0.017 & 0.398+-0.017 \\

 & Error Sculpt & 0.503+-0.022 & 0.504+-0.021 & 0.504+-0.024 & 0.502+-0.020 & 0.501+-0.02 & 0.492+-0.026 & 0.488+-0.027 & 0.494+-0.028 & 0.498+-0.028 & 0.495+-0.026 \\
 & CP Intervals Sculpt & 0.346+-0.03 & 0.433+-0.019 & 0.445+-0.050 & 0.447+-0.065 & 0.39+-0.052 & 0.512+-0.033 & 0.526+-0.022 & 0.532+-0.020 & 0.504+-0.025 & 0.509+-0.029 \\
 & NGBoost & 0.42+-0.04 & 0.369+-0.028 & 0.379+-0.019 & 0.324+-0.018 & 0.293+-0.018 & 0.402+-0.025 & 0.425+-0.062 & 0.359+-0.024 & 0.414+-0.049 & 0.408+-0.052 \\
 & Bayesian ridge & 0.357+-0.025 & 0.341+-0.042 & 0.332+-0.045 & 0.333+-0.050 & 0.371+-0.053 & 0.339+-0.036 & 0.299+-0.027 & 0.339+-0.047 & 0.34+-0.039 & 0.341+-0.053 \\
 & BNN & 0.311+-0.032 & 0.367+-0.046 & 0.327+-0.043 & 0.373+-0.034 & 0.364+-0.053 & 0.368+-0.027 & 0.413+-0.054 & 0.370+-0.026 & 0.377+-0.040 & 0.348+-0.060 \\
 & GP & 0.359+-0.055 & 0.302+-0.036 & 0.303+-0.033 & 0.321+-0.032 & 0.32+-0.022 & 0.333+-0.034 & 0.337+-0.049 & 0.357+-0.057 & 0.331+-0.045 & 0.328+-0.042 \\ \hline \hline
 &  $\Dcal$ examples & 10 & 20 & 30 & 40 & 50 & 100 & 200 & 300 & 400 & 500 \\ \hline\hline
\parbox[t]{2mm}{\multirow{11}{*}{\rotatebox[origin=c]{90}{Head of distribution (97\%)}}}  & \cellcolor{green!15} TRIAGE (Ours) - $WE$  & \cellcolor{green!15} \bf 0.028+-0.001 & \cellcolor{green!15} \bf 0.025+-0.002 & \cellcolor{green!15} \bf 0.025+-0.002 & \cellcolor{green!15} \bf 0.027+-0.001 & \cellcolor{green!15} \bf 0.027+-0.002 & \cellcolor{green!15} \bf 0.027+-0.002 & \cellcolor{green!15} \bf 0.026+-0.002 & \cellcolor{green!15} 0.026+-0.002 & \cellcolor{green!15} 0.026+-0.001 & \cellcolor{green!15} 0.027+-0.001 \\
 & TRIAGE (Rest)- $OE \cup UE$ & 0.099+-0.02 & 0.098+-0.019 & 0.101+-0.018 & 0.101+-0.018 & 0.103+-0.022 & 0.122+-0.023 & 0.111+-0.024 & 0.109+-0.022 & 0.112+-0.022 & 0.109+-0.022 \\
  & Baseline ($\Dtrain$) & 0.080+-0.012 & 0.080+-0.012 & 0.080+-0.012 & 0.080+-0.012 & 0.080+-0.012 & 0.080+-0.012 & 0.080+-0.012 & 0.080+-0.012 & 0.080+-0.012 & 0.080+-0.012 \\
 & Baseline ($\Dcal$) & 0.040+-0.011 & 0.037+-0.012 & 0.037+-0.004 & 0.030+-0.003 & 0.033+-0.002 & 0.033+-0.002 & \bf 0.026+-0.001 & \bf 0.025+-0.001 & \bf 0.024+-0.001 & 0.024+-0.001 \\
 & Baseline ($\Dtrain \cup \Dcal$) & 0.065+-0.011 & 0.056+-0.006 & 0.052+-0.006 & 0.047+-0.005 & 0.043+-0.005 & 0.036+-0.004 & 0.031+-0.003 & 0.027+-0.002 &  0.025+-0.001 & \bf 0.023+-0.001 \\
 & Error Sculpt & 0.071+-0.019 & 0.069+-0.020 & 0.068+-0.020 & 0.052+-0.020 & 0.065+-0.019 & 0.068+-0.018 & 0.067+-0.019 & 0.064+-0.019 & 0.066+-0.019 & 0.068+-0.019 \\
 & CP Intervals Sculpt & 0.085+-0.017 & 0.054+-0.026 & 0.035+-0.005 & 0.049+-0.016 & 0.043+-0.007 & 0.043+-0.013 & 0.048+-0.02 & 0.025+-0.002 & 0.023+-0.001 & 0.028+-0.002 \\
 & NGBoost & 0.067+-0.012 & 0.076+-0.009 & 0.076+-0.023 & 0.082+-0.008 & 0.102+-0.021 & 0.102+-0.039 & 0.081+-0.019 & 0.093+-0.016 & 0.076+-0.013 & 0.091+-0.02 \\
 & Bayesian ridge & 0.085+-0.015 & 0.087+-0.024 & 0.092+-0.018 & 0.099+-0.018 & 0.113+-0.027 & 0.088+-0.018 & 0.099+-0.013 & 0.108+-0.030 & 0.108+-0.033 & 0.094+-0.021 \\
 & BNN & 0.086+-0.015 & 0.108+-0.029 & 0.094+-0.026 & 0.076+-0.015 & 0.068+-0.010 & 0.066+-0.006 & 0.072+-0.016 & 0.058+-0.009 & 0.066+-0.010 & 0.089+-0.019 \\
 & GP & 0.105+-0.025 & 0.113+-0.019 & 0.116+-0.019 & 0.109+-0.018 & 0.103+-0.013 & 0.120+-0.017 & 0.131+-0.015 & 0.124+-0.022 & 0.142+-0.025 & 0.138+-0.025 \\

\end{tabular}}
\label{fit_for_purpose_longtail}
\vspace{-4mm}
\end{table*}

\endgroup

\clearpage

\subsection{Comparison with Data Valuation}

\paragraph{Goal.} We have presented a conceptual difference of TRIAGE to data valuation methods in Section \ref{sec:related}. We now experimentally compare TRIAGE with using data valuation methods for data selection. Specifically, we compare to two Shapley valuation methods: (i) TMC-Shapley \cite{datashap}, (ii) KNN-Shapley \cite{knnshap} and LAVA \cite{lava}, which uses optimal transport. We compare in the context of Table 2- data selection for sculpting. The TRIAGE $\Dcal$  is used as the validation set for these methods.

Note: TMC-Shapley: was computationally unfeasible for all 20k samples. Hence, we sample 5k and compare it to TRIAGE separately. (ii) KNN-Shapley and LAVA is run on the original 20k samples.

\paragraph{Takeaway.} 

\textit{Performance:} The results are shown in Tables \ref{dataval1} and \ref{dataval2}. While, these methods are competitive with TRIAGE, we find that TRIAGE tailored to regression outperforms them on downstream MSE performance.

\begin{figure*}[!h]
\vspace{-3mm}
\centering

    \subfigure[{Time vs TMC (Data)-Shapley} ]{\includegraphics[width=0.4\textwidth]{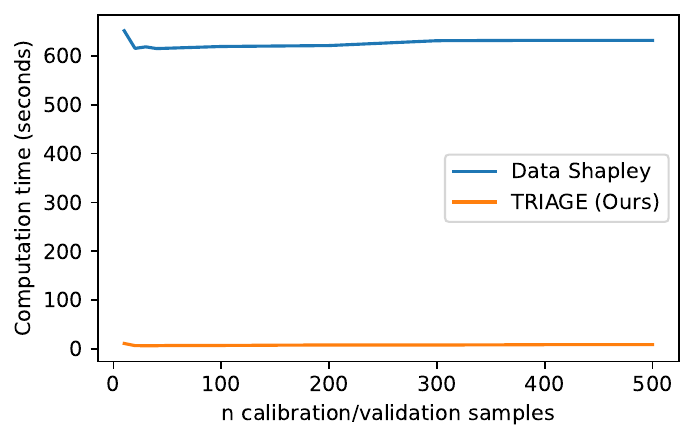}}
    ~
     \subfigure[{Time vs KNN-Shapley} ]{\includegraphics[width=0.3\textwidth]{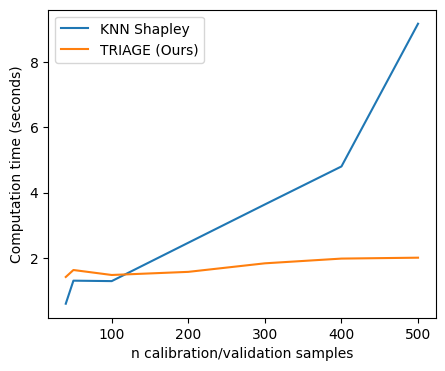}}

     \vspace{-0mm}
\caption{\footnotesize{Computational Time (in seconds): (a) TRIAGE is almost 600X more time efficient than TMC-Shapley, (b) TRIAGE is more efficient than KNN-Shapley, especially as $\D_{cal}$'s size increases.   }}
    \vspace{-0.5mm}
    \rule{\textwidth}{0.1pt}
        \label{fig:datavaltime}
    \vspace{-0mm}
\end{figure*}

\textit{Computational time:} We assess their compute time vs TRIAGE. In Figure \ref{fig:datavaltime}, we show for different sizes of $\Dcal$
 that TRIAGE is more time efficient. Additionally, unlike Shapley methods, our cost doesn’t increase with the size of $\Dcal$. KNN-Shapley is 1-3X and TMC-Shapley 600X more expensive than TRIAGE.

\begin{table*}[!h]

 \centering
 \captionsetup{font=footnotesize}
\caption{Comparison of TRIAGE to Data valuation for different sizes of $\Dcal$. TRIAGE's approach to data sculpting outperforms with lower test MSE ($\downarrow$ better). These rows will be added to Table 2 (main paper)}

\scalebox{0.7}{
\begin{tabular}{l|l|cccccccccc}

\hline
& $\Dcal$ sample size & 10 & 20 & 30 & 40 & 50 & 100 & 200 & 300 \\ \hline\hline
 \multirow{1}{*}{\makecell{Ours (Sculpting)}} & \cellcolor{green!15} \textbf{TRIAGE} ($WE$) & \cellcolor{green!15} \bf 0.051 & \cellcolor{green!15} \bf 0.050 & \cellcolor{green!15} \bf 0.046 & \cellcolor{green!15} \bf 0.046 & \cellcolor{green!15} \bf 0.046 & \cellcolor{green!15} \bf 0.046 & \cellcolor{green!15} \bf 0.045 & \cellcolor{green!15} \bf 0.045 &  \\ 
 \hline\hline

 \multirow{2}{*}{\makecell{Data Valuation}} &  \textbf{KNN-Shapley} 
 &  0.092 
 &  0.095 
 &  0.092
 &  0.122
 & 0.115
 &  0.086
 & 0.054
 & 0.075   \\
 
 & \textbf{LAVA} 
 & 0.055
 & 0.054
 & 0.058 
 & 0.055
 & 0.054 
 & 0.055 
 & 0.058
 & 0.056  \\ \hline\hline

\end{tabular}}
\label{dataval1}
\end{table*}

\begin{table*}[!h]

 \centering
 \captionsetup{font=footnotesize}
\caption{Comparison of TRIAGE to TMC-Shapley for data sculpting for different sizes of $\Dcal$. Due to computational infeasability of TMC-Shapley we subsample to 5k and compare to TRIAGE. TRIAGE's approach to sculpting has lower test MSE mostly compared to TMC-Shapley ($\downarrow$ better).}

\scalebox{0.7}{
\begin{tabular}{l|l|cccccccccc}

\hline

& $\Dcal$ sample size & 10 & 20 & 30 & 40 & 50 & 100 & 200 & 300 \\ \hline\hline
 \multirow{1}{*}{\makecell{Ours (Sculpting)}} & 
\cellcolor{green!15} \textbf{TRIAGE} ($WE$) & \cellcolor{green!15}  \bf  0.0477 & 
0.0488 &   0.0498 & \cellcolor{green!15} \bf 0.0457 & \cellcolor{green!15} \bf 0.0447 & \cellcolor{green!15} \bf 0.0468 & \cellcolor{green!15} \bf 0.0470 & \cellcolor{green!15} \bf  0.0461 &  \\ 
 
 \hline\hline
 \multirow{1}{*}{\makecell{Data Valuation}} & \textbf{TMC-Shapley}  
 &   0.0580
 &  \bf 0.0441 
 &  \bf  0.0490 
 &  0.0472 
 &   0.0580 
 &  0.0491 
 &  0.0501 
 & 0.0467 &  \\ 
\hline\hline
\end{tabular}}
\label{dataval2}
\end{table*}

\newpage
\subsection{Synthetic analysis under Huber's contamination model}\label{huber-appendix}

TRIAGE aims to sculpt data for robust regression. One can of course draw similarities to robust statistics which also aims to train robust models. (1) Post-hoc vs Built-in: TRIAGE wraps a regressor to detect and sculpt outlier samples, whereas Robust Statistics embeds outlier resilience within the model \cite{steinhardt,hubermaronna}, e.g. via Huber loss \cite{hubermaronna}. (2) Additional data-centric applications: TRIAGE tackles diverse "data-centric AI" tasks, like comparing synthetic data (Sec. 5.3.1) and feature acquisition/collection (Sec. 5.3.2), which is beyond the scope of robust statistics.

(ii) Theoretical Analysis: Connecting TRIAGE theoretically to robust statistics is an intriguing question. However, we highlight two important challenges that could be tackled by future work, for inspiration see \cite{steinhardt}:

(1) Interdependence of CPD and training dynamics in TRIAGE. Disentangling their impacts theoretically is non-trivial.
(2) Scores across training epochs are correlated due to iterative model training. This highlights the complexity of any theoretical guarantee, given the dynamic nature of the scores and their correlation.

As a step towards this we contrast TRIAGE with robust statistics. We 
provide a simulation using Huber's Contamination Model: Our simulation setup mirrors \cite{huberchen}, generating data from a linear model $y=X\beta + \eta$, where $X \sim U[0,1]$ and $\eta \sim N(0,1)$ .

Mimicking Huber's model we contaminate the response $y$ corrupting $\epsilon$ samples replacing $\eta_i$ with $\eta_i+c_i$, where $c_i$
comes from a different distribution. 

We compare TRIAGE with: (i) Error baseline, (ii) Training with Huber Loss, (iii) TRIAGE applied to a Huber Loss trained model.

\begin{figure*}[!h]

\centering
\subfigure[{Comparison to error baseline} ]{\includegraphics[width=0.4\textwidth]{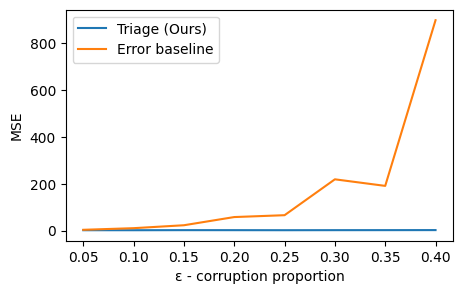}}
~
  \subfigure[{TRIAGE is compatible with robust losses} ]{\includegraphics[width=0.4\textwidth]{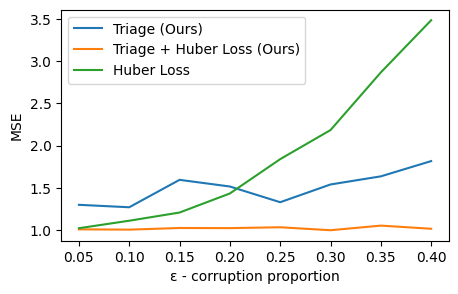}}
    \caption{\footnotesize{Simulation with Huber's contamination model: (a) TRIAGE has lower MSE as $\epsilon$ increases and (b) TRIAGE combined with a model trained with a Huber loss has lower MSE than both alone. }}\label{fig:huber}
    \vspace{-0.5mm}
    \rule{\textwidth}{0.1pt}
    \vspace{-0mm}
    \end{figure*}

The results in Figure \ref{fig:huber} show that TRIAGE has a lower MSE compared to the error baseline as $\epsilon$ rises. TRIAGE is also stable in response to contamination due to the clean calibration set. Interestingly, combining TRIAGE with a model trained using Huber’s loss proves superior to using either alone, highlighting the compatibility of TRIAGE with robust techniques.

\newpage
\subsection{Differentiation of general-purpose scoring methods with TRIAGE scores}\label{diff_other_scores}

\paragraph{Goal.} The main text (Section \ref{p2-exp}) illustrated that for the same magnitude of general-purpose scores these can be associated with different TRIAGE scores. Hence, highlighting that TRIAGE scores offer a viable alternative to differentiating samples.

\begin{figure}[!h]

  \centering
  \subfigure[
{Errors}]{\includegraphics[width=0.4\textwidth]{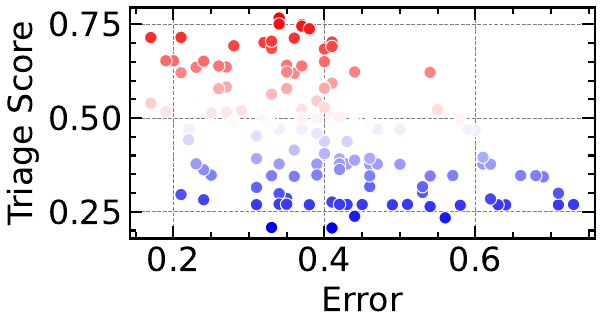}}\quad
  \subfigure[{Loss} ]{\includegraphics[width=0.4\textwidth]{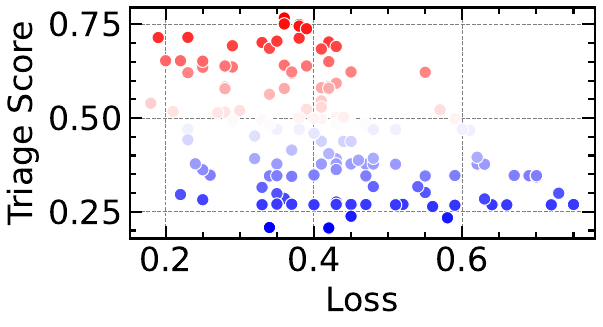}}\quad\\
  \subfigure[{Grad}]{\includegraphics[width=0.4\textwidth]{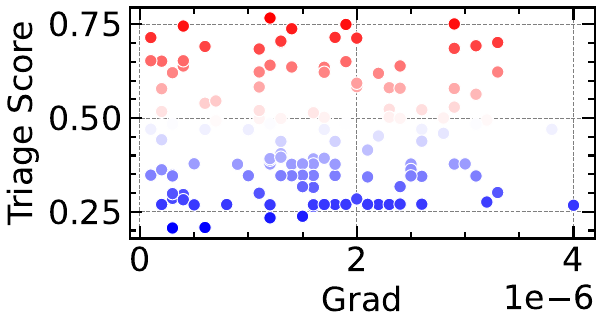}}\quad
  \subfigure[{VoG} ]{\includegraphics[width=0.4\textwidth]{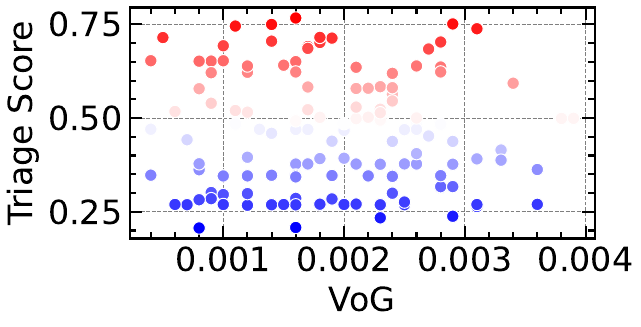}}\quad
  \caption{Samples with the same general-purpose scores are often associated with,  different Triage scores, highlighting the potential to differentiate sample}
   \label{fig:diff_all}
    \rule{\linewidth}{.45pt}
\end{figure}

In the main text, we looked at error vs TRIAGE scores . We now examine the remainder of general purpose scoring methods in Figure \ref{fig:diff_all}.

\paragraph{Takeaway.} The phenomena where TRIAGE scores can be used to differentiate samples with the same magnitude holds true across the different scoring approaches.

\clearpage

\subsection{Fine-grained filtering: Additional datasets}\label{filter-additional}

\paragraph{Goal.} In the main paper we demonstrated fine-grained filtering on a dataset to showcase the potential. We now repeat on all datasets.

\begin{figure}[!h]
  \centering
  \subfigure[
\scriptsize{Errors}]{\includegraphics[width=0.23\textwidth]{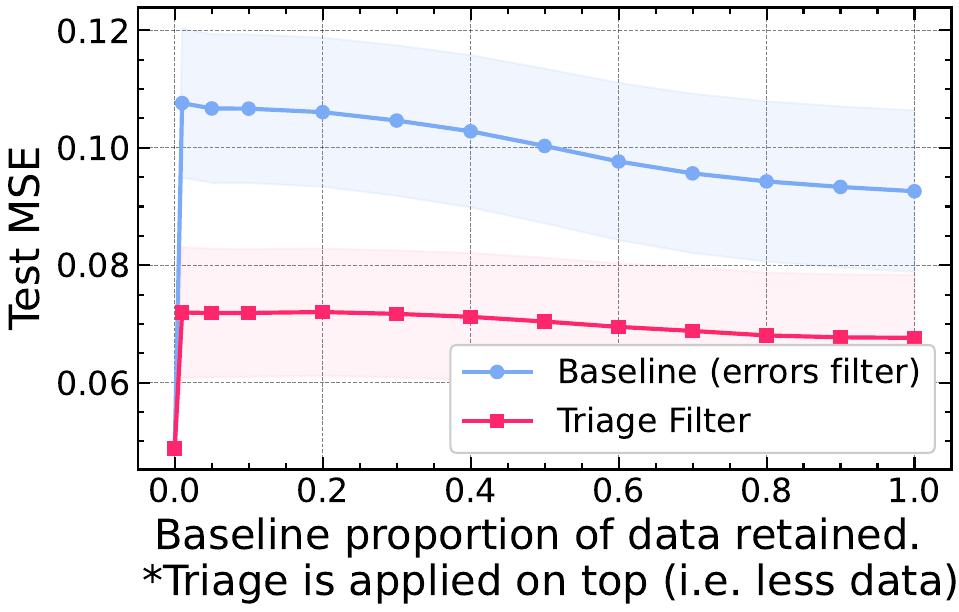}}\quad
  \subfigure[\scriptsize{Loss} ]{\includegraphics[width=0.23\textwidth]{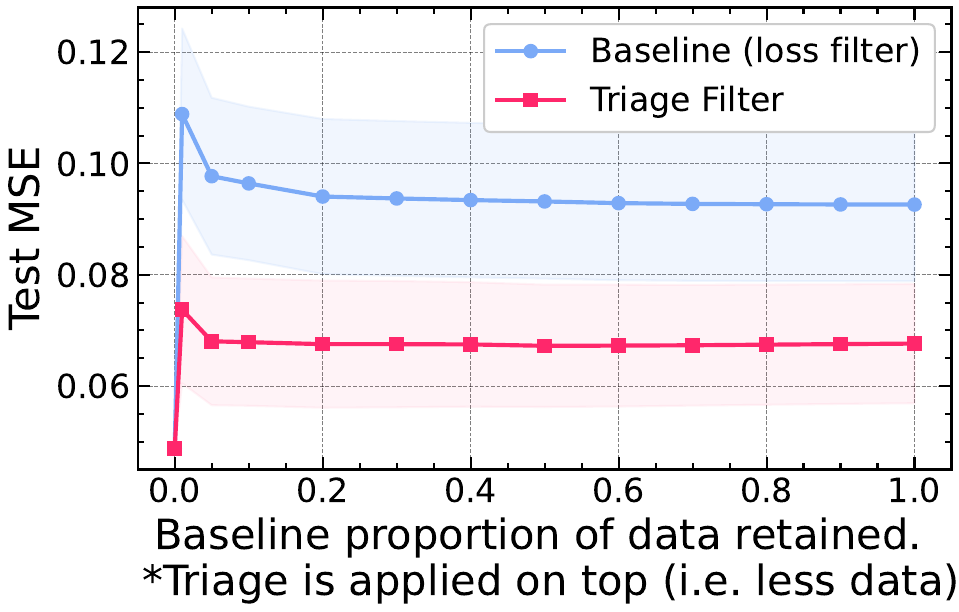}}\quad\\
  \vspace{-2mm}
  \subfigure[\scriptsize{Grad}]{\includegraphics[width=0.23\textwidth]{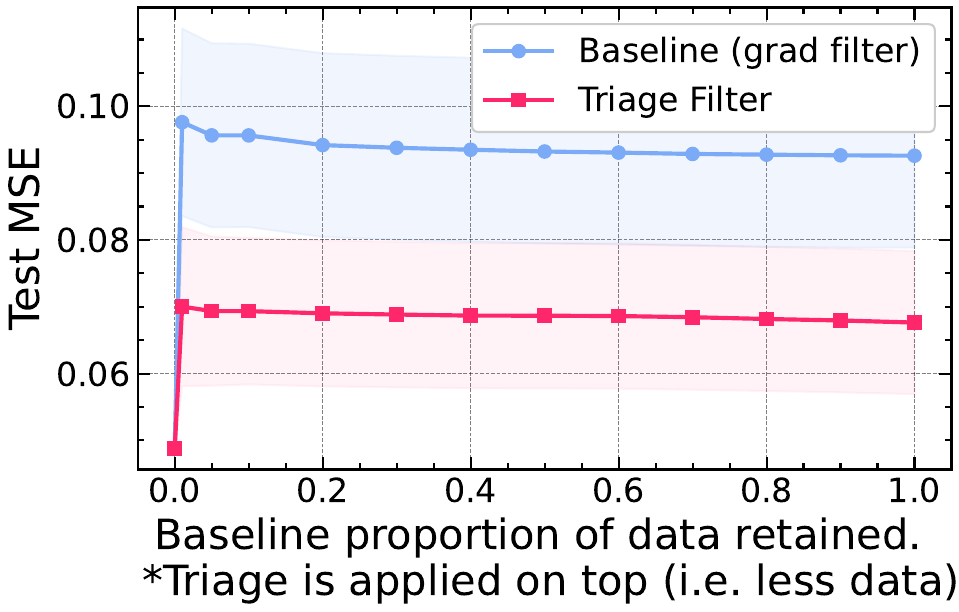}}\quad
  \subfigure[\scriptsize{VoG} ]{\includegraphics[width=0.23\textwidth]{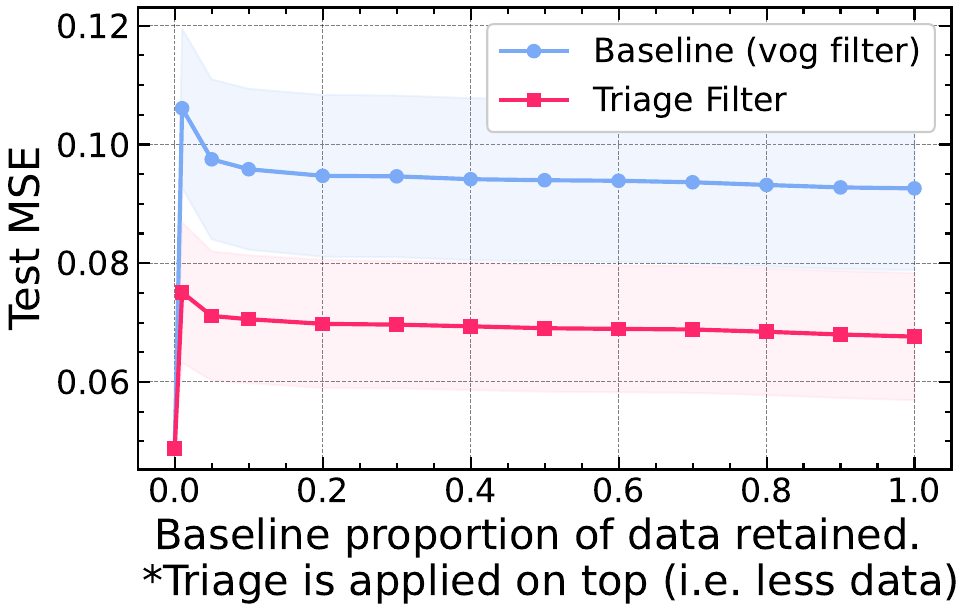}}\quad
  \vspace{-2mm}
  \caption{Fine-grained filter: Bike}
   \label{fig:filter_exp_bike}
\vspace{-2mm}
    \rule{\linewidth}{.45pt}
\vspace{-5mm}
\end{figure}

\begin{figure}[!h]
  \centering
  \subfigure[
\scriptsize{Errors}]{\includegraphics[width=0.23\textwidth]{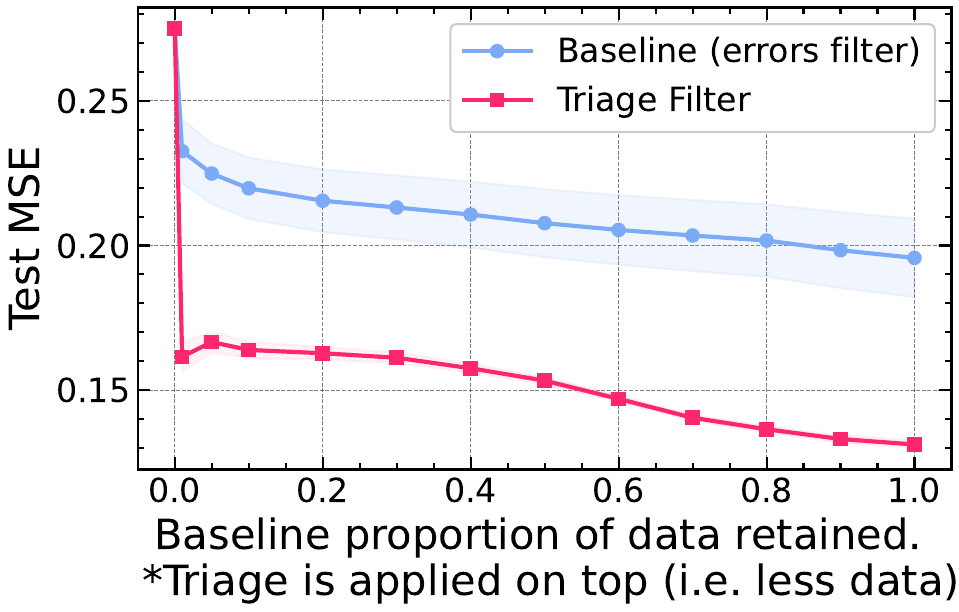}}\quad
  \subfigure[\scriptsize{Loss} ]{\includegraphics[width=0.23\textwidth]{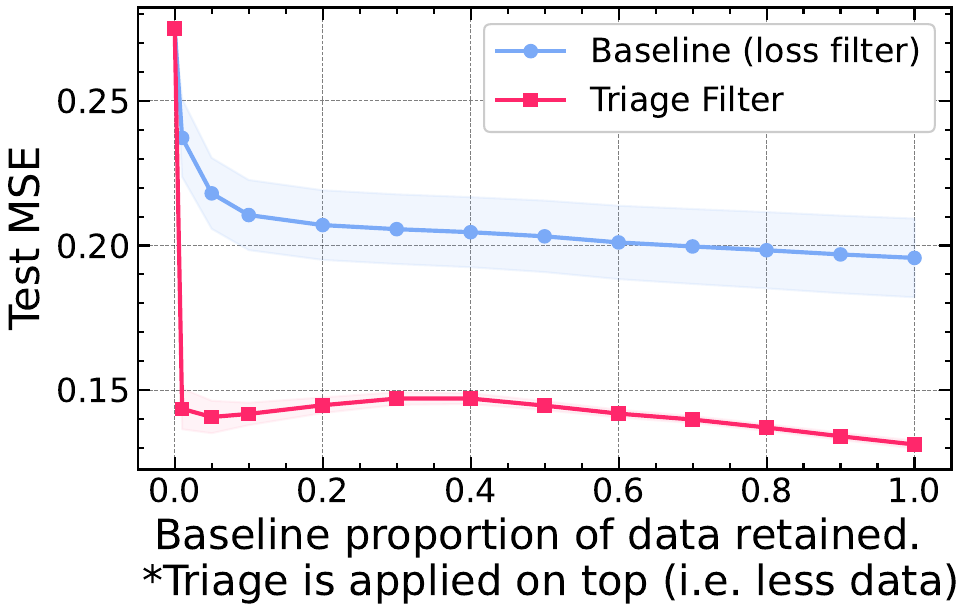}}\quad\\
  \vspace{-2mm}
  \subfigure[\scriptsize{Grad}]{\includegraphics[width=0.23\textwidth]{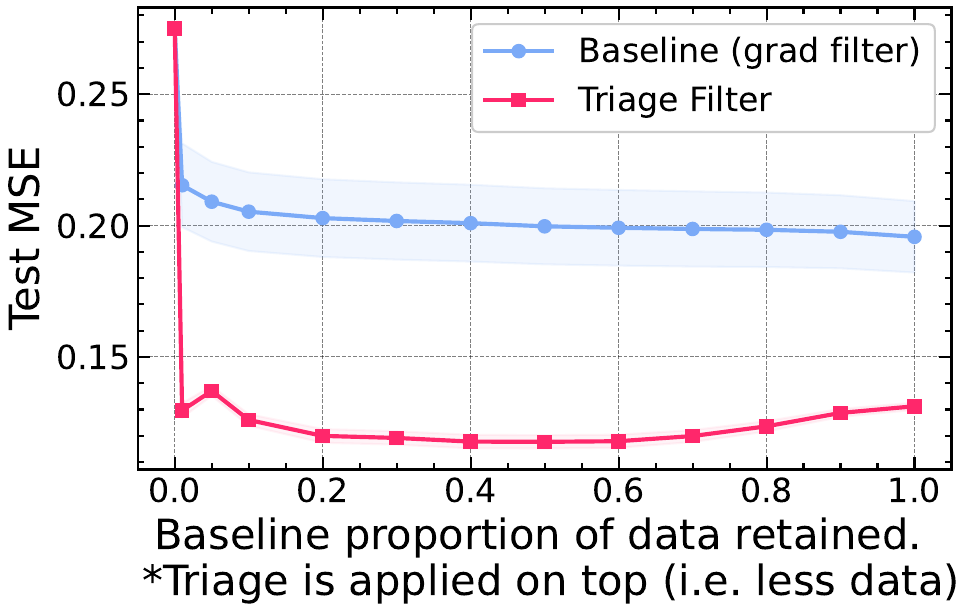}}\quad
  \subfigure[\scriptsize{VoG} ]{\includegraphics[width=0.23\textwidth]{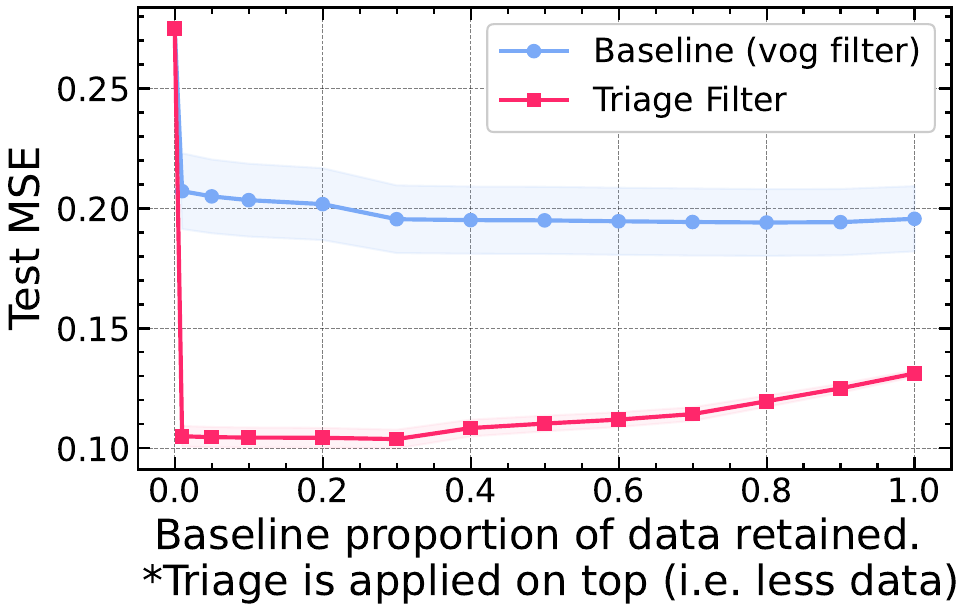}}\quad
  \vspace{-2mm}
  \caption{Fine-grained filter: Bio}
   \label{fig:filter_exp_bio}
\vspace{-2mm}
    \rule{\linewidth}{.45pt}
\vspace{-5mm}
\end{figure}

\begin{figure}[!h]
  \centering
  \subfigure[
\scriptsize{Errors}]{\includegraphics[width=0.23\textwidth]{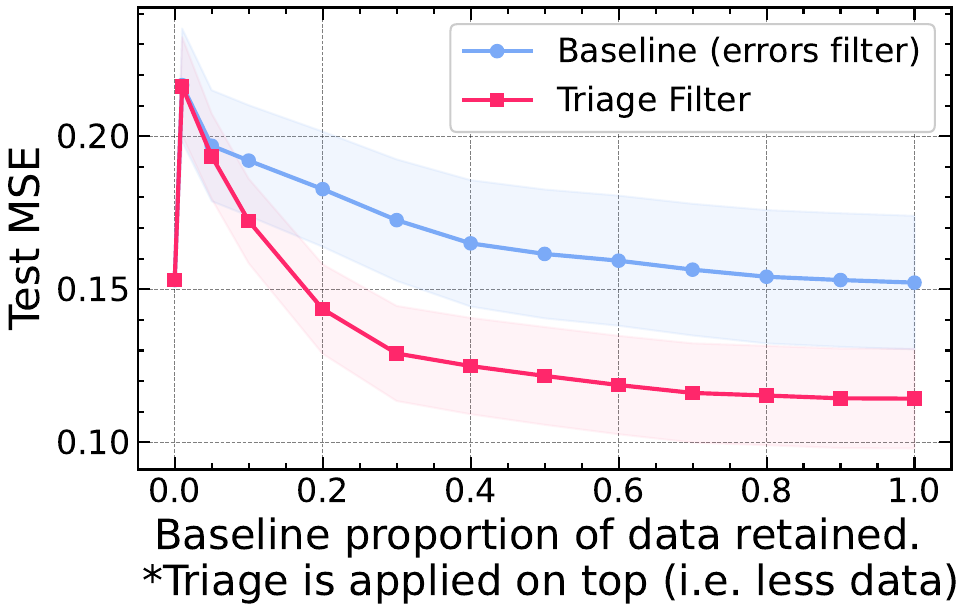}}\quad
  \subfigure[\scriptsize{Loss} ]{\includegraphics[width=0.23\textwidth]{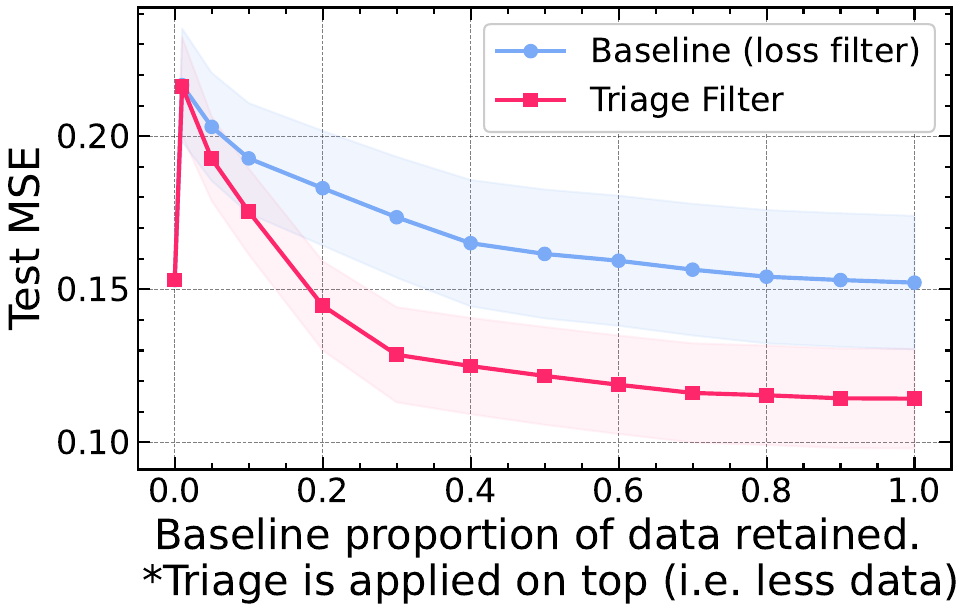}}\quad\\
  \vspace{-2mm}
  \subfigure[\scriptsize{Grad}]{\includegraphics[width=0.23\textwidth]{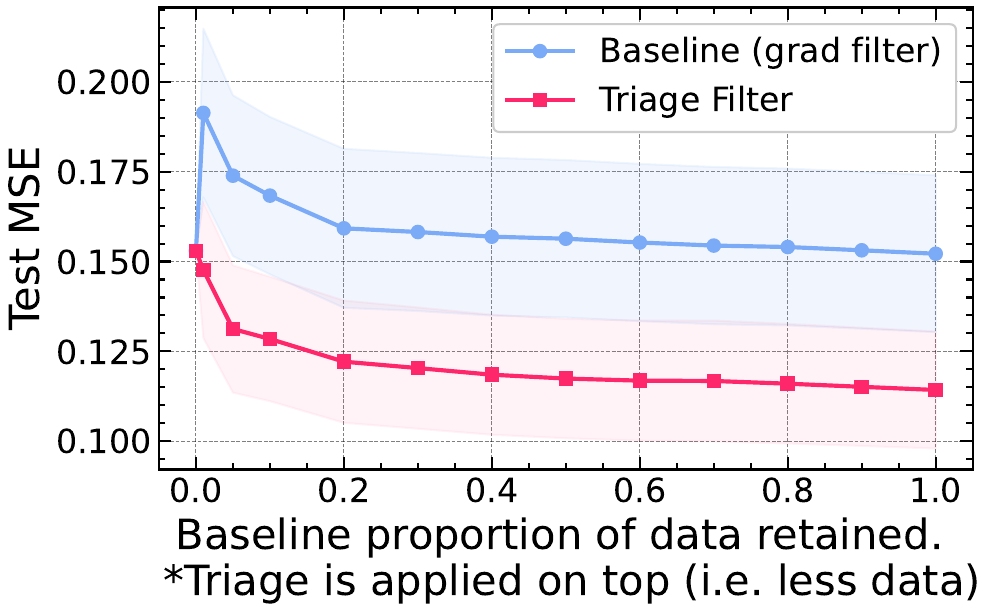}}\quad
  \subfigure[\scriptsize{VoG} ]{\includegraphics[width=0.23\textwidth]{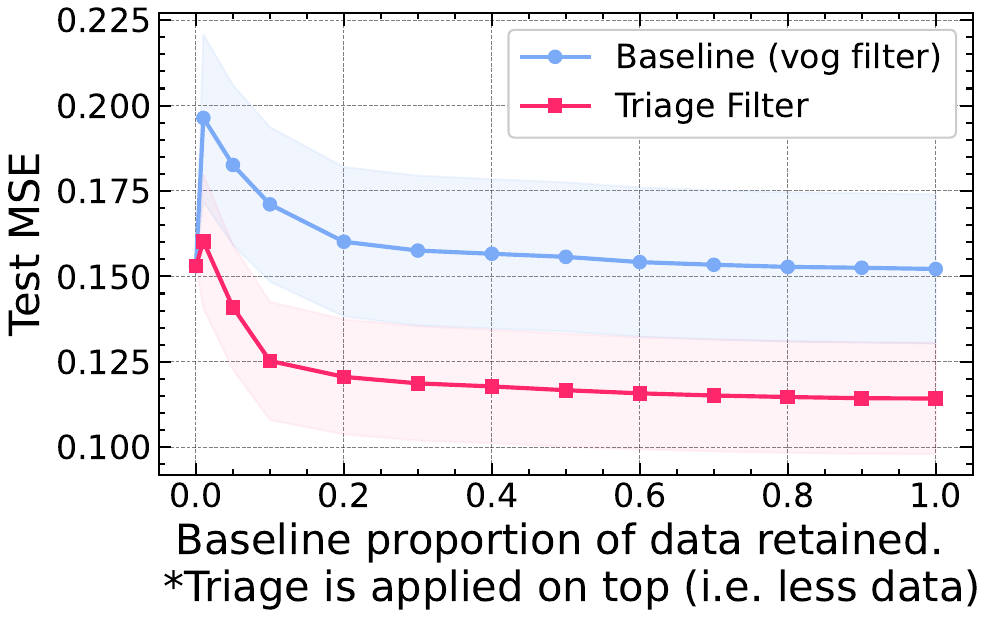}}\quad
  \vspace{-2mm}
  \caption{Fine-grained filter: Concrete}
   \label{fig:filter_exp_concrete}
\vspace{-2mm}
    \rule{\linewidth}{.45pt}
\vspace{-5mm}
\end{figure}

\begin{figure}[!h]
  \centering
  \subfigure[
\scriptsize{Errors}]{\includegraphics[width=0.23\textwidth]{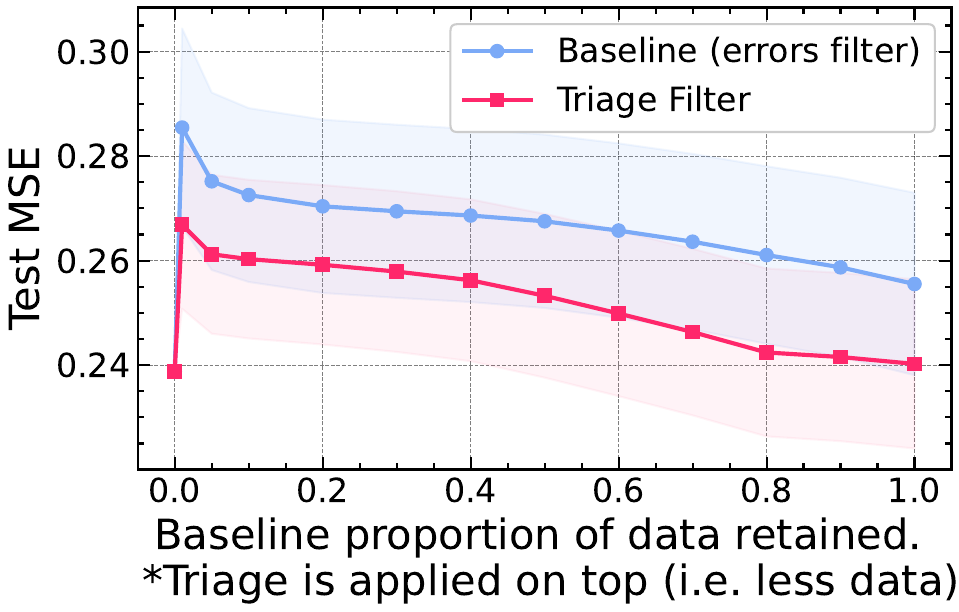}}\quad
  \subfigure[\scriptsize{Loss} ]{\includegraphics[width=0.23\textwidth]{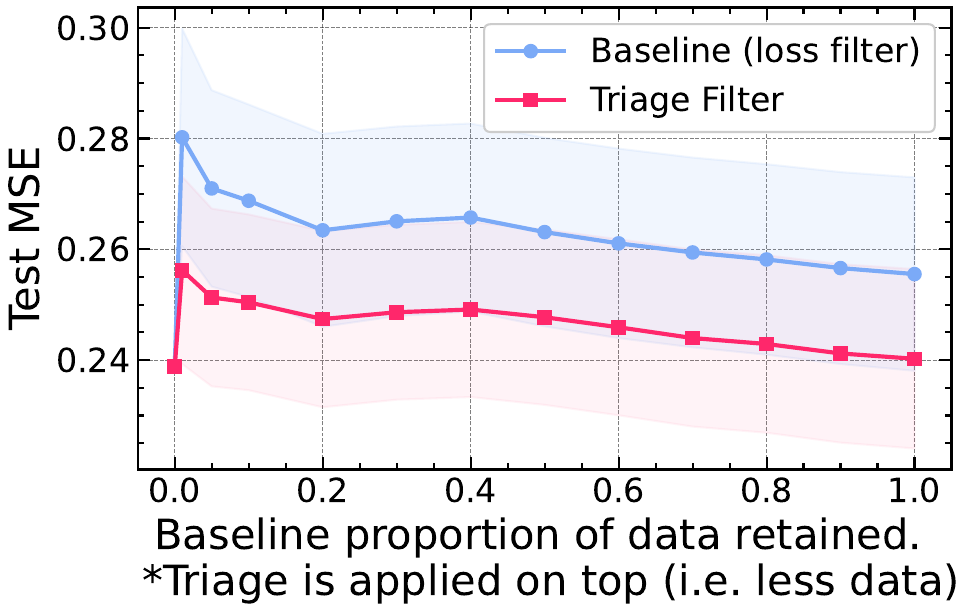}}\quad\\
  \vspace{-2mm}
  \subfigure[\scriptsize{Grad}]{\includegraphics[width=0.23\textwidth]{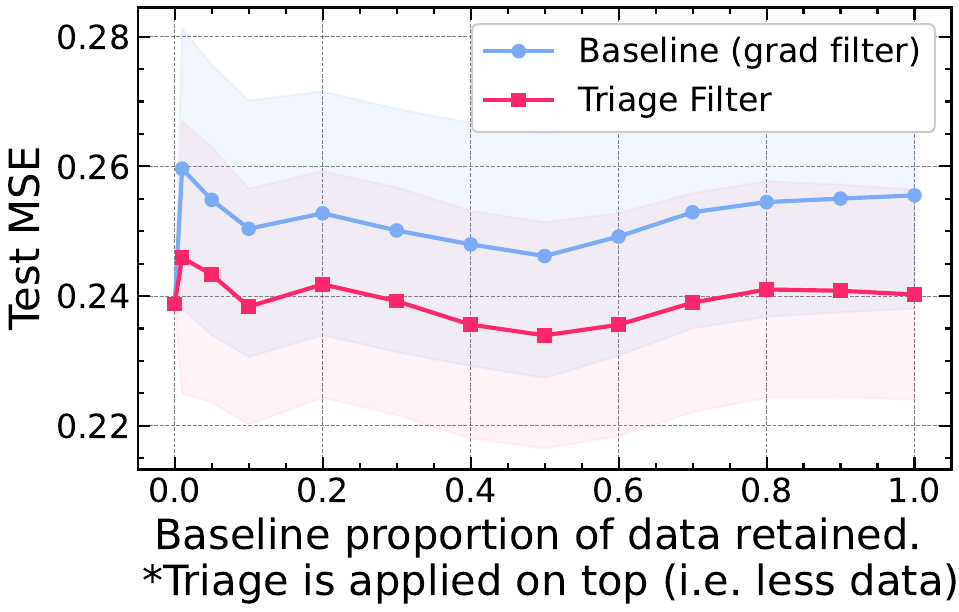}}\quad
  \subfigure[\scriptsize{VoG} ]{\includegraphics[width=0.23\textwidth]{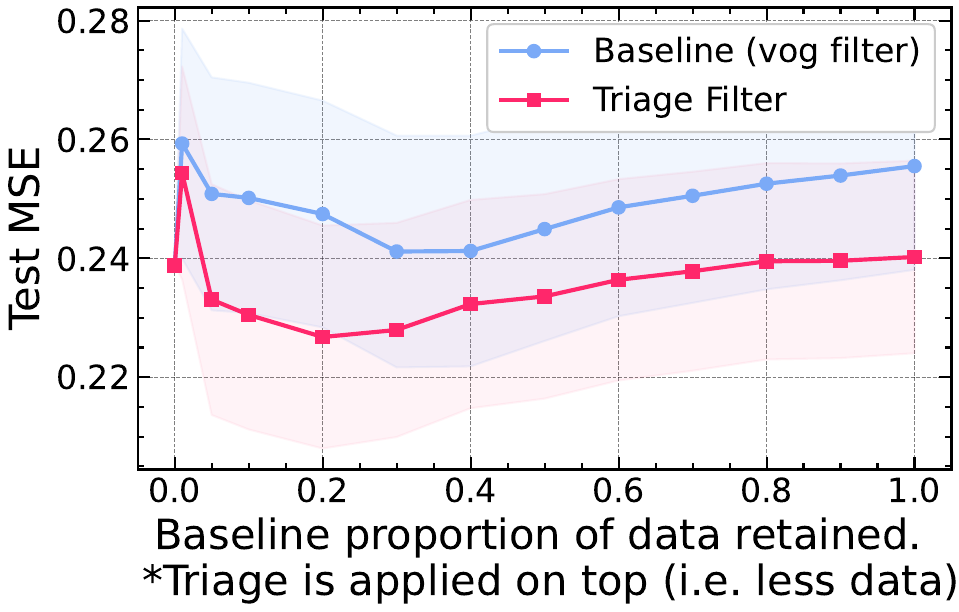}}\quad
  \vspace{-2mm}
  \caption{Fine-grained filter: Prostate}
   \label{fig:filter_exp_cancer}
\vspace{-2mm}
    \rule{\linewidth}{.45pt}
\vspace{-5mm}
\end{figure}

\begin{figure}[!h]
  \centering
  \subfigure[
\scriptsize{Errors}]{\includegraphics[width=0.23\textwidth]{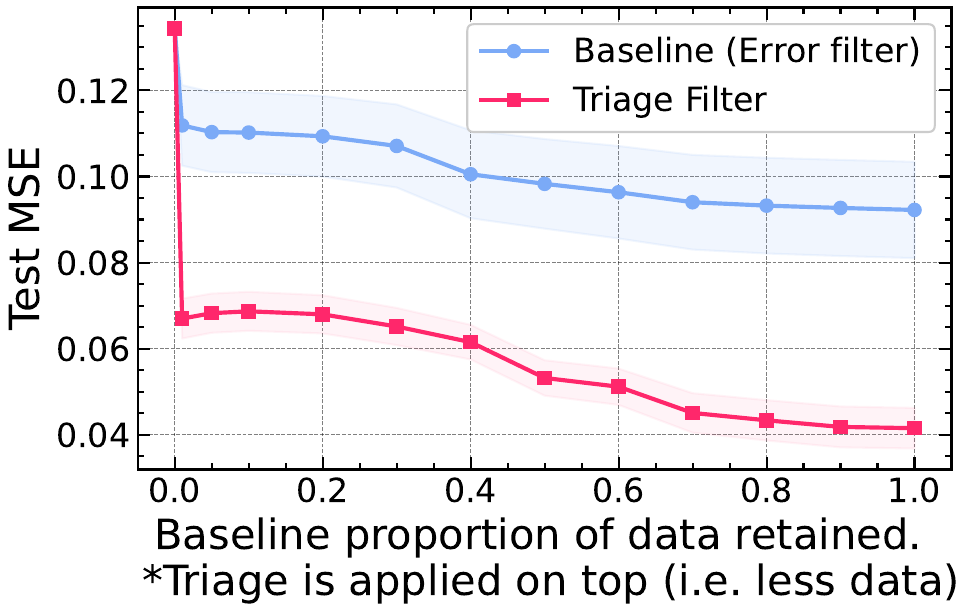}}\quad
  \subfigure[\scriptsize{Loss} ]{\includegraphics[width=0.23\textwidth]{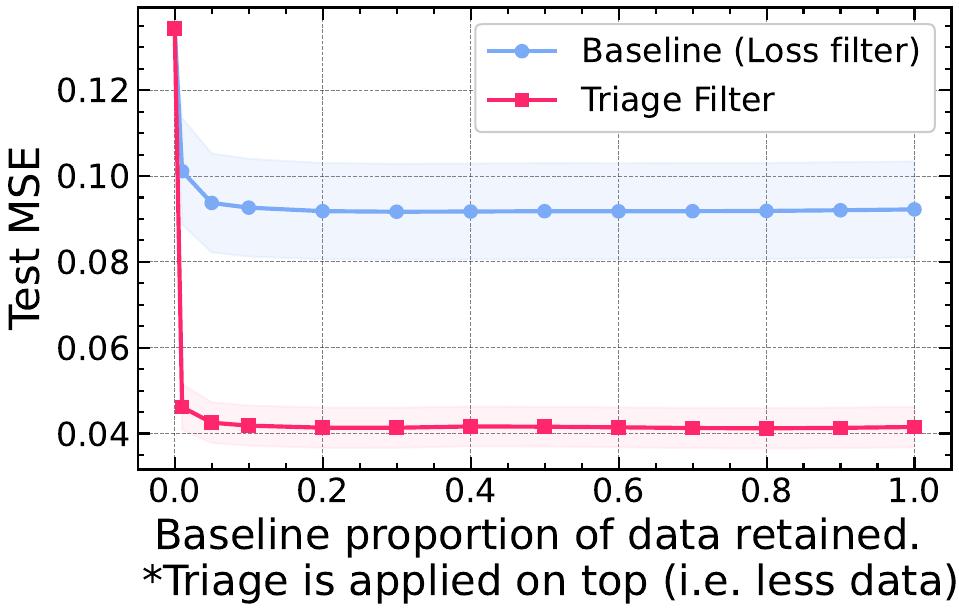}}\quad\\
  \vspace{-2mm}
  \subfigure[\scriptsize{Grad}]{\includegraphics[width=0.23\textwidth]{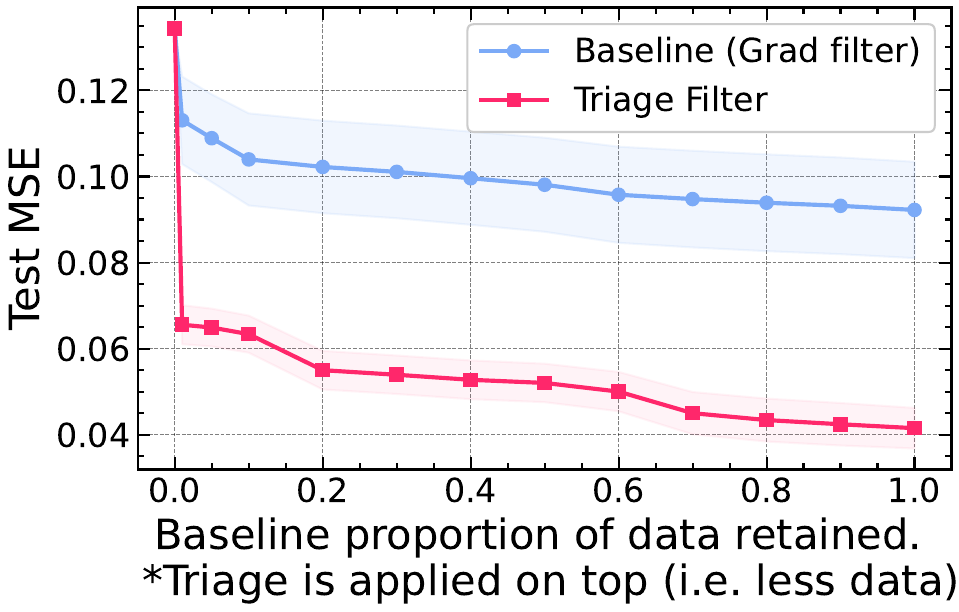}}\quad
  \subfigure[\scriptsize{VoG} ]{\includegraphics[width=0.23\textwidth]{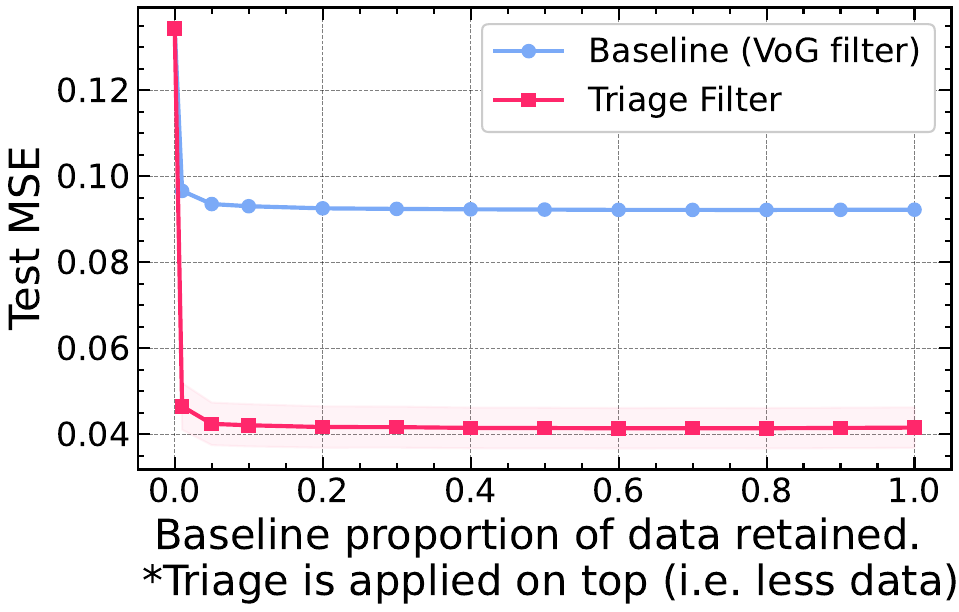}}\quad
  \vspace{-5mm}
  \caption{Fine-grained filter: LoS}
   \label{fig:filter_exp_los}
\vspace{-2mm}
    \rule{\linewidth}{.45pt}
\vspace{-10mm}
\end{figure}

\begin{figure}[!h]
  \centering
  \subfigure[
\scriptsize{Errors}]{\includegraphics[width=0.23\textwidth]{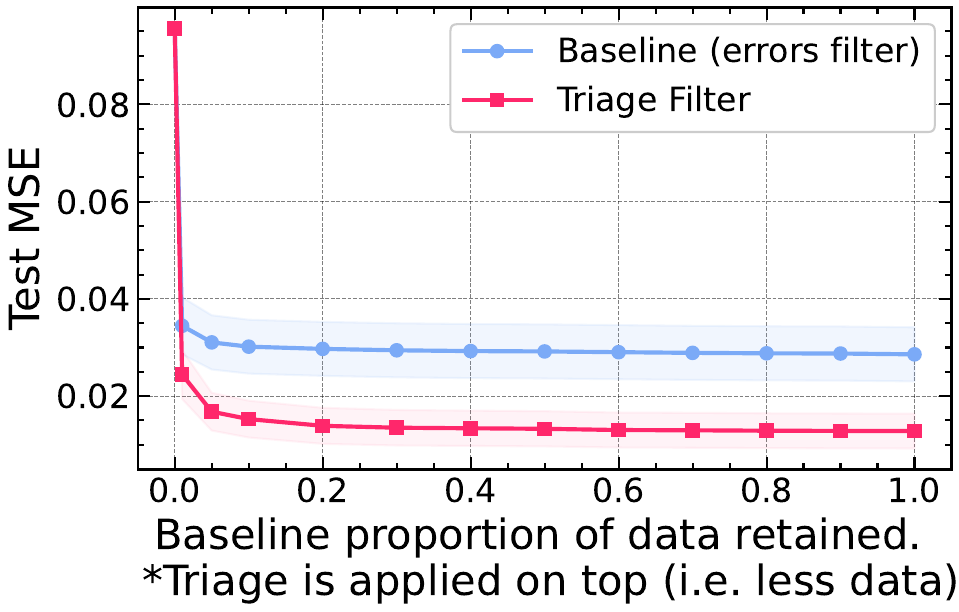}}\quad
  \subfigure[\scriptsize{Loss} ]{\includegraphics[width=0.23\textwidth]{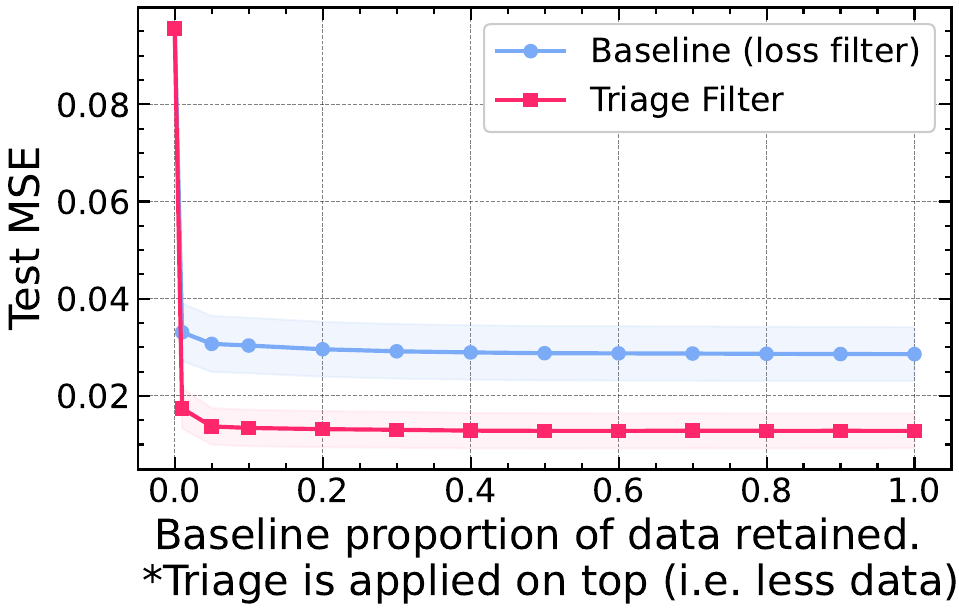}}\quad\\
  \vspace{-2mm}
  \subfigure[\scriptsize{Grad}]{\includegraphics[width=0.23\textwidth]{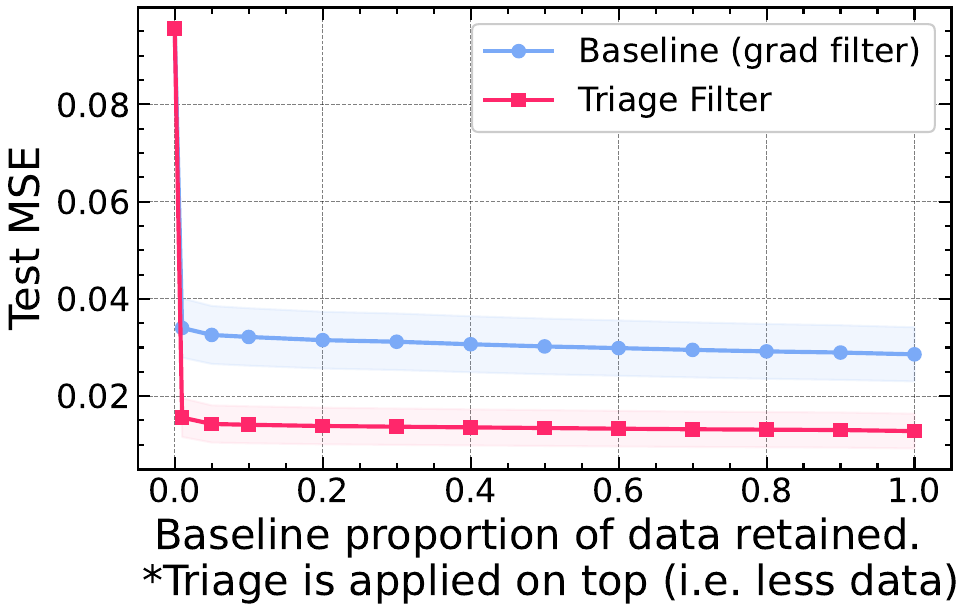}}\quad
  \subfigure[\scriptsize{VoG} ]{\includegraphics[width=0.23\textwidth]{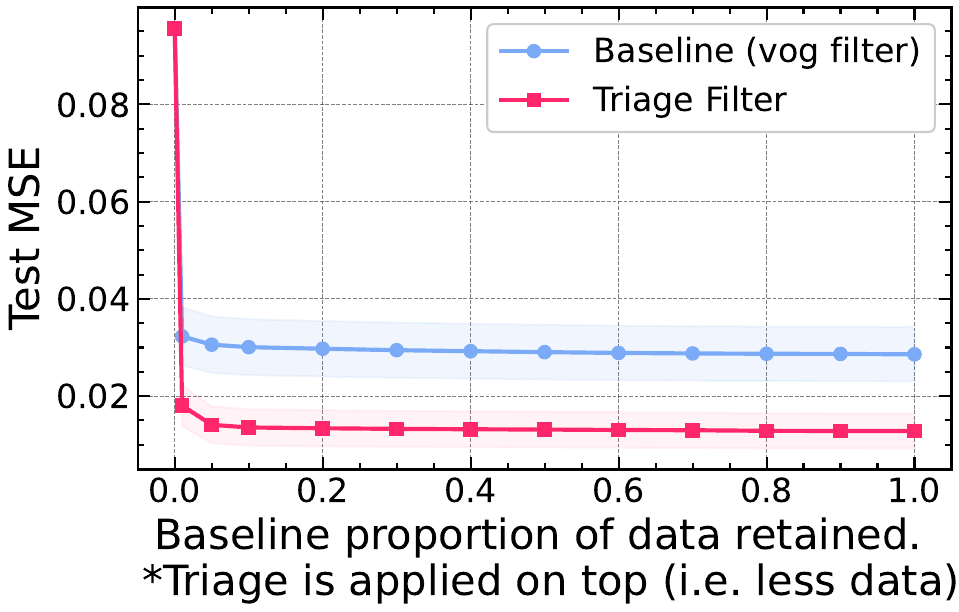}}\quad
  \vspace{-2mm}
  \caption{Fine-grained filter: Mimic}
   \label{fig:filter_exp_mimic}
\vspace{-2mm}
    \rule{\linewidth}{.45pt}
\vspace{-5mm}
\end{figure}

\begin{figure}[!h]
  \centering
  \subfigure[
\scriptsize{Errors}]{\includegraphics[width=0.23\textwidth]{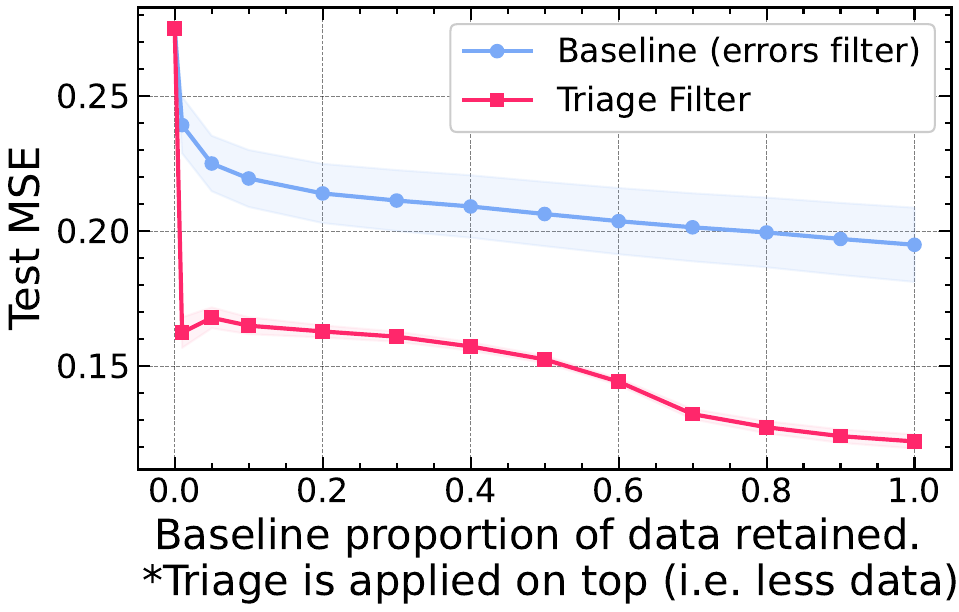}}\quad
  \subfigure[\scriptsize{Loss} ]{\includegraphics[width=0.23\textwidth]{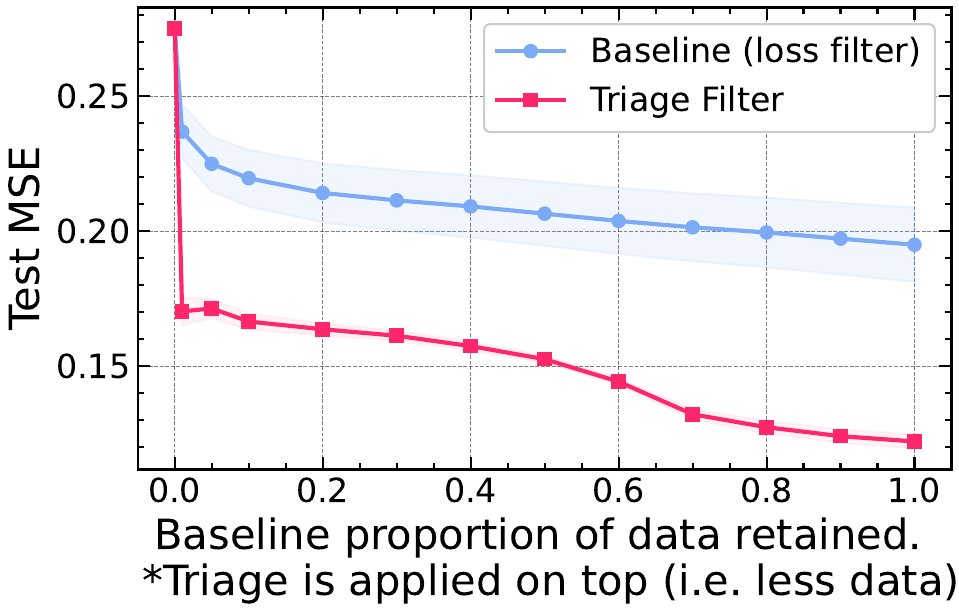}}\quad\\
  \vspace{-2mm}
  \subfigure[\scriptsize{Grad}]{\includegraphics[width=0.23\textwidth]{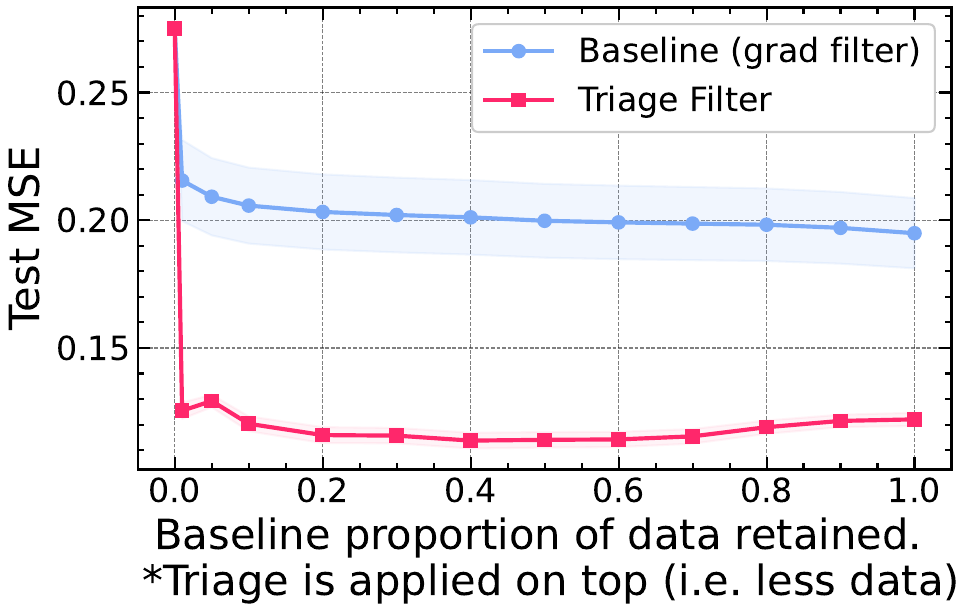}}\quad
  \subfigure[\scriptsize{VoG} ]{\includegraphics[width=0.23\textwidth]{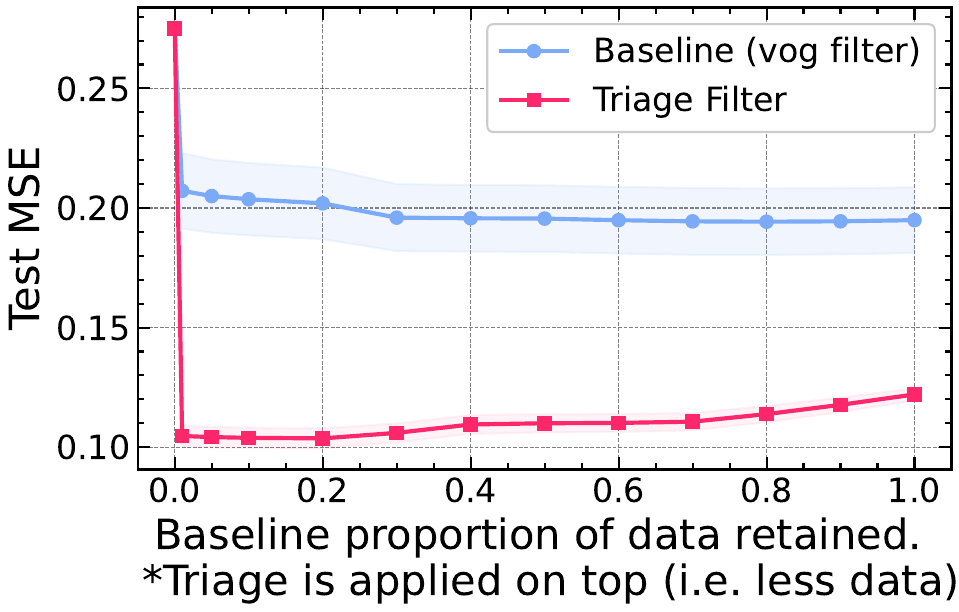}}\quad
  \vspace{-2mm}
  \caption{Fine-grained filter: Protein}
   \label{fig:filter_exp_protein}
\vspace{-2mm}
    \rule{\linewidth}{.45pt}
\vspace{-5mm}
\end{figure}

\begin{figure}[!h]
  \centering
  \subfigure[
\scriptsize{Errors}]{\includegraphics[width=0.23\textwidth]{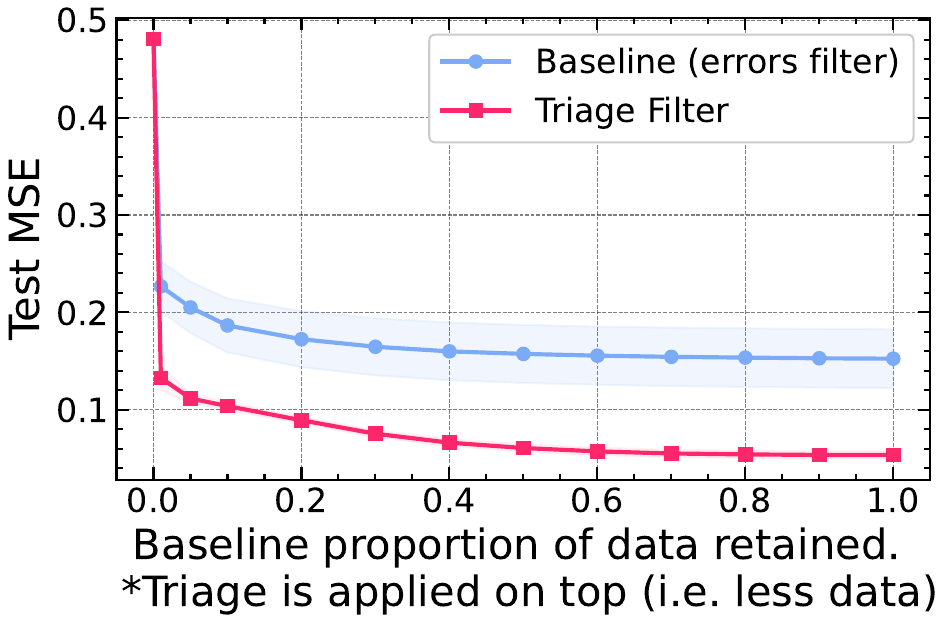}}\quad
  \subfigure[\scriptsize{Loss} ]{\includegraphics[width=0.23\textwidth]{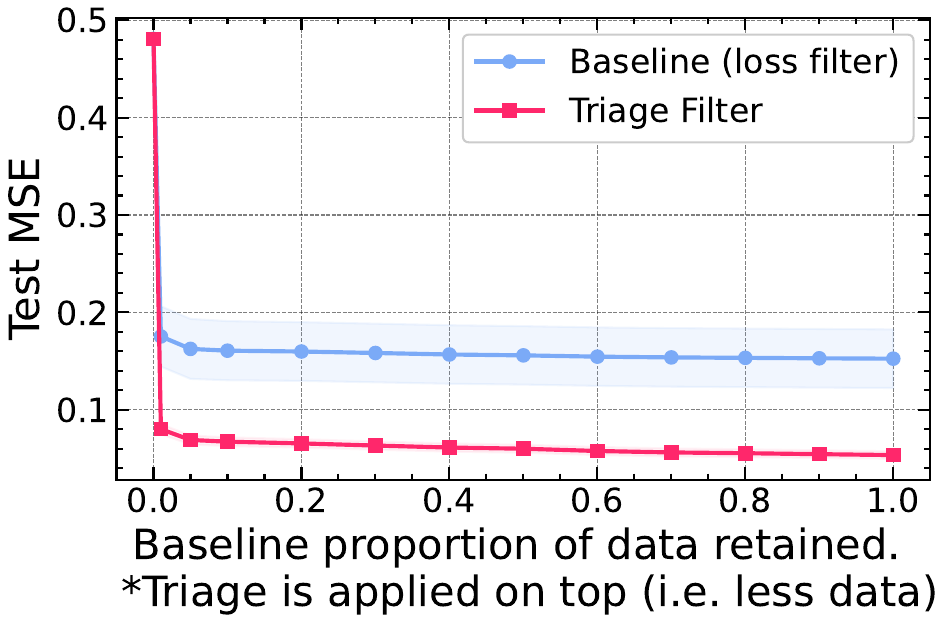}}\quad\\
  \vspace{-2mm}
  \subfigure[\scriptsize{Grad}]{\includegraphics[width=0.23\textwidth]{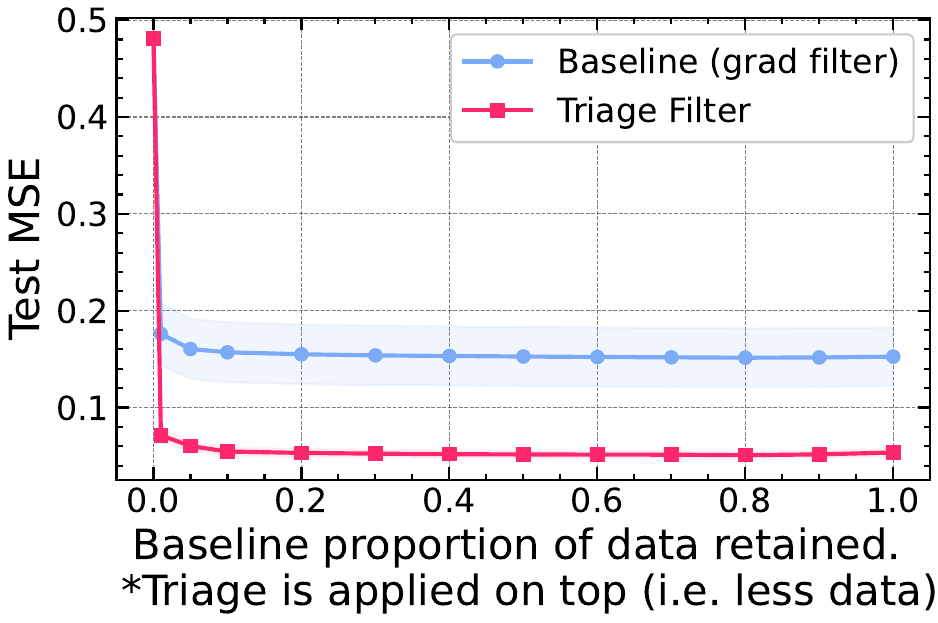}}\quad
  \subfigure[\scriptsize{VoG} ]{\includegraphics[width=0.23\textwidth]{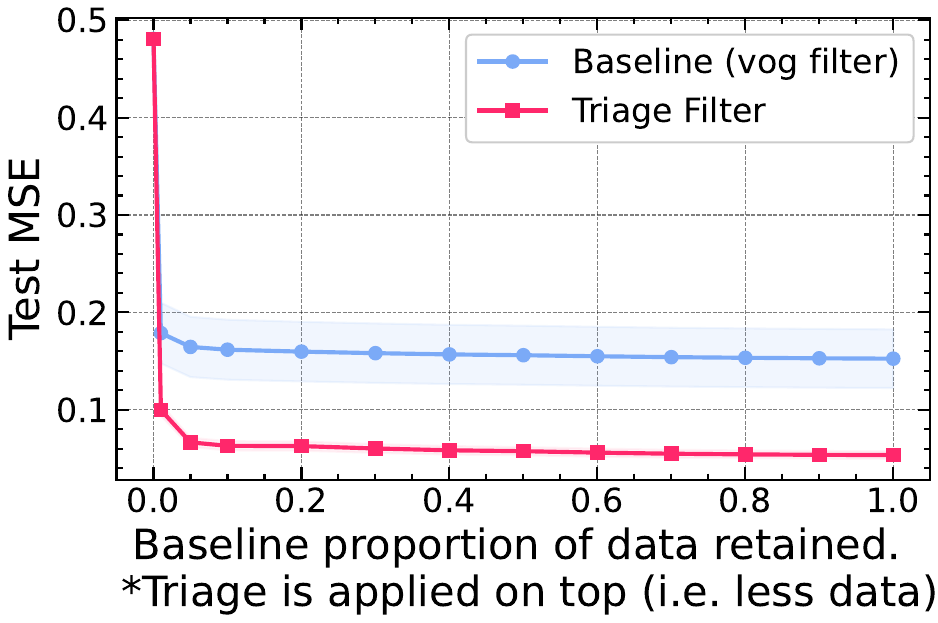}}\quad
  \vspace{-2mm}
  \caption{Fine-grained filter: Star}
   \label{fig:filter_exp_star}
\vspace{-2mm}
    \rule{\linewidth}{.45pt}
\vspace{-5mm}
\end{figure}

\paragraph{Takeaway.} The results shown in Figures \ref{fig:filter_exp_bike}-\ref{fig:filter_exp_star} highlight that indeed less is more. Fitting on more high-quality samples can result it better performance.

\clearpage

\subsection{Value of feature collection/acquisition: Additional datasets}\label{appx:acquire-extra}

\paragraph{Goal.}
The main paper aimed to assess if TRIAGE could quantify the value of data improvement when we are able to acquire features in Section \ref{p5-exp}. We now include results for all datasets (using the same setup,
based on the correlation of the feature with the target). 
Recall we assume that acquiring a valuable feature wrt. the task should increase the proportion of well-estimated samples.

\begin{figure*}[!h]
    \vspace{-4mm}
  \centering
  \subfigure[Well estimated subgroup proportion is increased as informative features are acquired.]{\includegraphics[width=0.4\textwidth]{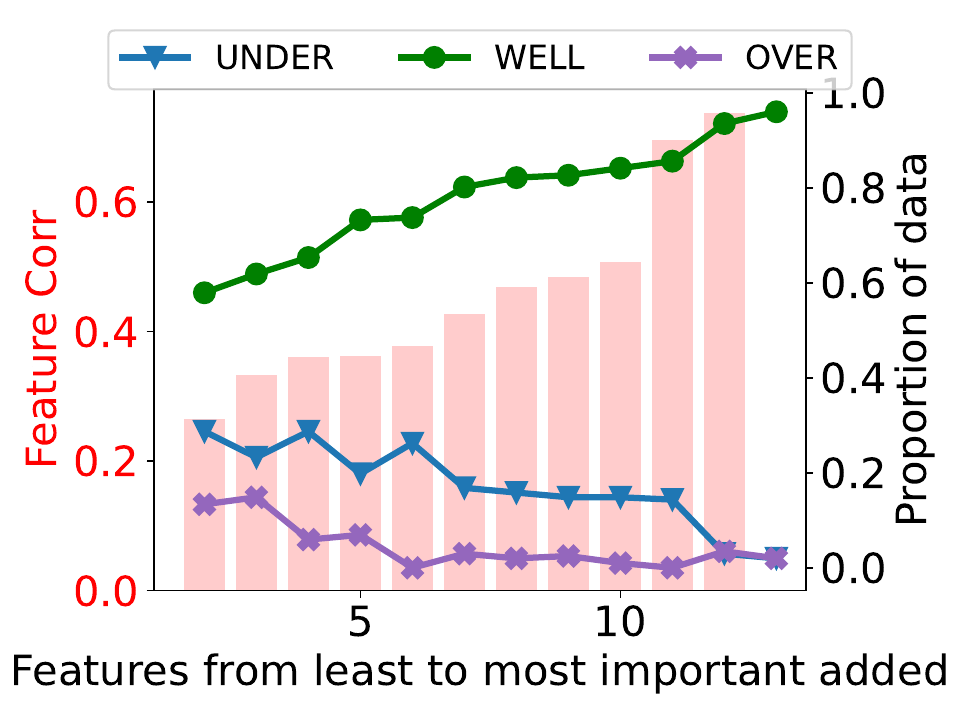}}\quad\quad
 \subfigure[The CPS probabilities for each category remain stable even as the proportions change]{\includegraphics[width=0.4\textwidth]{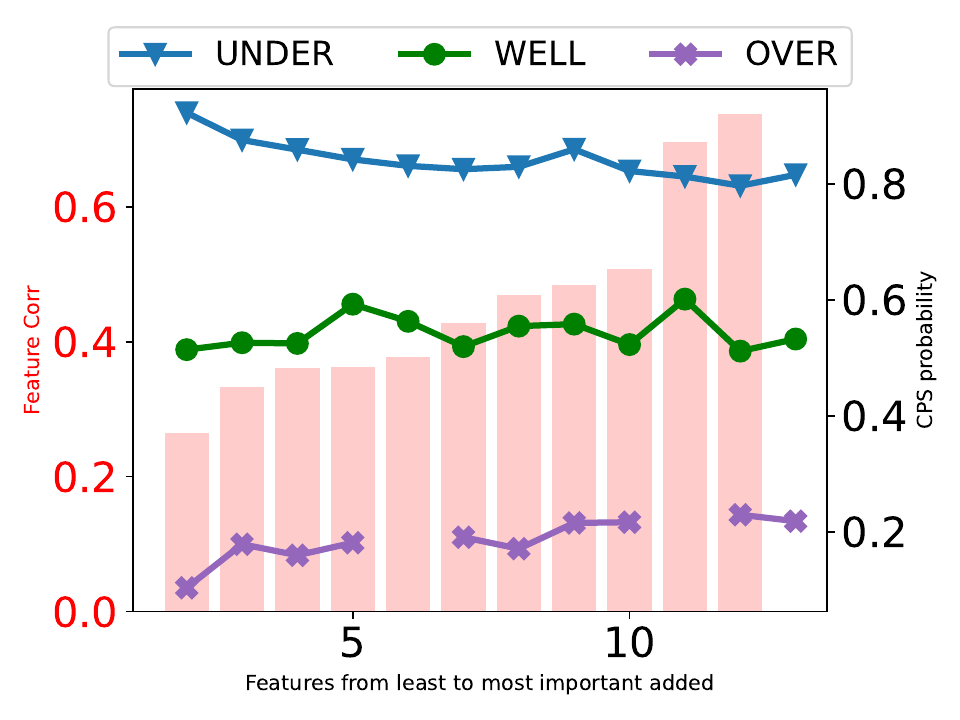}}
  \caption{Boston dataset}
  \label{fig:boston_feat}
  \vspace{-2mm}
\end{figure*}

\begin{figure*}[!h]
\vspace{-3mm}
  \centering
 \subfigure[Well estimated subgroup proportion is increased as informative features are acquired.]{\includegraphics[width=0.4\textwidth]{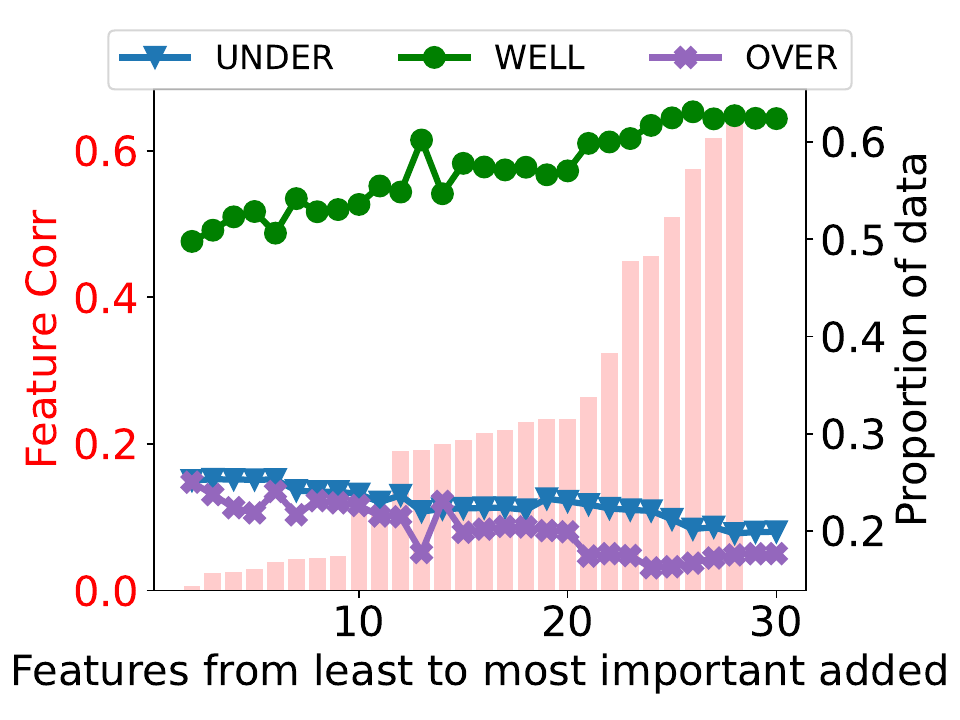}}\quad\quad
 \subfigure[The CPS probabilities for each category remain stable even as the proportions change]{\includegraphics[width=0.4\textwidth]{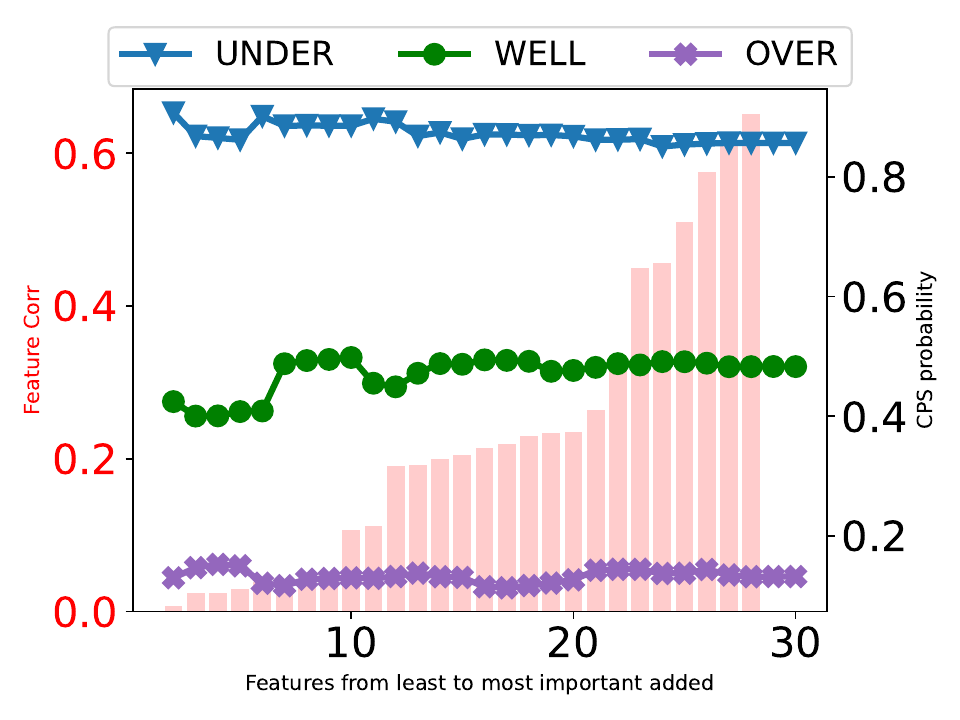}}
  \caption{Prostate dataset}
  \label{fig:prostate_feat}
  \vspace{-2mm}
\end{figure*}

\begin{figure*}[!h]
\vspace{-4mm}
  \centering
  \subfigure[Well estimated subgroup proportion is increased as informative features are acquired.]{\includegraphics[width=0.4\textwidth]{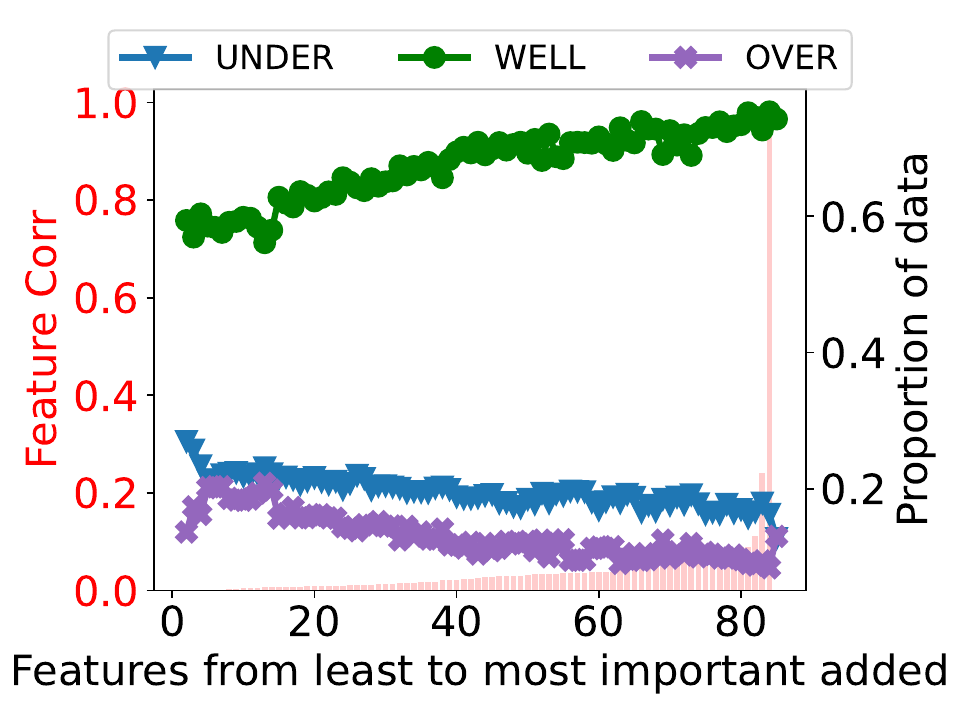}}\quad\quad
\subfigure[The CPS probabilities for each category remain stable even as the proportions change]{\includegraphics[width=0.4\textwidth]{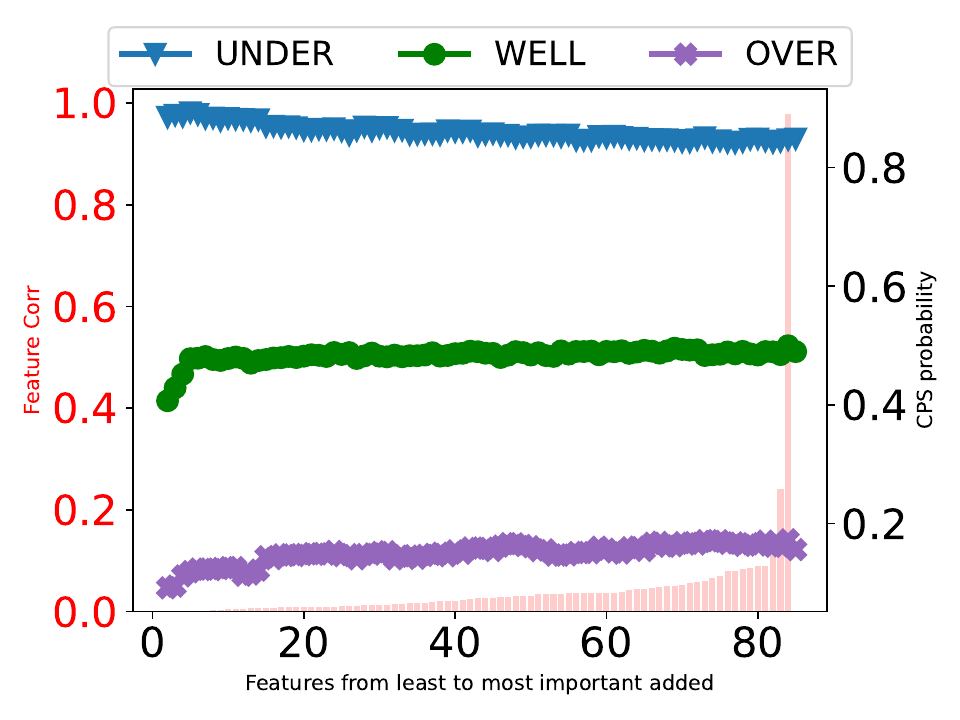}}
  \caption{Mimic dataset}
  \label{fig:mimic_feat}
  
\end{figure*}

\begin{figure*}[!h]

  \centering
  \subfigure[Well estimated subgroup proportion is increased as informative features are acquired.]{\includegraphics[width=0.4\textwidth]{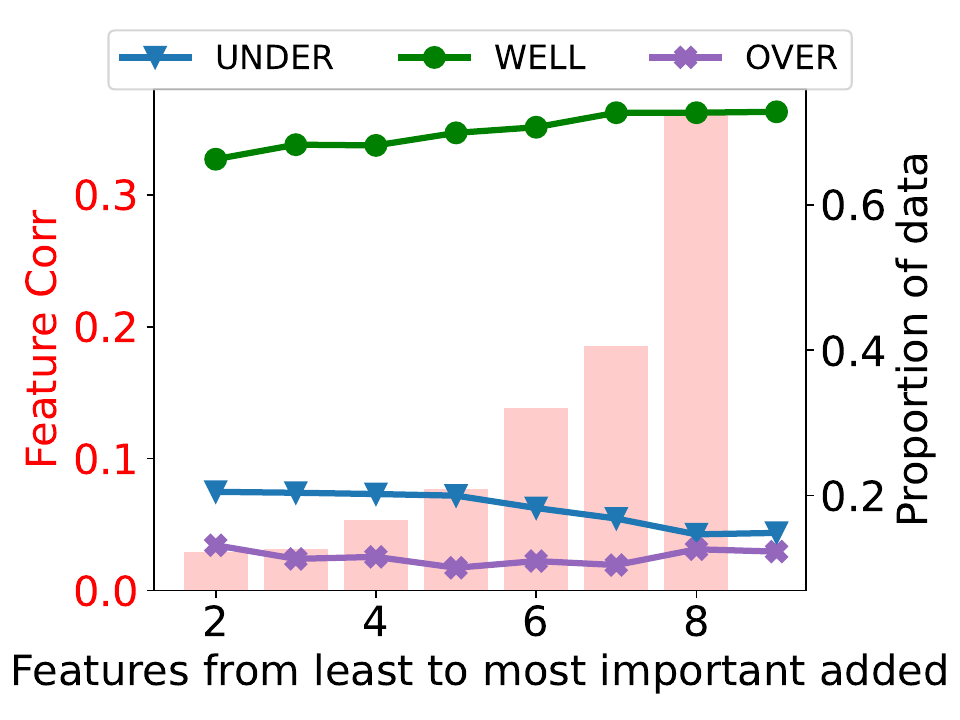}}\quad\quad
 \subfigure[The CPS probabilities for each category remain stable even as the proportions change]{\includegraphics[width=0.4\textwidth]{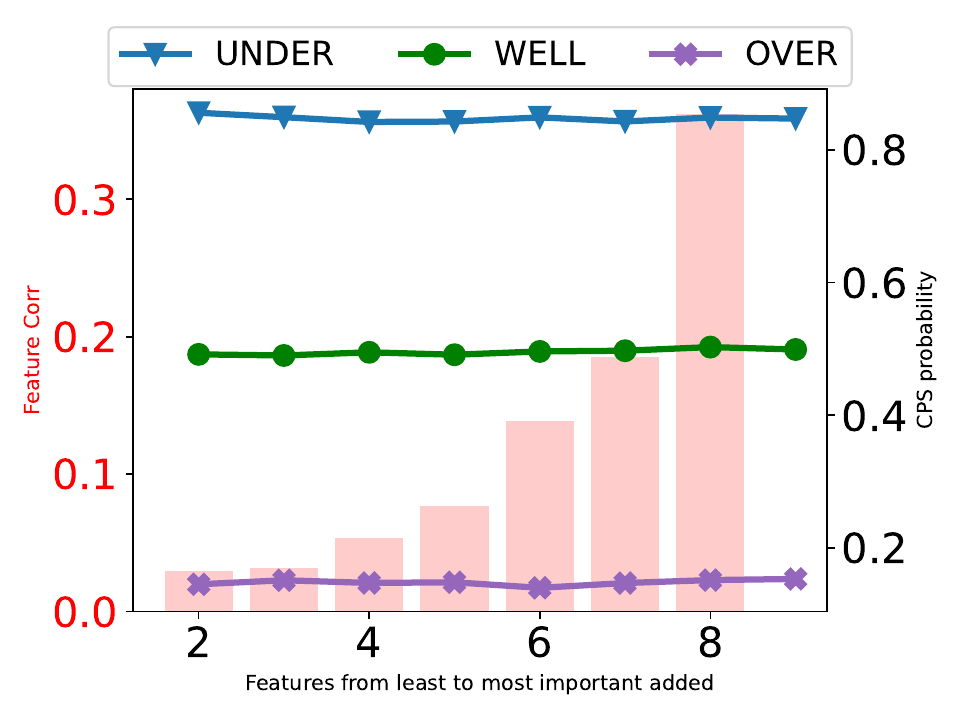}}
  \caption{Bio dataset}
  \label{fig:bio_feat}
\end{figure*}

\begin{figure*}[!h]

  \centering
 \subfigure[Well estimated subgroup proportion is increased as informative features are acquired.]{\includegraphics[width=0.4\textwidth]{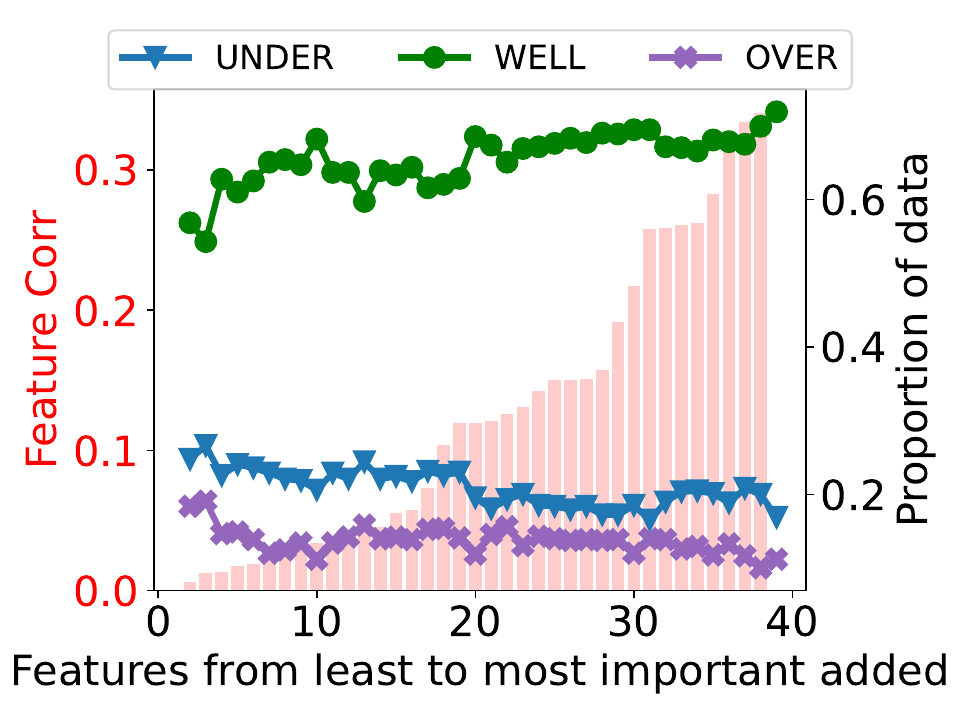}}\quad\quad
\subfigure[The CPS probabilities for each category remain stable even as the proportions change]{\includegraphics[width=0.4\textwidth]{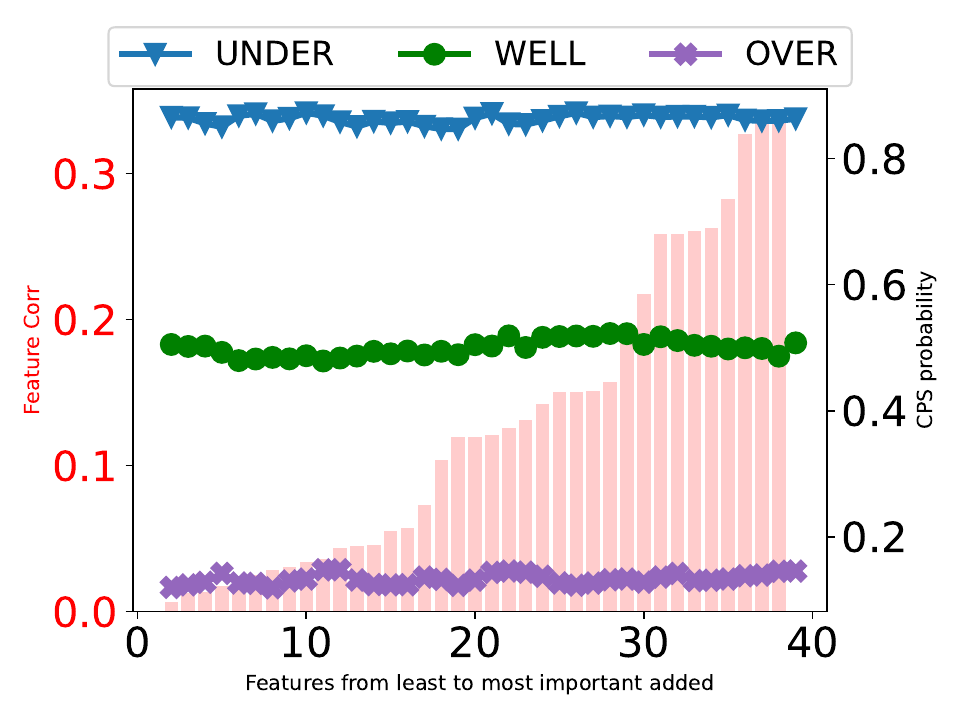}}
  \caption{Star dataset}
  \label{fig:star_feat}
\end{figure*}

\begin{figure*}[!h]

  \centering
  \subfigure[Well estimated subgroup proportion is increased as informative features are acquired.]{\includegraphics[width=0.4\textwidth]{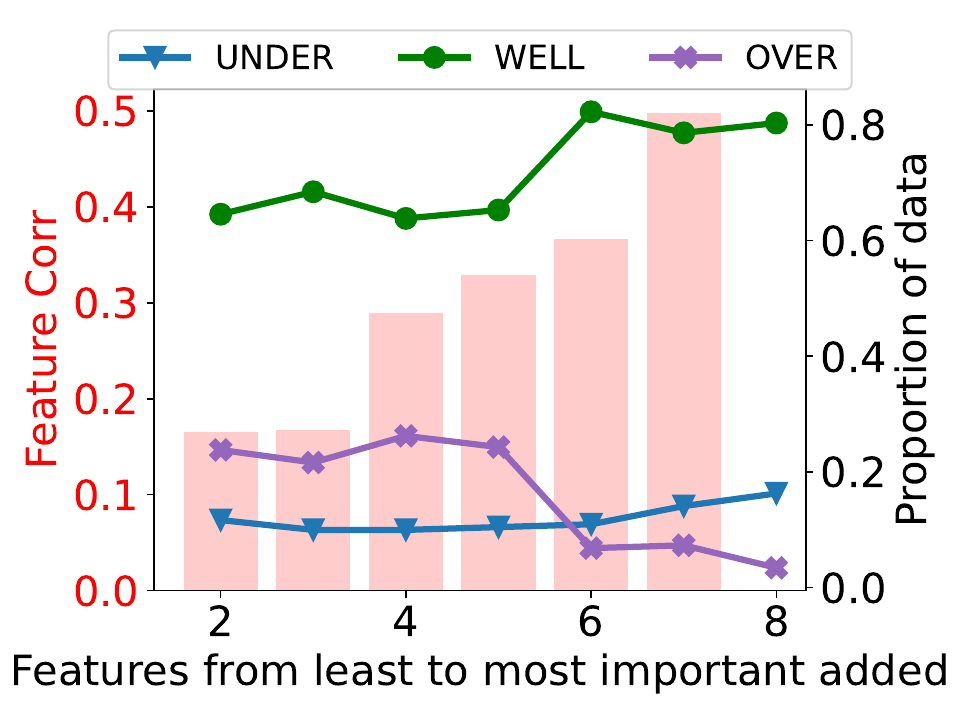}}\quad\quad
\subfigure[The CPS probabilities for each category remain stable even as the proportions change]{\includegraphics[width=0.4\textwidth]{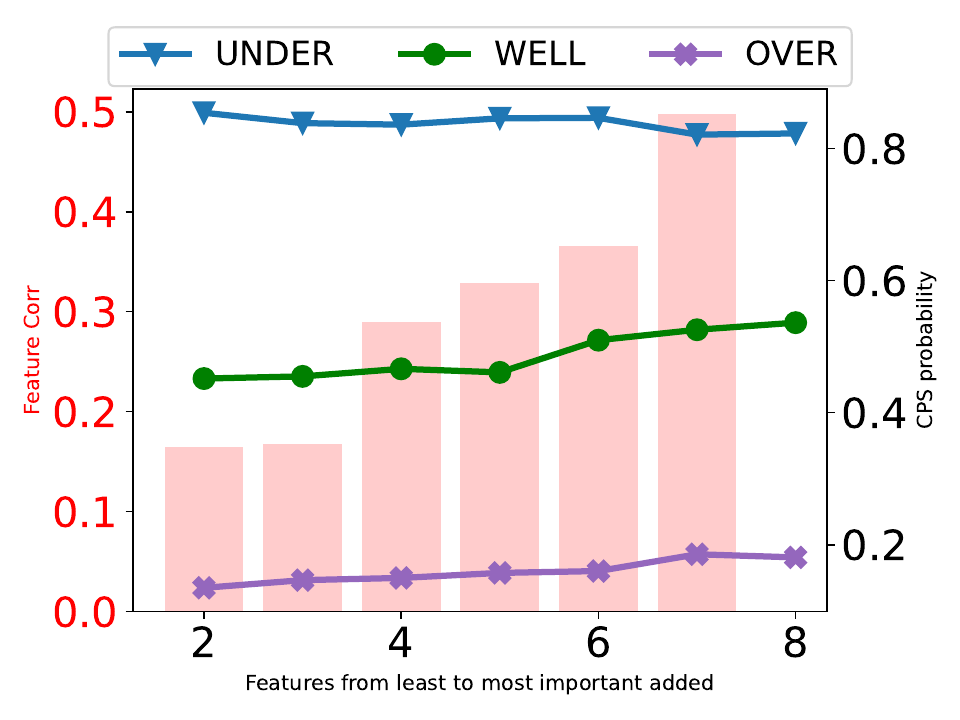}}
  \caption{Concrete dataset}
  \label{fig:concrete_feat}
\end{figure*}

\begin{figure*}[!h]

  \centering
  \subfigure[Well estimated subgroup proportion is increased as informative features are acquired.]{\includegraphics[width=0.4\textwidth]{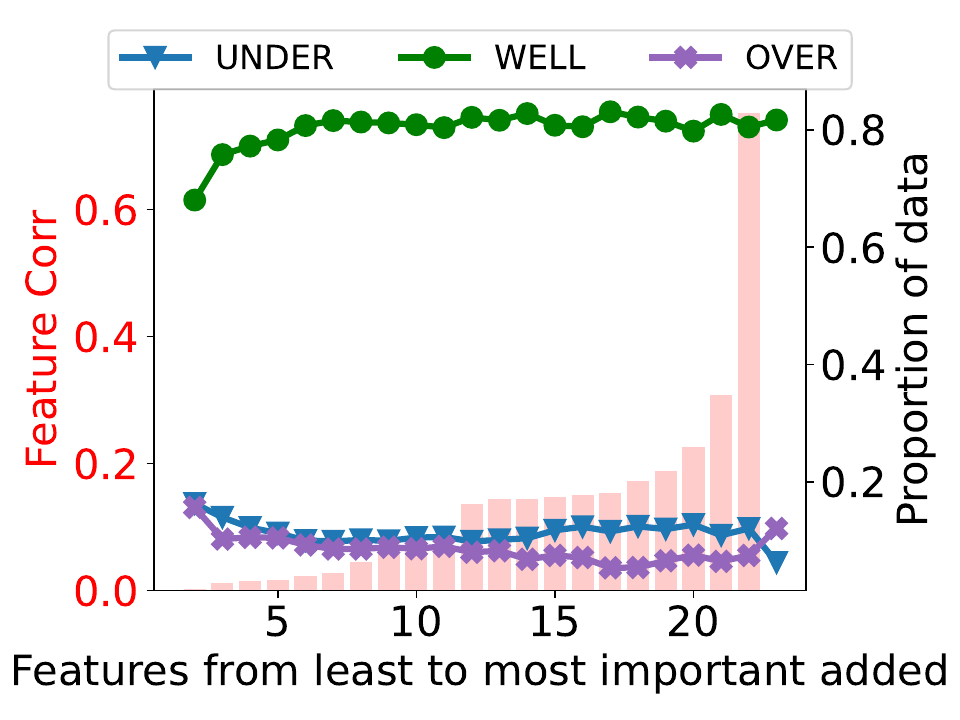}}\quad\quad
\subfigure[The CPS probabilities for each category remain stable even as the proportions change]{\includegraphics[width=0.4\textwidth]{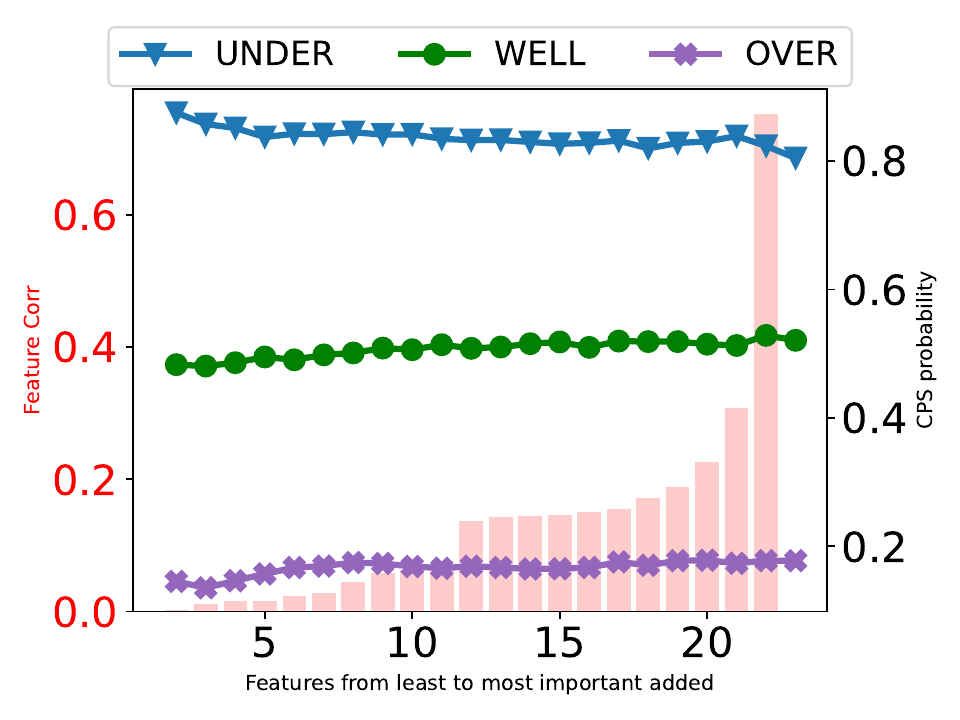}}
  \caption{LoS dataset}
  \label{fig:los_feat}
\end{figure*}

\begin{figure*}[!h]

  \centering
  \subfigure[Well estimated subgroup proportion is increased as informative features are acquired.]{\includegraphics[width=0.4\textwidth]{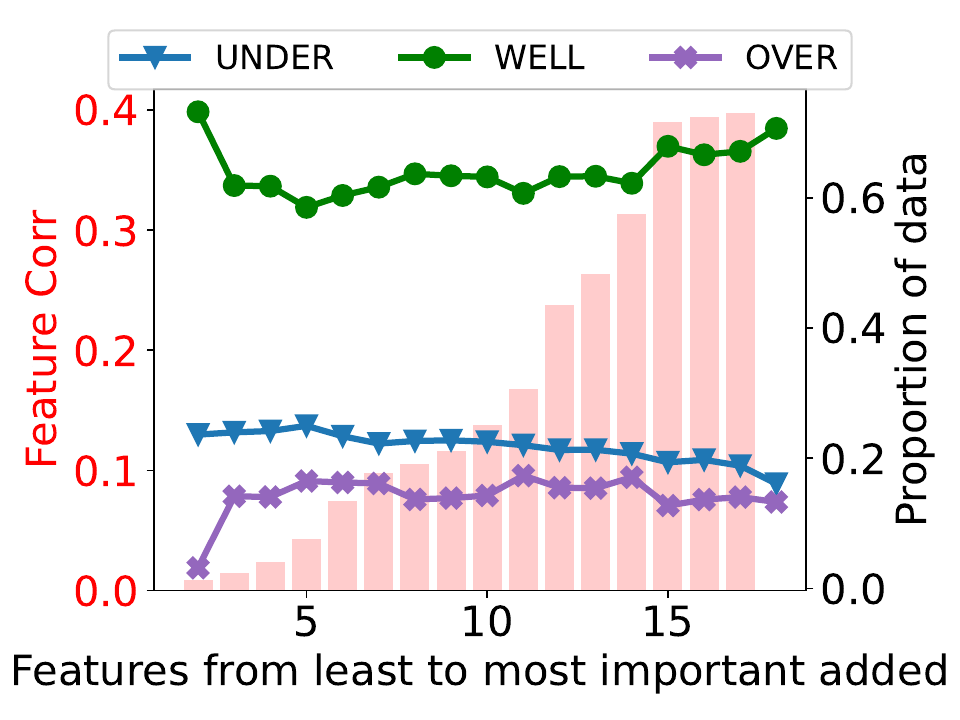}}\quad\quad
  \subfigure[The CPS probabilities for each category remain stable even as the proportions change]{\includegraphics[width=0.4\textwidth]{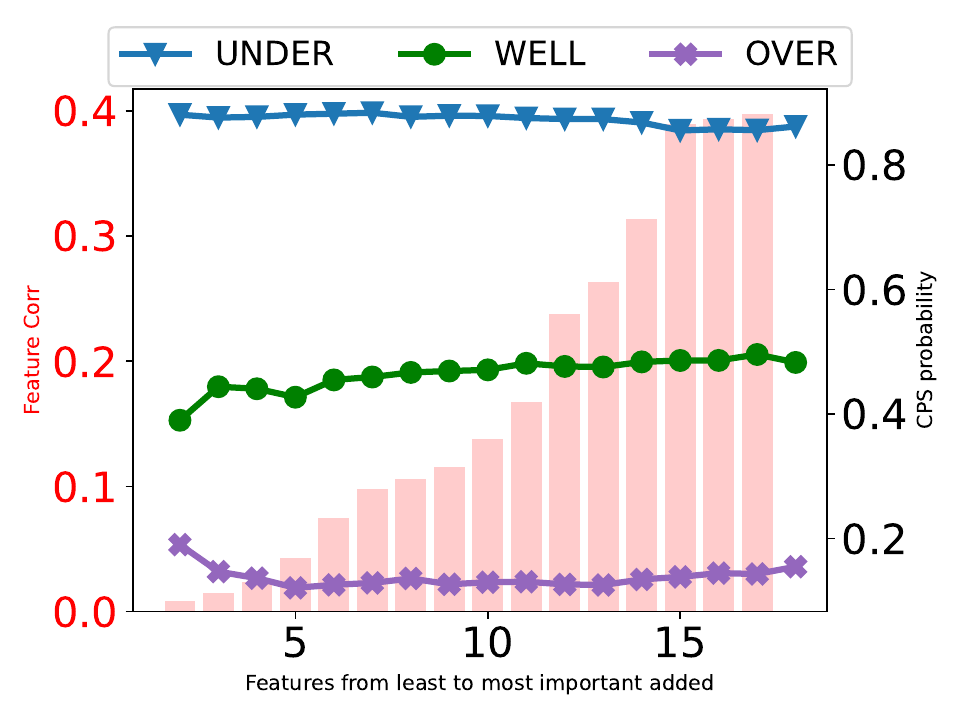}}
  \caption{Bike dataset}
  \label{fig:bike_feat}
\end{figure*}

\begin{figure*}[!h]

  \centering
 \subfigure[Well estimated subgroup proportion is increased as informative features are acquired.]{\includegraphics[width=0.4\textwidth]{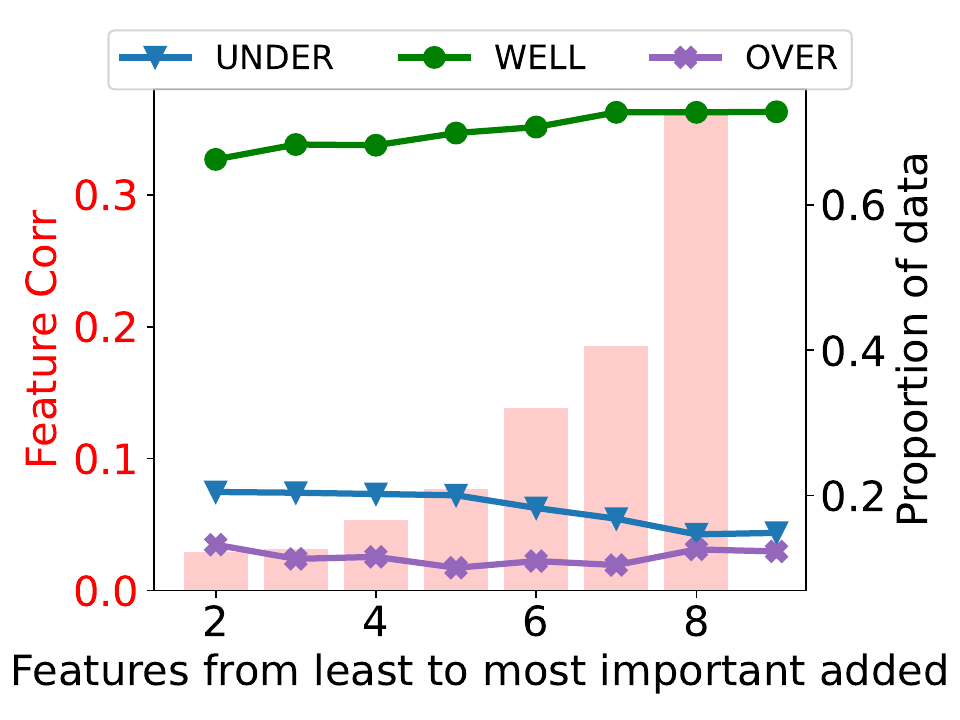}}\quad\quad
\subfigure[The CPS probabilities for each category remain stable even as the proportions change]{\includegraphics[width=0.4\textwidth]{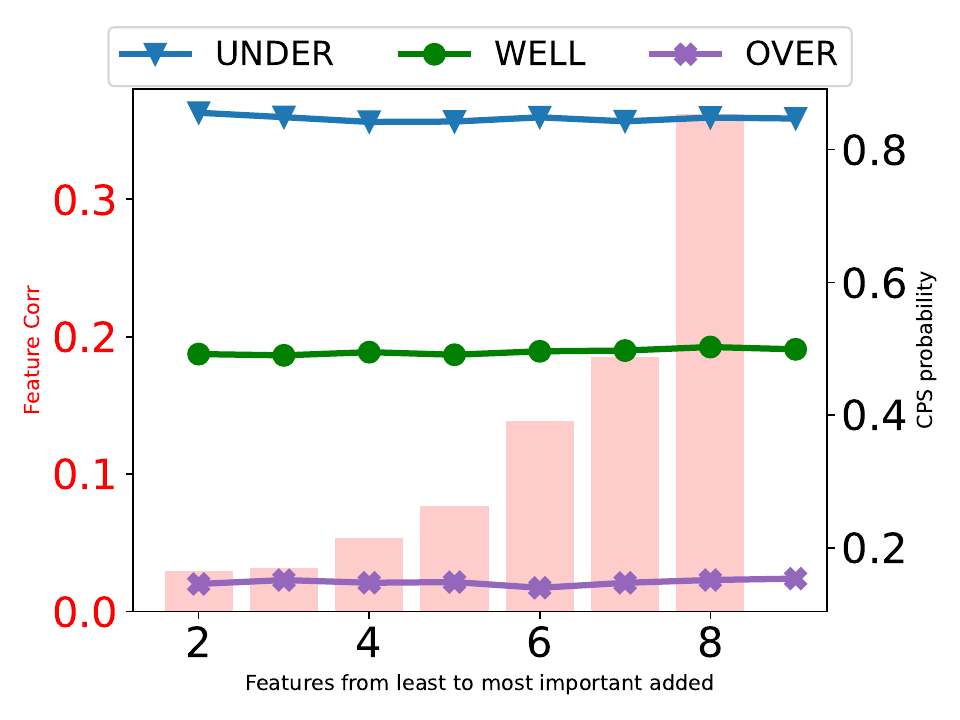}}
  \caption{Protein dataset}
  \label{fig:protein_feat}
\end{figure*}

\paragraph{Takeaway.} We show the results in Figs \ref{fig:boston_feat}-\ref{fig:protein_feat}.  We see similar results to the main text for all datasets. As we acquire more useful/valuable features, the proportions of well-estimated examples increases as desired. Indicating the measure can capture the value of a feature. We also see that despite the category proportions changing, the metrics themselves remain stable. Indicating that we capture inherent properties to these specific samples, which remain consistent.

\end{document}